\documentclass[10pt,journal,compsoc]{IEEEtran}
\usepackage{amsmath,graphicx}
\usepackage{amssymb,amsthm}
\usepackage{subfigure}
\usepackage{algorithm}
\usepackage{algorithmic}
\usepackage[dvipsnames]{xcolor}

\newcommand{\half}{\ensuremath{\frac{1}{2}}}
\newcommand{\rU}{\ensuremath{ \overline{U}}}

\newcommand{\ralpha}{\ensuremath{ \overline{\alpha}}}
\newcommand{\rT}{\ensuremath{ \overline{T}}}
\newcommand{\rt}{\ensuremath{ \overline{t}}}  

\newcommand{\rLambda}{\ensuremath{ {\Lambda}}}
\newcommand{\W}{\ensuremath{ M}}
\newcommand{\rW}{\ensuremath{ \overline{M}}}
\newcommand{\labS}{\ensuremath{I^s}}
\newcommand{\labT}{\ensuremath{I^t}}
\newcommand{\sizeU}{\ensuremath{R}}
\newcommand{\R}{\ensuremath{\mathbb{R}}}
\newcommand{\N}{\ensuremath{N}}
\newcommand{\Ns}{\ensuremath{N_s}}
\newcommand{\Nt}{\ensuremath{N_t}}
\newcommand{\Ks}{\ensuremath{K_s}}
\newcommand{\Kt}{\ensuremath{K_t}}
\newcommand{\Ls}{\ensuremath{L_s}}
\newcommand{\Lt}{\ensuremath{L_t}}
\newcommand{\V}{\ensuremath{\mathcal{V}}}
\newcommand{\E}{\ensuremath{\mathcal{E}}}
\newcommand{\dlambda}{\ensuremath{\delta}}
\newcommand{\dalpha}{\ensuremath{\Delta_\alpha}}
\newcommand{\dT}{\ensuremath{\Delta_T}}
\newcommand{\calpha}{\ensuremath{C}}
\newcommand{\lambdaR}{\ensuremath{\lambda_R}}

\DeclareMathOperator{\sgn}{sgn}

\newtheorem{proposition}{Proposition}

\ifCLASSOPTIONcompsoc
  \usepackage[nocompress]{cite}
\else
  \usepackage{cite}
\fi
\ifCLASSINFOpdf
\else
\fi
\begin{document}
%
\title{Domain Adaptation on Graphs by Learning Aligned Graph Bases}

\author{Mehmet Pilanc{\i}  and Elif Vural
\IEEEcompsocitemizethanks{
\IEEEcompsocthanksitem M. Pilanc{\i} and E. Vural are with the Department of Electrical and Electronics Engineering, METU, Ankara. This work has been supported by the T\"UB\.ITAK 2232 research scholarship.\protect\\
E-mail: mehmet.pilanci@metu.edu.tr, velif@metu.edu.tr
}
}

\IEEEtitleabstractindextext{%
\begin{abstract}

A common assumption in semi-supervised learning with graph models is that the class label function varies smoothly on the data graph, resulting in the rather strict prior that the label function has  low-frequency content. Meanwhile, in many classification problems, the label function may vary abruptly in certain graph regions, resulting in high-frequency components. Although the semi-supervised estimation of class labels is an ill-posed problem in general, in several applications it is possible to find a source graph on which the label function has similar frequency content to that on the target graph where the actual classification problem is defined. In this paper, we propose a method for domain adaptation on graphs motivated by these observations. Our algorithm is based on learning the spectrum of the label function in a source graph with many labeled nodes, and transferring the information of the spectrum to the target graph with fewer labeled nodes. While the frequency content of the class label function can be identified through the graph Fourier transform, it is not easy to transfer the Fourier coefficients directly between the two graphs, since no one-to-one match exists between the Fourier basis vectors of  independently constructed graphs in the domain adaptation setting. We solve this problem by learning a transformation between the Fourier bases of the two graphs that flexibly ``aligns'' them.  The unknown class label function on the target graph is then reconstructed such that its spectrum matches that on the source graph while also ensuring the consistency with the available labels. The proposed method is tested in the classification of  image, online product review, and social network data sets. Comparative experiments suggest that the proposed algorithm performs better than recent domain adaptation methods in the literature in most settings.

\end{abstract}

}

\maketitle

\IEEEdisplaynontitleabstractindextext

%
\IEEEpeerreviewmaketitle

\IEEEraisesectionheading{\section{Introduction}\label{sec:intro}}

%
%
%
%
\IEEEPARstart{M}{ost} 
classification algorithms rely on the assumption that the labeled and unlabeled data samples at hand are drawn from the same distribution. However, in many practical data classification problems, the labeled training samples and the unlabeled test samples may have different statistics \cite{PanY10}. Domain adaptation methods make use of the class labels sufficiently available in a source domain in order to infer the label information in a target domain where labeled data are much more scarce. In order to be able to ``transfer'' the information from one domain to another, some inherent relation must exist between the two domains. In this work, we focus on a setting where the source and the target data are represented with a graph in each domain. We consider that the source and the target graphs are related in such a way that the spectra of the source and the target class label functions on the two graphs share similar characteristics. We then propose a method that makes use of this relation in order to estimate the missing labels in the target domain based on the sufficiently available label information in the source domain.

The domain adaptation problem has attracted much attention in the recent years. Each domain adaptation solution is based on a certain assumption about how the source and the target domains are related. Some methods assume that the data samples from different domains can be aligned via projections and transformations \cite{Fernando2013}, \cite{GongSSG12}, while some try to establish a relation between their distributions \cite{LopezPazHS12}, \cite{SunFS16}, or learn joint feature representations \cite{GaninL15}. Meanwhile, what is common between all these methods is that they are strictly based on the assumption that the data samples reside in an ambient space such as an Euclidean domain, hence they have physical coordinates. Although this may be true in various settings, there are also many data classification problems where the source and the target data are defined or described solely through the pairwise affinities or the relations between data samples. Some examples are social networks \cite{MyersL10}, where no physical coordinates are associated with a user but relations or links between different users define the network; or sensor networks \cite{Jablonski17}, where the pairwise similarities between different sensors are identifiable via their geographical or other kinds of proximities. Graph models provide very convenient tools for such problems. For instance, in a social network each user can be represented as a graph node and relationships between users can be captured with edges. One can then consider an inference problem on the graph, e.g., whether a user is likely to be interested in a product or not.  Similarly, in a sensor network one may infer the missing data at a broken sensor based on the data obtained from the other sensors.

In this work, we propose a new domain adaptation method that uses a source graph and a target graph representing the source and the target data. We consider the problem of estimating a label function on the target graph where very few labels are available. Depending on the application, the label function we consider can be any function defined on a graph domain, whose missing values are to be inferred from the available values. In particular, in a data classification problem, which is the main application area of our work, the values of the label function are class labels. Our assumption about the relation between the source and the target domains is that the spectrum, i.e., the frequency content of the label function has similar characteristics over the source and the target graphs. Given the observations of the label function on the source graph, we estimate the label function on the target graph under the prior that its frequency spectrum resembles that on the source graph.

\begin{figure}[t]
  \centering
  \centerline{\includegraphics[width=5.0cm]{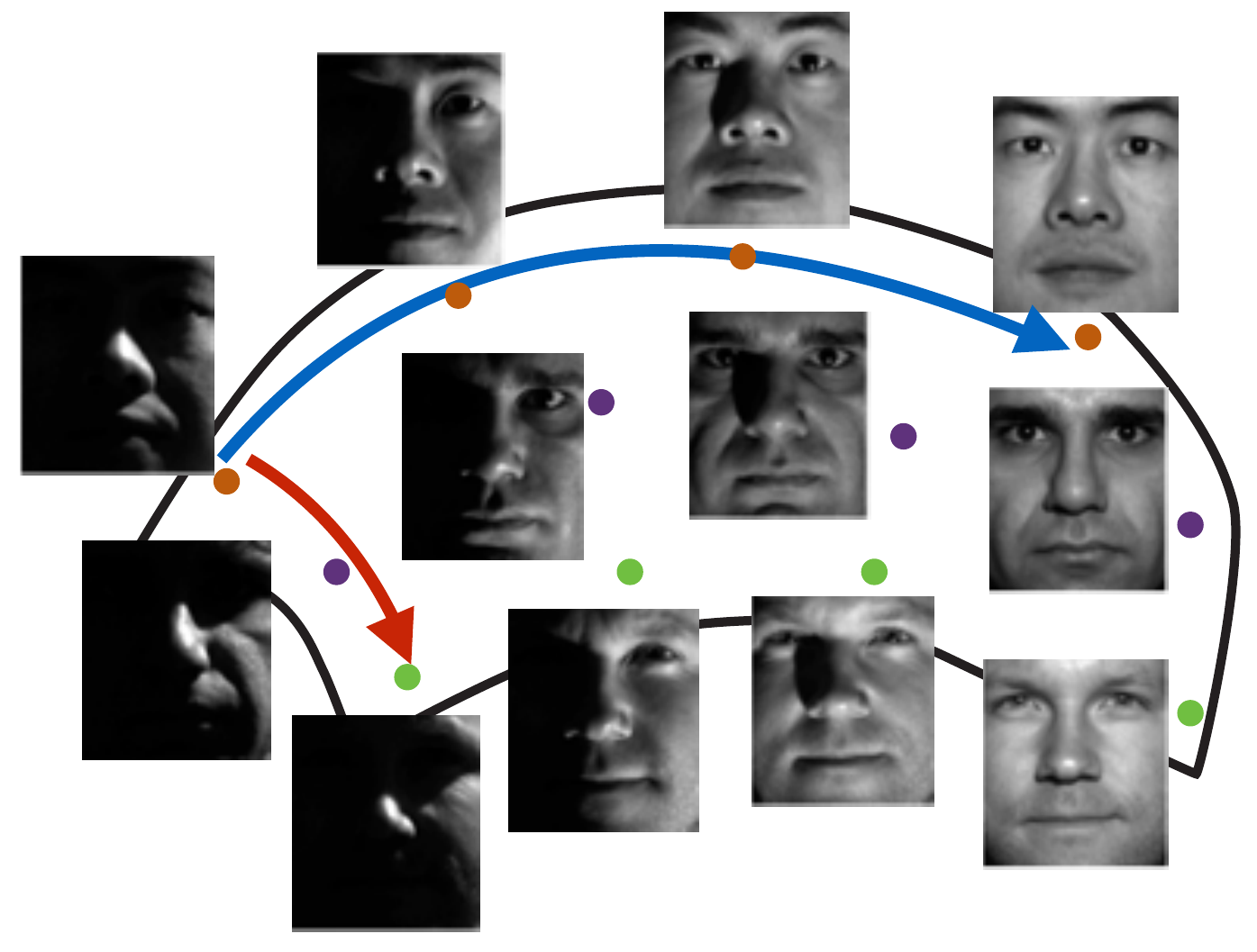}}
  \caption{Illustration of a generic face manifold. Face images  \cite{GeBeKr01} of three different individuals are indicated with different colors. While the class label function varies slowly along the blue direction, it has a relatively fast variation along the red direction.}\medskip
  \label{fig:gen_face_manif}
\end{figure}

\begin{figure}[t]
  \centering
  \centerline{\includegraphics[width=9.0cm]{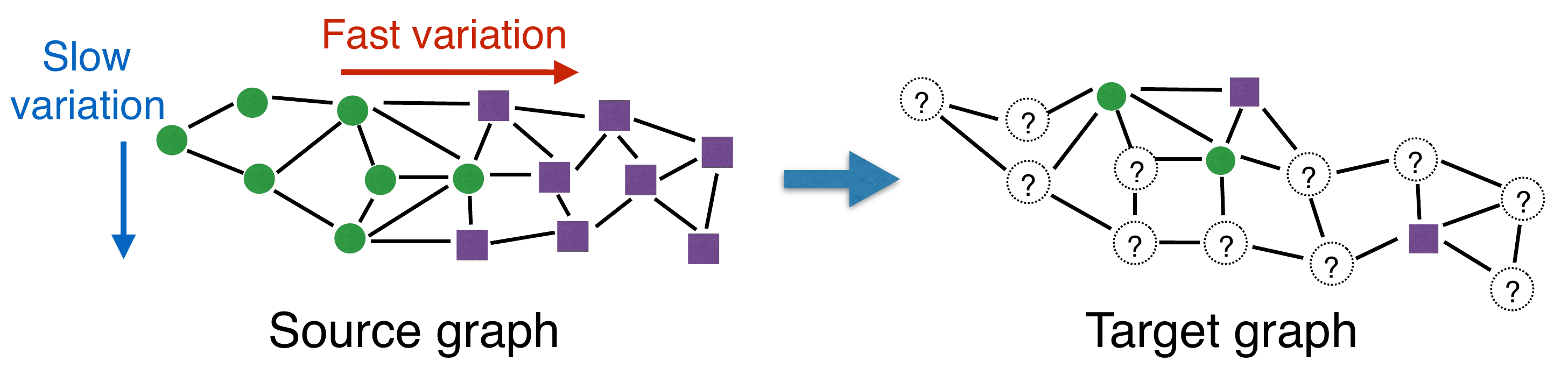}}
  \caption{Illustration of the graph domain adaptation problem studied in this work. Given that the source label function has slow and fast variations along the indicated directions, we would like to transfer this label spectrum information to the target graph in order to estimate the target label function more accurately.}\medskip
  \label{fig:illus_gda}
\end{figure}

Harmonic analysis on graph domains, which permits the definition of the Fourier transform on graphs,  has been an active and popular research topic of the recent years  \cite{HammondVG11}, \cite{ShumanNFOV13}. However, the notion of smoothness or smoothly-varying functions on graphs has actually been essential to many dimensionality reduction and semi-supervised learning methods since a long time \cite{Belkin03}, \cite{ZhouBLWS03}, \cite{ZhuGL03}. Graph-based semi-supervised learning algorithms in a single domain typically rely on the assumption that the label function has a smooth variation on the graph \cite{Zhu08}, \cite{BelkinN04}.  Meanwhile, the validity of the smoothness assumption is questionable in the general sense. For instance, in Figure \ref{fig:gen_face_manif}, a generic face manifold is illustrated, where the face images of different individuals may get arbitrarily close to each other due to extreme lighting conditions. Consequently, the label function has fast variation along certain directions on the data graph and its spectrum contains some non-negligible high-frequency content. Although the assumption that the label function should vary slowly on the graph is reasonable especially in a single domain where no information about its spectral content is available, the spectrum can actually be learnt in a setting with more than one domain. Our method is then based on the idea of learning the spectral content of the label function from the source graph, and transferring it to the target graph as illustrated in Figure \ref{fig:illus_gda}. 

Given a source and a target graph that are independently constructed, we propose to learn a pair of ``aligned'' bases on the two graphs through which information can be transferred or shared. In particular, the ``aligned'' source and target bases are such that the coefficients of the source and target label functions are similar when represented in the corresponding bases. We formulate the basis learning problem as the learning of a linear transformation between the source and the target graph Fourier bases so that each source Fourier basis vector is mapped to a new basis vector in the target graph  obtained as a linear combination of the target Fourier basis vectors. The learning of this transformation then becomes a key problem of the proposed scheme. In particular, the linear transformation to be learnt must be sufficiently flexible to indeed ``align'' the two graphs even if they are independently constructed, while retaining the capability of transferring the spectral content of the label function between the two graph  bases. In order to achieve this, we impose suitable priors on the linear transformation, and then learn the transformation matrix jointly with the source and the target label functions under the constraint that the source and the target label functions must have similar coefficients over the learnt bases. The resulting objective function is not jointly convex in the coefficients and the transformation matrix; nevertheless, it is separately convex in one when the other is fixed. We thus minimize the objective function with an alternating optimization procedure. The output of the algorithm is the estimated label function on the target graph, which provides the class labels of the initially unlabeled data samples. To the best of our knowledge, our treatment is the first to study the domain adaptation problem based on explicitly analyzing the spectrum of label functions in a pure graph setting. Our proposed method is applicable not only to data analysis problems defined purely on graphs, but also to data embedded in an ambient space via the construction of graphs with respect to, e.g., nearest-neighborhoods. We demonstrate the usage of the algorithm in several data classification and regression applications. Classification results on synthetic data, face and object images, and social network data as well as regression results for the prediction of product ratings of users show that the proposed algorithm often outperforms traditional classifiers and reference domain adaptation methods in comparison.

The paper is organized as follows. In Section \ref{sec:related_work}, we overview the related literature. In Section \ref{sec:prob_form}, we present a brief introduction to frequency analysis on graphs and formulate the problem of  graph domain adaptation. In Section \ref{sec:prop_method}, we describe the proposed algorithm for domain adaptation on graphs via spectral graph alignment. In Section \ref{sec:exp_res}, we evaluate the performance of the method with comparative experiments. Finally, we conclude in Section \ref{sec:concl}.


\section{RELATED WORK}
\label{sec:related_work}

The domain adaptation problem has been treated in several settings and under different assumptions so far \cite{PanY10}. Some works focus on a problem where the source and the target distributions are defined on the same data space \cite{HuangSGBS06}, \cite{DaiXYY07}, \cite{BlitzerMP06}. In the case that the conditional distributions of labels remain unchanged and only the marginal distributions of data coordinates vary between the source and target domains, the domain adaptation problem is referred to as covariate shift or sample selection bias, where solutions based on sample reweighting are applicable \cite{HuangSGBS06}, \cite{SunCPY11}. Daum\'e III et al.~and Duan et al.~have proposed to map the source and the target features to a higher dimensional domain via feature augmentation, where a common classifier can be learnt \cite{DaumeIII07}, \cite{DaumeKS10}, \cite{Daume2010}, \cite{DuanXT12}. In settings with multiple sources domains, a common approach is to learn the target hypothesis based on a weighted combination of the source hypotheses \cite{CrammerKW08}, \cite{WuWZTXYH17}.  

Another domain adaptation solution consists of learning a transformation or a projection that aligns the source and the target data \cite{Fernando2013}, \cite{GongSSG12},  \cite{PanKY08}, \cite{FangZ13}, \cite{WangM08},  \cite{Wang2010}, \cite{ZhangLO17}. In fact, the idea of aligning the source and the target domains by mapping them to an intermediate space through a transformation has been at the core of many domain adaptation algorithms, some of which can also be applied to problems where the source and the target samples reside in different ambient spaces \cite{WangM11}. Several authors have proposed to reduce the distance between the samples from different domains by learning a transformation \cite{MuandetBS13}, \cite{ShiLFYZ10}, \cite{CourtyFTR17},  where the maximum mean discrepancy is a common choice as a distribution distance \cite{GongZLTGS16}, \cite{BaktashmotlaghHLS13}, \cite{PanTKY11}, \cite{LongWDPY14} or scatter measure \cite{GhifaryBKZ17}. The approaches in \cite{LopezPazHS12}, \cite{SunFS16}, \cite{TranRP17} rely on matching the densities or the second-order statistics of the source and the target domains via copula functions or transformations. A metric adapted to the domain adaptation problem is learnt in \cite{XuPXWLMS17},  \cite{HerathHP16a}.  In some works, a classifier is learnt in a joint manner with the mapping \cite{YaoPNLM15}, \cite{ChengP14}, or directly in the original data domain based on a self-training principle \cite{LuSC0H18}. 

Deep networks have also gained popularity in domain adaptation applications in the recent years. These methods are typically based on the extraction of domain-invariant features that are shared between \cite{GaninL15, LongC0J15} or adapted specifically \cite{TzengHSD17} to the source and the target domains. Domain classifier layers aiming to reduce the distribution discrepancy are often learnt along with the label predictors in an adversarial manner \cite{LongZ0J17, TzengHSD17}.

While all of the above domain adaptation methods rely strictly on the availability of a representation of the data in an ambient space, in thus study, we focus on a setting where the data does not need to have a physical embedding and the problem may be directly defined over an abstract data graph. Frequency analysis on graph domains is now a well-established framework, thanks to the recent advances in the field of graph signal processing. The convergence of the graph Laplacian operator to the continuous Laplace-Beltrami operator on manifolds has been studied in several works \cite{HeinAv05}, \cite{Singer06}, which provides a foundation for graph signal processing. Characterizing the Fourier basis vectors as the eigenvectors of the Laplacian operator, the Fourier transform and Fourier bases can be extended to graph domains via the eigenvalue decomposition of the graph Laplacian matrix \cite{HammondVG11},  \cite{ShumanNFOV13}, \cite{Chung96}.

The idea of matching graph bases with transformations or pairwise correspondences  has been explored before in the previous works \cite{EynardKBGB15}, \cite{PokrassBBSS13}, \cite{RodolaCBTC17}; however, in different settings related to unsupervised clustering or 3D shape analysis problems. Note that, several previous methods have already incorporated manifold models or graph models in domain adaptation. The algorithm in \cite{WangM11} employs a manifold model and learns projections by preserving the topology of the data set while achieving discrimination between different classes. The works in \cite{ChengP14},  \cite{XiaoG15},  \cite{DhillonTC12} similarly impose priors on the smoothness of the label function over the data graph. The K-NN graphs used in \cite{DhillonTC12} are iteratively refined with the aid of a supervised metric learner. A pair of source and target graphs are constructed in \cite{BanerjeeBBBCB15}, which is followed by a graph matching stage to map source classes to target clusters for multispectral image classification.

Finally, a preliminary version of our work has been presented in \cite{PilanciV16}, where the idea of transferring the label spectrum between a source and a target graph has been explored for the first time. However, a major limitation of the algorithm in \cite{PilanciV16} is that it relies on a one-to-one match between the graph Fourier basis vectors. This restricts its applicability to settings where the source and the target graphs are highly similar so as to admit a direct match between the two graph Fourier bases. This limitation is circumvented in the current study by learning a transformation between the two Fourier bases.



\section{DOMAIN ADAPTATION ON GRAPHS}
\label{sec:prob_form}

In this section, we first give an overview of the extension of classical frequency analysis techniques to graph domains \cite{ShumanNFOV13}. Then, we propose a problem formulation for domain adaptation on graphs. In the following, matrices are represented with uppercase letters, and vectors are denoted with lowercase or Greek letters. Vectors are considered as column vectors unless stated otherwise.  $A_{ij}$ stands for the $(i,j)$-th entry of a matrix $A$, and $|\cdot|$ denotes the cardinality of a set.

\subsection{Overview of Frequency Analysis on Graphs}

In graph-based methods, a data set with $\N$ data samples is typically represented with a graph with $\N$ vertices, such that each vertex corresponds to a data sample.
Let $G=(\V, \E, W)$ be a weighted graph with $\N$ vertices (nodes), where $\V=\{ x_i\}_{i=1}^\N$ is the set of vertices, $\E$ is the set of edges, and $W \in \R^{\N \times \N}$ is the weight matrix. If there is an edge between the nodes  $x_i$ and $x_j$, then $W_{ij}$ consists of the weight of this edge. If the nodes $x_i$ and $x_j$ are not connected with an edge, then $W_{ij}=0$.

A graph signal is a function $f: \V \rightarrow \R$  taking a real value on each graph vertex, which can equivalently be represented as an $\N$-dimensional vector $f \in \R^\N$. A set $\{ v_k\}_{k=1}^\N \subset \R^\N$ of  linearly independent graph signals  form a graph basis, so that any graph signal $f$ can be represented as
\begin{equation}
f= \sum_{k=1}^\N \alpha_k v_k
\end{equation}
in terms of the graph basis vectors $v_k$ with coefficients $\alpha_k$. Representing the basis as a matrix $V= [v_1 \dots v_\N]  \in \R^{\N \times \N}$ and the coefficient vector as $\alpha = [\alpha_1 \dots \alpha_\N ]^T \in \R^\N $, the graph signal can be expressed  as $f= V \alpha$.

The graph Laplacian matrix $L \in \R^{\N \times \N}$ is defined as $L=D-W$, where $D \in \R^{\N \times \N}$ is the diagonal degree matrix given by $D_{ii} = \sum_j W_{ij}$. The graph Laplacian is an essential element in spectral graph theory, since its application to a graph signal $f$ as an operator via the matrix multiplication
\begin{equation}
(Lf) (x_i) = \sum_{j=1}^\N W_{ij} (f (x_i) - f (x_j))
\end{equation}
is the graph equivalent of applying the Laplacian operator to a signal in classical signal processing \cite{ShumanNFOV13}, \cite{HeinAv05}, \cite{Singer06}. This analogy allows the extension of classical Fourier analysis to graph domains as follows. First recall that for one-dimensional signals, the complex exponentials $e^{j \Omega t}$ defining the Fourier transform are given by the eigenfunctions of the Laplacian operator $\Delta$ 
\begin{equation}
\label{eq:comp_exp_eigfunc}
- \Delta (e^{j \Omega t}) =  \Omega^2 e^{j \Omega t}.
\end{equation}  
The eigenvalue $\Omega^2$ of the Laplacian operator increases with the frequency of the complex exponential $e^{j \Omega t}$. Characterizing the Fourier transform via the eigenfunctions of the Laplacian operator, the graph counterparts of complex exponentials are then the eigenvectors of the graph Laplacian given by
\begin{equation}
\label{eq:fourier_basis_comp}
L u_k = \lambda_k u_k.
\end{equation}
The set of eigenvectors $\{ u_k \}_{k=1}^\N \subset \R^\N$ of the graph Laplacian corresponding to the eigenvalues $\lambda_1=0 \leq \lambda_2 \leq \dots \leq \lambda_\N$ thus defines a graph Fourier basis. In analogy with \eqref{eq:comp_exp_eigfunc}, the eigenvalues $\lambda_k$ bear a notion of frequency in a graph. The eigenvectors $u_k$ for increasing values of $k$ indeed have an increasing speed of variation over the graph when regarded as graph signals \cite{ShumanNFOV13}. In particular, a common measure for the speed of variation of a graph signal $f$ over the graph is
\begin{equation}
f^T L f = \half \sum_{i,j=1}^\N W_{ij} (f(x_i) - f(x_j))^2,
\end{equation}
which takes larger values if the function $f$ varies more abruptly between neighboring graph nodes. The above term becomes the corresponding eigenvalue $\lambda_k$ of the graph Laplacian when the graph signal is taken as a Fourier basis vector $f=u_k$
\begin{equation}
u_k^T L u_k = \lambda_k.
\end{equation}
Once the Fourier basis $\{ u_k \}_{k=1}^\N$ of a graph is computed, the graph Fourier transform $\hat f (\lambda_k)$ of a graph signal $f$ is simply given by its inner product with the basis vectors
\begin{equation}
\hat f (\lambda_k) = \langle f, u_k  \rangle = \sum_{i=1}^\N f (x_i) u_k (x_i).
\end{equation}
This can be equivalently written as $\hat f = U^T f \in \R^\N$ in matrix notation, where $\hat f = [\hat f(\lambda_1) \dots \hat f (\lambda_\N) ]^T$ and $U=[u_1 \dots u_\N] \in \R^{\N \times \N}$. Here $\hat f (\lambda_k)$ is the $k$-th Fourier coefficient of $f$ corresponding to the basis vector $u_k$ with frequency $\lambda_k$. The inverse Fourier transform is then obtained as the reconstruction of the signal from its representation over the Fourier basis as
\begin{equation}
f = \sum_{k=1}^\N \hat f (\lambda_k) u_k  = U \hat f.
\end{equation}
%


\subsection{Problem Formulation for Graph Domain Adaptation}
\label{sec:prob_formul}

We now propose our problem formulation for domain adaptation in graph settings. We consider a source graph $G^s=(\V^s, \E^s, W^s)$ that consists of $\Ns$ vertices $\V^s=\{ x^s_i \}_{i=1}^{\Ns}$ and edges $\E^s$, and a target graph $G^t=(\V^t, \E^t, W^t)$ with $\Nt$ vertices $\V^t=\{ x^t_i \}_{i=1}^{\Nt}$ and edges $\E^t$. The weighted edges of the source and the target graphs are respectively represented in the weight matrices $W^s$, $W^t$. Let $U^s \in \R^{\Ns \times \Ns}$ and $U^t \in \R^{\Nt \times \Nt}$ denote the matrices containing the Fourier basis vectors, respectively on the source and the target graphs. These are computed using the eigenvalue decompositions of the respective graph Laplacians $L^s \in \R^{\Ns \times \Ns}$ and $L^t \in \R^{\Nt \times \Nt}$ as explained in \eqref{eq:fourier_basis_comp}.

Consider a label function  $f^s \in \R^{\Ns}$ on the source graph and a label function $f^t \in \R^{\Nt}$ on the target graph. We assume that the labels of some of the nodes are available. We denote the known labels as  $ y^s_i = f^s(x^s_i) $ on the source graph (for labeled $x^s_i$), and as $ y^t_i = f^t(x^t_i) $ on the target graph (for labeled $x^t_i$). The sets containing the indices of the labeled data samples are denoted as  $\labS \subset \{ 1, \dots, \Ns \}$ and $\labT \subset \{ 1, \dots, \Nt \}$ in the source and the target domains.
%
The label functions $f^s$ and $f^t$ take discrete values in a classification problem and  continuous values in a regression problem. For instance, in a classification problem with two classes, one can set $y^t_i$ as equal to $1$ if the labeled data sample $x^t_i$ belongs to the first class and as $-1$ if it is from the second class. The problem is then to compute the labels of all unlabeled data samples, which is done by estimating the label vector $f^t$. Domain adaptation methods often assume a setting with many labeled samples in the source domain and much fewer labeled samples in the target domain, i.e., $| \labT |  \ll | \labS | $. 

Let $V^s \in \R^{\Ns \times \Ns}$ and $V^t \in \R^{\Nt \times \Nt}$ denote a pair of bases for the functions respectively on the source and the target graphs. We can then decompose the label functions $f^s$ and $f^t$ to be predicted in the source and target graphs over the bases $V^s$ and $V^t$ as
\begin{equation}
f^s = \sum_{k=1}^{\Ns} \alpha^s_k v^s_k = V^s \alpha^s,
\qquad
f^t = \sum_{k=1}^{\Nt} \alpha^t_k v^t_k = V^t \alpha^t .
\end{equation}
Here $V^s$ and  $V^t $ contain respectively the basis vectors $\{ v^s_k \}$ and $\{ v^t_k \}$ in their columns; and $\alpha^s \in \mathbb{R}^{\Ns}$ and $\alpha^t \in \mathbb{R}^{\Nt}$ are coefficient vectors.

Domain adaptation methods assume the presence of a relationship between the source and the target domains and aim to transfer the knowledge in the source domain to the target domain in order to better predict the target label function. In the following, we consider a domain adaptation setting where a relationship can be established between the source and the target domains via a ``coherent'' pair of bases $V^s$, $V^t$ for the space of functions on the source and the target graphs.  In particular, if $V^s$ and $V^t$ are a ``coherent'' pair of bases, then one can transfer the label information from the source graph to the target graph based on the representations of the label functions on these bases. We can then formulate the following problem:

 \textbf{Problem 1.} 
 \begin{equation}
\label{eq:prob_pass_alpha}
 \min_{\alpha^s, \alpha^t}  
  \| S^s V^s \alpha^s - y^s  \|^2  
 +  \| S^t V^t \alpha^t - y^t  \|^2  
 + \mu \|  \ralpha^s - \ralpha^t \|^2 
\end{equation}

Here $y^s \in \R^{\Ks}$ and $y^t \in \R^{\Kt}$ are vectors consisting respectively of the available labels $\{y^s_i\}$ and $\{y^t_i\}$  in the source and the target domains where $\Ks=| \labS |$ and $\Kt = |\labT| $ are the number of known labels. The matrices $S^s \in \R^{\Ks \times \Ns}$ and $S^t \in \R^{\Kt \times \Nt}$ are binary selection mask matrices consisting of 0's and 1's, which enforce the label prediction functions $f^s$, $f^t$ to match the available labels  $y^s$, $y^t$ over the subsets $\labS$, $\labT$ of labeled data; and $\mu>0$ is a weight parameter. The coefficients $\alpha^s$ and $\alpha^t$ of the source and target label functions must be found such that the resulting estimation of the label predictions correspond to the given labels, while $\alpha^s$ and $\alpha^t$ (or their appropriately restricted versions $\ralpha^s$,  $\ralpha^t$ in the case that the graph sizes are different $\Ns \neq \Nt$) are close over the source and the target graphs.
 
Then, an important question is what properties a ``coherent'' pair of bases $V^s$ and $V^t$ should have, and how such bases can be found in practice. If a one-to-one match exists between the source and the target graphs, e.g., as in a problem where each source node has a known corresponding target node, then one can simply select the bases as the Fourier bases $V^s = U^s$, $V^t = U^t$, so that the spectra of the source and the target label functions can be directly matched by solving the problem in \eqref{eq:prob_pass_alpha}. However, in a realistic setting such a one-to-one match often does not exist. For instance, the experiments reported in Section \ref{ssec:data_sets}, Figure \ref{fig:mitcbcl_spectra} study the frequency content of the label function on the source and the target graphs. The results in Figure \ref{fig:mitcbcl_spectra} indicate that the general shape (envelope) of the spectrum resembles between the two graphs; however, corresponding Fourier coefficients across the two graphs are not always the same. This suggests that although it would be too restrictive to transmit the exact Fourier coefficients, it is possible to exploit the similarity between the shapes of the source and the target spectra. Based on these observations, we propose to learn $V^s$, $V^t$ relying on the available observations of the label function, in a manner that allows the transfer of the spectral content between the graphs. In particular, we propose to choose
\begin{equation}
 V^s = U^s,  \quad V^t = U^t T 
 \end{equation}
where $U^s$ and $U^t$ are the Fourier bases, and the matrix $T \in \R^{\Nt \times \Nt} $ represents a transformation between $U^t$ and $V^t$. From Problem 1, one can observe that $T$ matches the source basis vector $v_i^s = u_i^s$ to the target basis vector 
\begin{equation}
\label{eq:form_vkt}
v_i^t = \sum_{j=1}^{\Nt} T_{ji} u_j^t
\end{equation}
obtained as a linear combination of the Fourier vectors $u_j^t$.


\begin{figure}[t]
  \centering
  \centerline{\includegraphics[width=7.0cm]{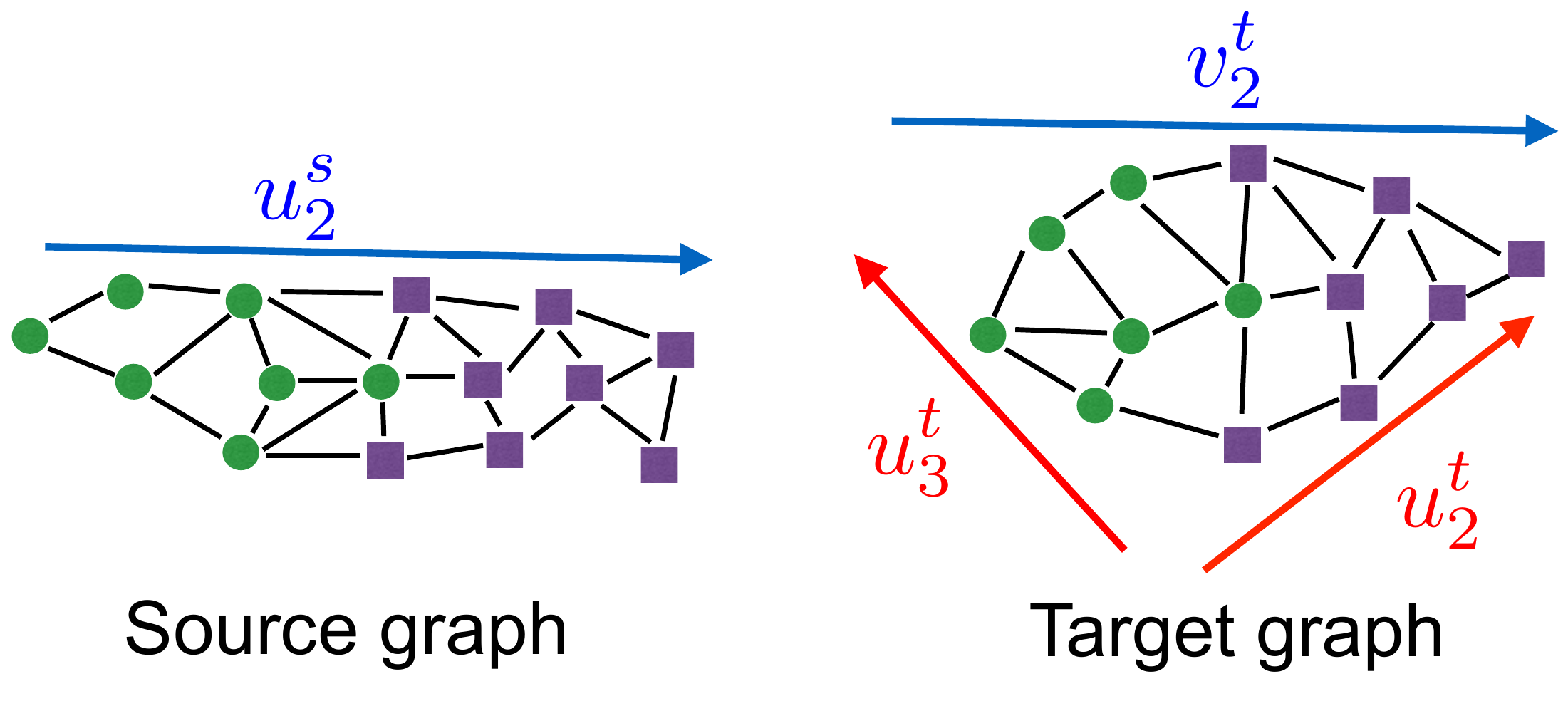}}
  \caption{Illustration of the transformation between similar frequencies in the proposed method. The figure illustrates a case where the second Fourier signal $u_2^s$ in the source graph and the second and third Fourier signals $u_2^t$ and $u_3^t$ in the target graphs oscillate mainly along the indicated directions. Due to the differences between the graph topologies, $u_2^s$ can be successfully matched to neither $u_2^t$ nor $u_3^t$. Nevertheless, $u_2^s$ might possibly be matched to some signal $v_2^t$ in the target graph that can be written as a linear combination of $u_2^t$ and $u_3^t$.}\medskip
  \label{fig:T_illus}
\end{figure}

When learning the transformation $T$, our purpose is to learn a representation that is flexible enough to properly ``align'' the two individually constructed graphs, while also preserving the spectral relation between the two graphs. The rate of variation of the $i$-th source Fourier vector $v_i^s = u_i^s$ is proportional to the $i$-th eigenvalue $\lambda_i^s$ of the source graph Laplacian $L^s$. In order to preserve the spectral relation between the graphs, the corresponding target vector $v_i^t$ in \eqref{eq:form_vkt} must have a similar rate of variation on the target graph, so that slowly (or rapidly) varying source label functions are matched to slowly (or rapidly) varying target label functions. For this reason, we propose to learn $T$ such that the weight $T_{ji}$ of the $j$-th target Fourier vector $u_j^t$ in the representation of $v_i^t$ is encouraged to be higher for $j$ values close to $i$, and to decay as $j$ deviates from $i$. In this way, the source Fourier vector $ u_i^s = v_i^s $ is mapped to a target vector that is mainly composed of the target Fourier vectors $u_j^t$ having frequencies close to that of $u_i^s$, as illustrated in Figure \ref{fig:T_illus}. This can be achieved by penalizing high magnitudes for the entries of $T$ distant from the diagonal, by including a term $ \| \W \odot T \|^2$ in the overall objective, where $\W \in \R^{\Nt \times \Nt}$ is a symmetric weight matrix of the form
\begin{equation}
\label{eq:Wij_defn}
\W_{ij} = \exp \left( \frac{(i-j)^2}{\sigma^2} \right),
\end{equation}
the scale parameter $\sigma$ adjusts the width of the window of matched frequencies, and $\odot$ denotes the Hadamard (element-wise) product between two matrices. The overall objective function to minimize then becomes the following:

\textbf{Problem 2.}

\begin{equation}
\label{eq:prob_form_fullbasis}
\begin{split}
 \min_{\alpha^s, \alpha^t, T}  
 & \| S^s U^s \alpha^s - y^s  \|^2  
 +  \| S^t U^t T \alpha^t - y^t  \|^2  \\
 &+ \mu_1 \|  \ralpha^s - \ralpha^t \|^2 + \mu_2 \| \W \odot T \|_F^2 \\
&  \text{subject to } \sum_{i=1}^{\Nt}T_{ij}^2 = 1 , \text{ for } \ j = 1,\ \dots,  \Nt. \\
 \end{split}
\end{equation}
Here $\mu_1>0$, $\mu_2>0$ are weight parameters, and $\| \cdot \|_F$ denotes the Frobenius norm of a matrix. The equality constraints ensure that the columns of the transformation matrix $T$ have unit norm, in order not to approach the trivial solution $T=0$.

While Problem 2 aims to learn a pair of complete bases on the two graphs, it is often not necessary to use all basis vectors for obtaining a good reconstruction of the label function: Fourier basis vectors $u_i^s$, $u_i^t$ with very high frequencies (eigenvalues) $\lambda_i^s$, $\lambda_i^t$, have a quite rapid variation over the graph, and discarding some of these not only reduces the complexity of the problem, but also serves the important purpose of regularization. For these reasons, we select a subset of the basis vectors $\{u_i^s\}_{i=1}^\sizeU$, $\{ u_i^t \}_{i=1}^\sizeU$, corresponding to the $\sizeU$ smallest frequencies in both domains, where $\sizeU< \Ns$ and $\sizeU< \Nt$. Let $\rU^s \in \mathbb{R}^{\Ns \times \sizeU }$, $\rU^t \in \mathbb{R}^{\Nt \times \sizeU }$ denote the reduced source and target Fourier bases consisting of the first $\sizeU$ basis vectors. When label functions are reconstructed with the reduced bases, Problem 2 can be reformulated as 

\textbf{Problem 3.}
\begin{equation}
\label{eq:prob_form_general}
\begin{split}
 \min_{\ralpha^s, \ralpha^t, \rT}  
 & \| S^s \rU^s \ralpha^s - y^s  \|^2  
 +  \| S^t \rU^t \rT \ralpha^t - y^t  \|^2  \\
 &+ \mu_1 \|  \ralpha^s - \ralpha^t \|^2 + \mu_2 \| \rW \odot \rT \|_F^2 \\
&  \text{subject to } \sum_{i=1}^{\sizeU} \rT_{ij}^2 = 1 , \text{ for } \ j = 1,\ \dots,  \sizeU. \\
 \end{split}
\end{equation}

Here, the matrix $\rT \in \R^{\sizeU \times \sizeU}$ is the submatrix of $T$ consisting of its first $\sizeU$ rows and columns, which match the source vectors $\{u_i^s\}_{i=1}^\sizeU$ to linear combinations of $\{ u_i^t \}_{i=1}^\sizeU$. The reduced weight matrix  $\rW \in R^{\sizeU \times \sizeU}$ has entries as defined in \eqref{eq:Wij_defn}. The vectors $\ralpha^s$, $\ralpha^t$ consist of the projections of the label functions onto the Fourier vectors in the reduced bases $\rU^s$, $\rU^t$ such that the source and the target label functions $f^s$ and $f^t$ are reconstructed as
\begin{equation}
f^s = \rU^s  \ralpha^s, 
\quad
f^t = \rU^t \rT  \ralpha^t
\end{equation}
once Problem 3 is solved. Note that, although the main focus in domain adaptation is to estimate the target labels, the above formulation also allows the estimation of the missing source labels in case of interest. 

Estimating the label functions by solving Problem 3, one may then wonder how well the variations of the source and target label functions on the two graphs agree. In the following, we provide an upper bound on the difference between the rates of change of the source and the target label functions $f^s$ and $f^t$. Let $0= \lambda_1^s \leq \lambda_2^s \leq \dots \leq \lambda_{\sizeU}^s$ and $0 =\lambda_1^t \leq \lambda_2^t \leq \dots \leq \lambda_{\sizeU}^t$ respectively denote the smallest $\sizeU$  eigenvalues of the source and the target graph Laplacians $L^s$ and $L^t$. Let the similarity of the source and the target graph topologies be so that the deviation between the corresponding eigenvalues of the two graph Laplacians are bounded as $| \lambda_i^s - \lambda_i^t | \leq \dlambda$, for all $i=1, \dots, \sizeU$. Let us define $\lambdaR = \max(\lambda_\sizeU^s, \lambda_\sizeU^t)$, which indicates a spectral upper bound (bandwidth) for the frequencies of the first $\sizeU$ source and target Fourier basis vectors. Assume that the difference between the source and the target coefficients is bounded as $\| \ralpha^s - \ralpha^t \| \leq \dalpha$, and the deviation between $\rT$ and the $\sizeU \times \sizeU$ identity matrix $I$ is bounded as $\| \rT - I \| \leq \dT$, with $\| \cdot \|$ denoting the operator norm for matrices. Finally let $\calpha$ be a bound for the norms of the computed coefficients with $\| \ralpha^s \|, \| \ralpha^t \| \leq \calpha$. We then have the following result.

\begin{proposition}
\label{prop:diff_fs_ft_rates}
Assume that the constants $\lambdaR>0$, $\dlambda \geq 0$, $\dT \geq 0 $, $\dalpha \geq 0$, and $\calpha >0$ are such that the above conditions hold for the solution $\ralpha^s$, $\ralpha^t$ , $\rT$ of Problem 3. Then, the difference between the rates of variation of the estimated source and target label functions $f^s$, $f^t$ on the source and target graphs  is bounded as
\vspace{-0.5cm}
\begin{equation*}
\begin{split}
|  (f^s)^T L^s f^s - (f^t)^T L^t f^t | \leq  \calpha^2 \dlambda &+ 2 \calpha \lambdaR \dalpha \\
   &+ \calpha^2 \lambdaR ( 2 \dT + \dT^2 ).
\end{split}
\end{equation*}

\end{proposition}
%
The proof of Proposition \ref{prop:diff_fs_ft_rates} is given in Appendix \ref{app:pf:prop:diff_fs_ft_rates}. In the light of this theoretical bound, the formulation proposed in Problem 3 can be interpreted as follows. In the considered setting, due to the assumption of the similarity of their spectra, the source and target label functions must have similar rates of variation over the two graphs. The bound in Proposition \ref{prop:diff_fs_ft_rates} shows that the source and target label functions have similar rates of variation if the constants $\dlambda$, $\lambdaR$, $\dalpha$, $\dT$ are sufficiently small. The constant $\dlambda$ depends on the topological similarity between the two graphs and cannot be controlled by the learning algorithm. Meanwhile, the constant  $\lambdaR$ in the above bound suggests that preventing $\lambdaR$ from taking very large values should have a positive effect on the learning. This is in line with the choice of representing the label functions with a relatively small number $\sizeU$ of basis vectors in Problem 3, in contrast to Problem 2. Then, another objective of Problem 3 is to minimize the difference between the coefficient vectors $\ralpha^s$ and $\ralpha^t$, which reduces $\dalpha$. Finally, the term $\| \rW \odot \rT \|_F^2$ in the learning objective aiming to discourage large off-diagonal entries will eventually help reduce the constant $\dT$ in the above bound. Note, however, that we deliberately avoid imposing $\rT \approx I$ in Problem 3, which would restrict the flexibility of the learnt bases in aligning the two graphs to account for the differences in the graph topologies. This is discussed in more detail in Section \ref{ssec:overall_opt}.


\section{Proposed Method: Domain Adaptation via Spectral Graph Alignment}
\label{sec:prop_method}


In this section, we present the proposed domain adaptation method, which we call Domain Adaptation via Spectral Graph Alignment (DASGA). Our algorithm aims to learn a pair of ``aligned'' bases on the source and target graphs based on Problem 3.


The problem in \eqref{eq:prob_form_general} is not jointly convex in all optimization variables $\ralpha^s$, $\ralpha^t$, $\rT$. Nevertheless, it is convex separately in the overall coefficient vector $\ralpha= [ (\ralpha^s)^T  (\ralpha^t)^T ]^T$, and the transformation matrix $\rT$. Hence, we propose to minimize the objective \eqref{eq:prob_form_general} with an iterative and alternating optimization approach, by first fixing $\rT$ and optimizing $\ralpha^s$, $\ralpha^t$; and then fixing the coefficient vectors $\ralpha^s$, $\ralpha^t$ and optimizing $\rT$ in each iteration. We describe these two optimization steps in the sequel.

\subsection{Optimization of the Coefficient Vectors}
\label{ssec:opt_coef_vects}

In the first step of an iteration, the transformation matrix $\rT$ is fixed, and the coefficient vectors $\ralpha^s$ and $\ralpha^t$ are optimized. Fixing $\rT$, the optimization problem in \eqref{eq:prob_form_general} becomes the following unconstrained problem in $\ralpha^s$ and $\ralpha^t$

\begin{equation}
\label{eq:prob_general_optalpha}
\begin{split}
 \min_{\ralpha^s, \ralpha^t} G(\ralpha^s, \ralpha^t ) = 
& \min_{\ralpha^s, \ralpha^t}  \ \ 
  \| S^s \rU^s \ralpha^s - y^s  \|^2    \\
&+  \| S^t \rU^t \rT \ralpha^t - y^t  \|^2 
  + \mu_1 \, \|  \ralpha^s - \ralpha^t \|^2  .\\
 \end{split}
\end{equation}

The function $G(\ralpha^s, \ralpha^t )$ is convex in the coefficients $\ralpha^s$ and $ \ralpha^t $ and its global minimum can be found by setting its derivatives to 0: 

\begin{equation}
\begin{split}
\frac{\partial G(\ralpha^s, \ralpha^t ) }{\partial \ralpha^s} &= 2A^s \ralpha^s - 2 B^s y^s + 2 \mu_1 \ralpha^s - 2 \mu_1 \ralpha^t = 0 \\
\frac{\partial G(\ralpha^s, \ralpha^t ) }{\partial \ralpha^t}  &= 2A^t \ralpha^t - 2 B^t y^t + 2 \mu_1 \ralpha^t - 2 \mu_1 \ralpha^s = 0 \\
\end{split}
\end{equation}
where 
\begin{equation}
\begin{split}
A^s &= (\rU^s)^T (S^s)^T  S^s \rU^s,
\qquad
B^s = (\rU^s)^T (S^s)^T \\
A^t &= (\rU^t \rT)^T (S^t)^T S^t \rU^t \rT,
\qquad
B^t = (\rU^t \rT)^T (S^t)^T .
\end{split}
\end{equation}
This gives the coefficient vectors as 
\begin{equation}
\begin{split}
\ralpha^s &= ( \mu_1^{-1}  A^t A^s + A^t + A^s  )^{-1}  
		(    \mu_1^{-1}  A^t  B^s y^s +   B^s y^s + B^t y^t   ) \\
\ralpha^t &=  (	\mu_1^{-1} A^s \ralpha^s + \ralpha^s 	 - \mu_1^{-1}  B^s y^s  ).
\end{split}
\end{equation}

\subsection{Optimization of the Transformation Matrix}
\label{ssec:opt_trans_mat}

In the second step of an iteration, the coefficient vectors $\ralpha^s$ and $\ralpha^t$ are fixed and the transformation matrix $\rT$ is optimized. Then the minimization of the objective in \eqref{eq:prob_form_general} becomes equivalent to the following problem 
\begin{equation}
\label{eq:prob_general_opt_T2}
\begin{split}
 \min_{\rT} H( \rT ) = 
 &\min_{\rT}  \ \  \| S^t \rU^t \rT \ralpha^t - y^t  \|^2 + \mu_2 \, \|  \rW \odot \rT \|_F^2  \\
 &  \text{subject to } \sum_{i=1}^{\sizeU} \rT_{ij}^2 = 1 , \text{ for } \ j = 1,\ \dots,  \sizeU. \\
 \end{split}
\end{equation}

The above problem involves the minimization of a quadratic convex function $H(\rT)$ in $\rT$ subject to $\sizeU$ equality constraints that are also quadratic and convex in $\rT$. We solve the problem in \eqref{eq:prob_general_opt_T2} using the Sequential Quadratic Programming (SQP) algorithm \cite{NocedalW06}, which is a method to numerically solve constrained nonlinear optimization problems. The SQP algorithm is based on iteratively approximating the original problem with a Quadratic Programming problem, where the objective function is replaced with its local quadratic approximation, and the equality and inequality constraints are replaced with their local affine approximations. In our problem \eqref{eq:prob_general_opt_T2}, the objective function $H(\rT)$ is already a quadratic function of $\rT$ and we only have equality constraints.

The first and second order derivatives to be used in the solution of \eqref{eq:prob_general_opt_T2} are found as follows. Let $\rt \in \R^{\sizeU^2}$ denote the column-wise vectorized form of the matrix $\rT$, such that its $k$-th entry is given by $\rt_k = T_{ij}$, with $k=(j-1)\sizeU + i$, for $i,j=1, \dots, \sizeU$. We denote by $h(\rt) = H(\rT)$ the objective in \eqref{eq:prob_general_opt_T2} when considered as a function of $\rt$. The objective function $h(\rt)=H(\rT)$ can then be rewritten in terms of $\rt$ as
\begin{equation}
\label{eq:form_h_t}
h(\rt) = \| A \rt - y^t \|^2 + \mu_2  \|  F \rt \|^2.
\end{equation}
Here $A \in \R^{\Lt \times \sizeU^2}$ is a matrix with entries given by $A_{lk}=(S^t \rU^t)_{li} \, \ralpha^t_j$ and $F \in \R^{\sizeU^2 \times \sizeU^2}$ is a diagonal matrix with entries given by
$
F_{kk}=\rW_{ij}, 
$
where $l=1, \dots, \Lt$ and  $k= \sizeU(j - 1) + i$,  for $i, j=1, \dots, \sizeU$. The variable $\Lt$ here is the number of labeled target samples. Next, the $j$-th equality constraint of the problem \eqref{eq:prob_general_opt_T2} can be written in terms of $\rt$ as
\begin{equation}
\label{eq:form_jth_const}
g_j(\rt) = \sum_{i=1}^{\sizeU} \rT_{ij}^2 - 1 = 0
\end{equation}
for $j=1, \dots, \sizeU$. 

The problem \eqref{eq:prob_general_opt_T2} is then solved by forming the Lagrangian function
\begin{equation}
\L(\rt, \eta)= h(\rt) - g(\rt, \eta)
\end{equation}
where
\begin{equation}
\label{eq:Lagr_constraint_term}
g(\rt, \eta) =  \sum_{j=1}^\sizeU \eta_j g_j(\rt),
\end{equation}
$\eta_j>0$ are the Lagrange multipliers, and $\eta = [\eta_1 \dots \eta_\sizeU]^T$. From \eqref{eq:form_h_t}, we obtain the gradient of the objective $h(\rt)$ as
\begin{equation}
\label{eq:grad_h_v2}
\nabla_{\rt} h = 2 (A^T A  +  \mu_2 F^T F ) \rt
\end{equation}
and its Hessian as
\begin{equation}
\label{eq:form_hessian_h}
\nabla^2_{\rt \rt} h(\rt) = 2 (A^T A + \mu_2 F^T F).
\end{equation}
Next, from \eqref{eq:form_jth_const}, the $k$-th entry of the gradient of $g_j(\rt)$ is found as
\begin{equation}
\label{eq:constraint_gradient}
\left (  \nabla_{\rt}  \, g_j \right)_k=
  \begin{cases} 
 2 \rt_k  \, , & \text{if } (j-1) \sizeU + 1 \leq k \leq  j \sizeU \\
 0, & \text{otherwise}
   \end{cases}
\end{equation}
for $k=1, \dots, \sizeU^2$. From \eqref{eq:constraint_gradient}, the Hessian $\nabla^2_{\rt \rt} \, g(\rt, \eta)$ of the second term $g(\rt, \eta)$ of the Lagrangian in \eqref{eq:Lagr_constraint_term} is obtained as a diagonal matrix with entries given by
\begin{equation}
\label{eq:hess_constraints}
[ \nabla^2_{\rt \rt} \, g(\rt, \eta) ]_{kk}= 2 \eta_j
\end{equation}
for $\sizeU(j - 1) + 1 \leq k \leq \sizeU j$. Putting \eqref{eq:form_hessian_h} and \eqref{eq:hess_constraints} together, we get the Hessian of the Lagrangian as
\begin{equation}
\nabla^2_{\rt \rt}  \L(\rt, \eta)= \nabla^2_{\rt \rt}  h(\rt) - \nabla^2_{\rt \rt}  g(\rt, \eta).
\end{equation}

The SQP algorithm optimizes objectives with equality constraints by iteratively updating the solution $(\rt, \eta)$, where a linear system representing the approximate solution of the KKT conditions with the Newton's method is solved in each iteration \cite[Algorithm 18.1]{NocedalW06}. The linear system is constructed from the objective $h(\rt)$, the constraints $g_j(\rt)$, their gradients, and the Hessian of the Lagrangian.

\textbf{Remark:} Although the SQP algorithm often converges to a solution in practice, it is not  easy to establish a general theoretical convergence guarantee. For our problem, the convergence can be theoretically guaranteed under certain conditions: Let the algorithm parameter $\mu_2$ be chosen such that there exists a local solution $(\rt^*, \eta^*)$ to the Lagrangian function $\L(\rt, \eta)$ of the constrained problem \eqref{eq:prob_general_opt_T2} such that $\mu_2> \eta_j^*$ for all $j=1, \dots, \sizeU$. Then, if the initialization $(\rt, \eta)$ of the SQP algorithm is sufficiently close to $(\rt^*, \eta^*)$, the algorithm converges to $(\rt^*, \eta^*)$. The details of this convergence analysis are provided in Appendix \ref{app:conv_anly_sqp}.

\subsection{Overall Optimization Procedure}
\label{ssec:overall_opt}

We now overview the overall optimization procedure employed in the proposed DASGA method. First, the optimization variables $\rT$, $\ralpha^s$, and $\ralpha^t$ are initialized  as follows. Since the objective in Problem 3 aims to find a transformation that aligns the source and target Fourier bases, a natural choice would be to initialize $\rT$ as the identity matrix, so that each source vector $u_i^s$ is mapped to the target vector $u_i^t$. However, even in a simple scenario where the source and target graphs are very similar, as the eigenvalue decomposition determines eigenvectors up to a sign, mapping each $u_i^s$ to $u_i^t$ might in fact constitute a bad initialization; e.g., consider the very simple case where the source and target graphs are identical but $u_i^t = - u_i^s$. An unfavorable initialization of the transformation matrix may consequently influence the estimates of the coefficient vectors $\ralpha^s$, $\ralpha^t$ and affect the overall solution of the alternating optimization procedure.

In order to obtain a more favorable initialization, we propose to set the initial $\rT$ matrix with a strategy that corrects the sign of each target vector according to its best match among the source basis vectors. This strategy is based on the method presented in our previous work \cite{PilanciV16}, where the best match of a target vector $u_i^t$ among the source vectors is determined by finding

\begin{equation}
\label{eq:restrict_us_ut}
\max_j  |  \langle \tilde u^s_j , \tilde u^t_i  \rangle |.
\end{equation}

Here $\tilde u^s_j$, $\tilde u^t_i $ are subvectors of the basis vectors $u^s_j$, $ u^t_i $ obtained by restricting them to a subset of their entries indexed by some $\{ s_i \}_{i=1}^K$ and $\{ t_i \}_{i=1}^K$. It is difficult to directly compare the vectors $u^s_j$, $ u^t_i $ as the nodes of the source and target graphs are ordered arbitrarily and independently of each other. If a set of corresponding source and target node pairs $\mathcal{N} = \{ (x^s_{s_i}, x^t_{t_i}) \}_{i=1}^K $ is known, then this set can be used for the restriction of the basis vectors to a subset of their entries in the problem \eqref{eq:restrict_us_ut}, so that the vectors $u^s_j$, $ u^t_i $ can be compared throughout their chosen entries. However, in our method we do not rely on the availability of a set of corresponding node pairs and propose to form the set $\mathcal{N} = \{ (x^s_{s_i}, x^t_{t_i}) \}_{i=1}^K $ based on the class labels, such that each pair of matched nodes $(x^s_{s_i}, x^t_{t_i}) $ is formed randomly among the source and target nodes having the same class labels. We then compare the vectors $u^s_j$, $ u^t_i $ over their entries $\tilde u^s_j$, $\tilde u^t_i $ corresponding to these nodes.  Although very few labeled target nodes are typically available in a domain adaptation application, we have observed that only a few pairs is often sufficient to determine the correct signs for initializating $\rT$, which is next done as follows
\begin{equation}
\label{eq:init_T}
\rT_{ii}= \sgn( \langle \tilde u^s_{J_i} , \tilde u^t_i \rangle ), \qquad 
J_i = \arg \max_j  |  \langle \tilde u^s_j , \tilde u^t_i  \rangle |.
\end{equation}
Here $\sgn$ denotes the sign function and $\rT$ is initialized as a diagonal matrix with $-1$'s or $1$'s on the diagonals that matches the sign of each target vector $u^t_i$ to the source vector $u^s_j$ best corresponding to it. This initialization respects the normalization constraint \eqref{eq:prob_form_general} on the entries of $\rT$.

Once the transformation matrix $\rT$ is initialized in this way, the alternating optimization procedure starts, where the coefficient vectors $\ralpha^s$ and $\ralpha^t$ are computed by fixing $\rT$ first, and then $\rT$ is optimized by fixing  $\ralpha^s$ and $\ralpha^t$ in each iteration, as described in Sections \ref{ssec:opt_coef_vects} and \ref{ssec:opt_trans_mat}. In each iteration, both the updates on $\ralpha^s$ and $\ralpha^t$, and the update on $\rT$ either reduce or retain the value of the objective function in \eqref{eq:prob_form_general}. Since the objective function is nonnegative and thus bounded from below, it converges throughout the proposed iterative alternating optimization process. We continue the iterations until the convergence of the objective function. The proposed Domain Adaptation via Spectral Graph Alignment (DASGA) algorithm is summarized in Algorithm \ref{alg:SDA_Algorithm}.

\begin{algorithm}[t]
\caption{Domain Adaptation via Spectral Graph Alignment (DASGA)}

\begin{algorithmic}[1]

\STATE
\textbf{Input:} \\
$W^s$, $W^t$: Source and target graph weight matrices\\
$y^s$, $y^t$: Available source and target labels\\

\STATE
\textbf{Initialization:}

Set the transformation matrix $\rT$ as  in  \eqref{eq:init_T}.

\REPEAT

\STATE
\label{alg:state_coef_upd}
 Update coefficients $\ralpha^s$, $\ralpha^t$ by solving \eqref{eq:prob_general_optalpha}.
 
\STATE
\label{alg:state_trans_upd}
Update transformation matrix $\rT$ by solving \eqref{eq:prob_general_opt_T2}.

\UNTIL the objective function  \eqref{eq:prob_form_general} converges

\STATE
\textbf{Output}:\\

$f^t =  \rU^t \rT  \ralpha^t $: Estimated target label function\\
$f^s = \rU^s  \ralpha^s$: Estimated source label function

\end{algorithmic}
\label{alg:SDA_Algorithm}
\end{algorithm}

\subsection{Complexity Analysis}

We now present the complexity analysis of the proposed method. The overall complexity is mainly determined by the complexity of Steps \ref{alg:state_coef_upd} and \ref{alg:state_trans_upd} of Algorithm \ref{alg:SDA_Algorithm} executed iteratively until convergence. Let $\Ls$ and $\Lt$ denote the number of labeled samples respectively in the source and the target domains.

We first derive the complexity of Step \ref{alg:state_coef_upd}. In the solution of \eqref{eq:prob_general_optalpha}, the matrices $B^s$ and $A^s$ are respectively computed with $O(\Ls \Ns \sizeU)$ and $O(\Ls \Ns \sizeU + \Ls \sizeU^2)$ operations. Meanwhile,  these are constant matrices that do not depend on $\rT$ and they are computed only once; hence, we may ignore their calculation in the overall complexity. Next, $O(\Nt \sizeU^2 + \Lt \Nt \sizeU)$ and $O(\Nt \sizeU^2 + \Lt \Nt \sizeU + \Lt \sizeU^2)$ operations are needed to compute the matrices $B^t$ and $A^t$ respectively. The matrices $ \mu_1^{-1}  A^t A^s + A^t + A^s $ and $  \mu_1^{-1}  A^t  B^s y^s +   B^s y^s + B^t y^t  $ in the expression of $\ralpha^s$ are computed respectively with $O(\sizeU^3)$  and $O(\Ls \sizeU^2 + \Lt \sizeU )$ operations. Considering also the matrix inversion in its expression, $\ralpha^s$ is computed with $O(\sizeU^3)$ operations. The target coefficients $\ralpha^t$ are then obtained from $\ralpha^s$ with $O(\sizeU^2)$ operations. From the complexities of all these computations, we get the overall complexity of Step \ref{alg:state_coef_upd} of Algorithm \ref{alg:SDA_Algorithm} as $O(\sizeU^3 + (\Ls + \Nt) \sizeU^2 +  \Lt \Nt \sizeU)$.

Next, we examine the complexity of executing Step \ref{alg:state_trans_upd} with the SQP algorithm. The complexity of the evaluation of $h(\rt)$ in \eqref{eq:form_h_t} is of $O(\Lt \sizeU^2 + \sizeU^4)$. From \eqref{eq:grad_h_v2}, we observe that the gradient $\nabla_{\rt} h$ is computed with $O(\sizeU^4)$ operations as well. Finally, since the Hessian $\nabla^2_{\rt \rt} h(\rt) $ of the objective in \eqref{eq:form_hessian_h} is a constant matrix that does not depend on $\rt$, we can exclude it from the complexity of the iterative SQP algorithm. Next, from \eqref{eq:form_jth_const}, the complexity of computing all $\sizeU$ gradients is obtained as $O(\sizeU^2)$. From \eqref{eq:constraint_gradient} and \eqref{eq:hess_constraints}, we observe that the gradients $\nabla_{\rt} g_j(\rt)$ of the constraints and the Hessian $\nabla^2_{\rt \rt} \, g(\rt, \eta)$ are obtained directly from $\rt$ and $\eta$ without any operations. We thus conclude that the Hessian  $\nabla_{\rt \rt}^2 \L(\rt, \eta)$ of the Lagrangian can also be obtained with negligible complexity. Finally, the optimization variables are updated by solving the linear system given in \cite[Algorithm 18.1]{NocedalW06} with $O(\sizeU^6)$ operations in a single iteration of the SQP algorithm. Putting together the complexities of all these operations, we conclude that the complexity of solving Step \ref{alg:state_trans_upd} with the SQP algorithm is of $O(\sizeU^6 + \Lt \sizeU^2  )$.

Finally, considering together the Steps \ref{alg:state_coef_upd} and \ref{alg:state_trans_upd} of Algorithm \ref{alg:SDA_Algorithm}, we get the overall complexity of the DASGA algorithm as $ O(\sizeU^6 + (\Ls + \Nt) \sizeU^2 +  \Lt \Nt \sizeU)$.


\section{Experimental results}
\label{sec:exp_res}

In the following, we first introduce the datasets and then evaluate the performance of the proposed method with comparative experiments. Next, we study the behavior of the algorithm throughout the iterative optimization procedure and examine its sensitivity to the choice of the algorithm parameters.

\subsection{Data sets}
\label{ssec:data_sets}

\begin{figure}[t]
\begin{center}
     \subfigure[Synthetic dataset-1]
       {\label{fig:toydata_mix1}\includegraphics[height=3cm]{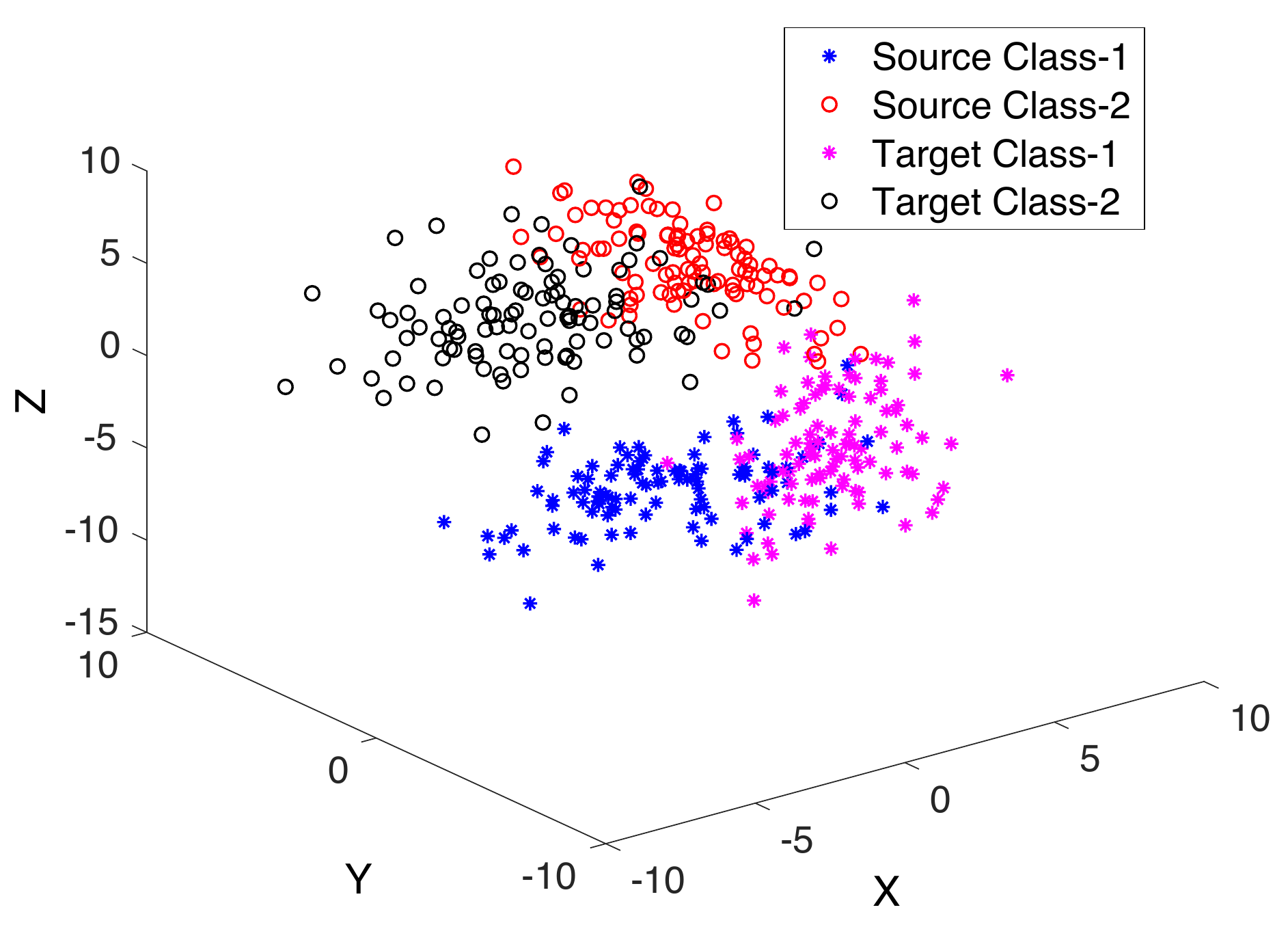}}
       \subfigure[Synthetic dataset-2]
       {\label{fig:toydata_mix3}\includegraphics[height=3cm]{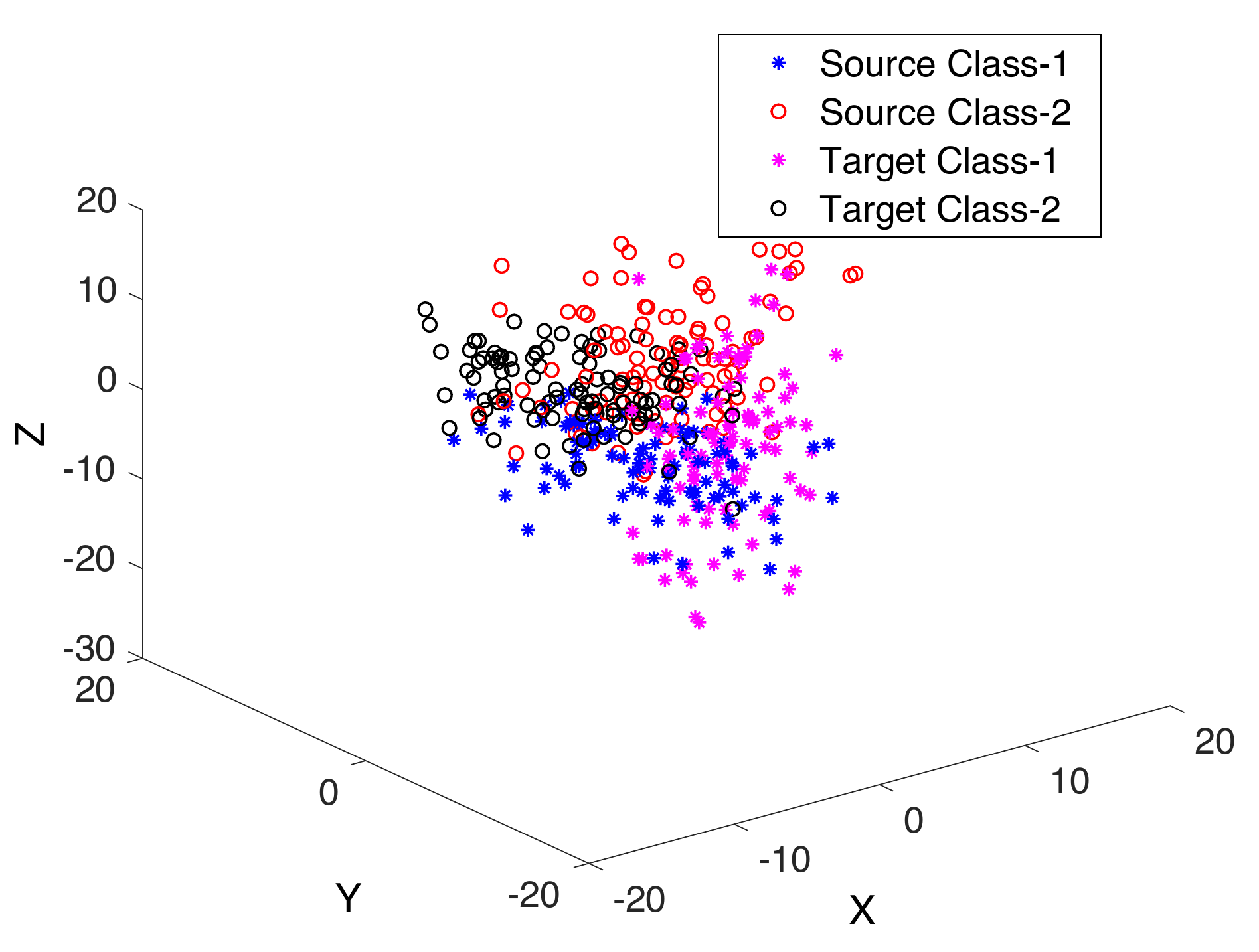}}  
 \end{center}
 \caption{Synthetic data sets with two classes}
 \label{fig:toydata}
\end{figure}

\begin{figure}[t]
  \centering
  \centerline{\includegraphics[width=7.0cm]{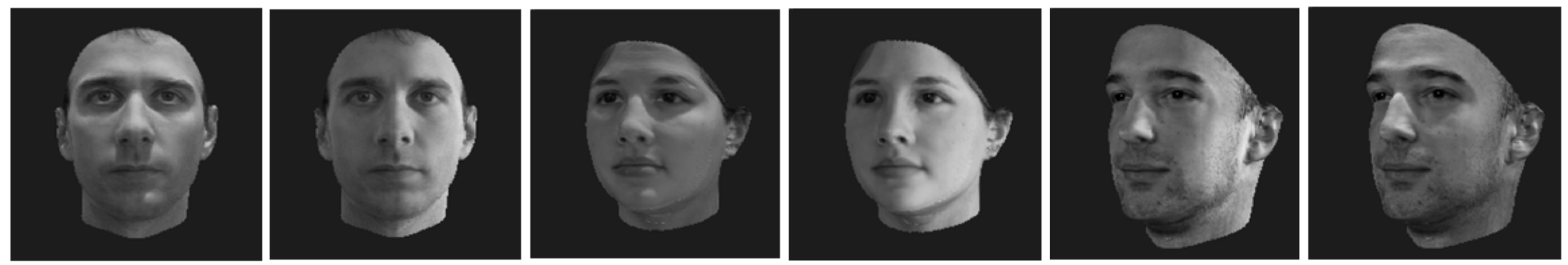}}
  \caption{Sample images from the MIT-CBCL face data set for three different subjects \cite{MITCBCL}. Leftmost two, middle two, and rightmost two images are rendered respectively under poses 1, 5, and 9 for various illumination conditions. }
  \label{fig:face_dataset}
\end{figure}

\begin{figure}[t]
  \centering
  \centerline{\includegraphics[width=8cm]{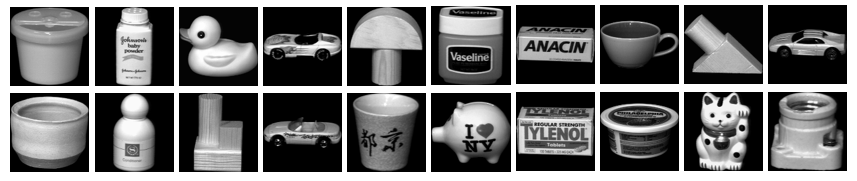}}
 \caption{Sample images from the COIL-20 data set. The upper and lower rows show the objects respectively in the source domain and the target domain. Each source domain object is matched to the target domain object right below it. Matched object pairs are considered to have the same class label in the experiments.}
  \vspace{-0cm}
  \label{fig:Coil_data_matches}
\end{figure}

\begin{figure}[t]
  \centering
  \centerline{\includegraphics[width=6.5cm]{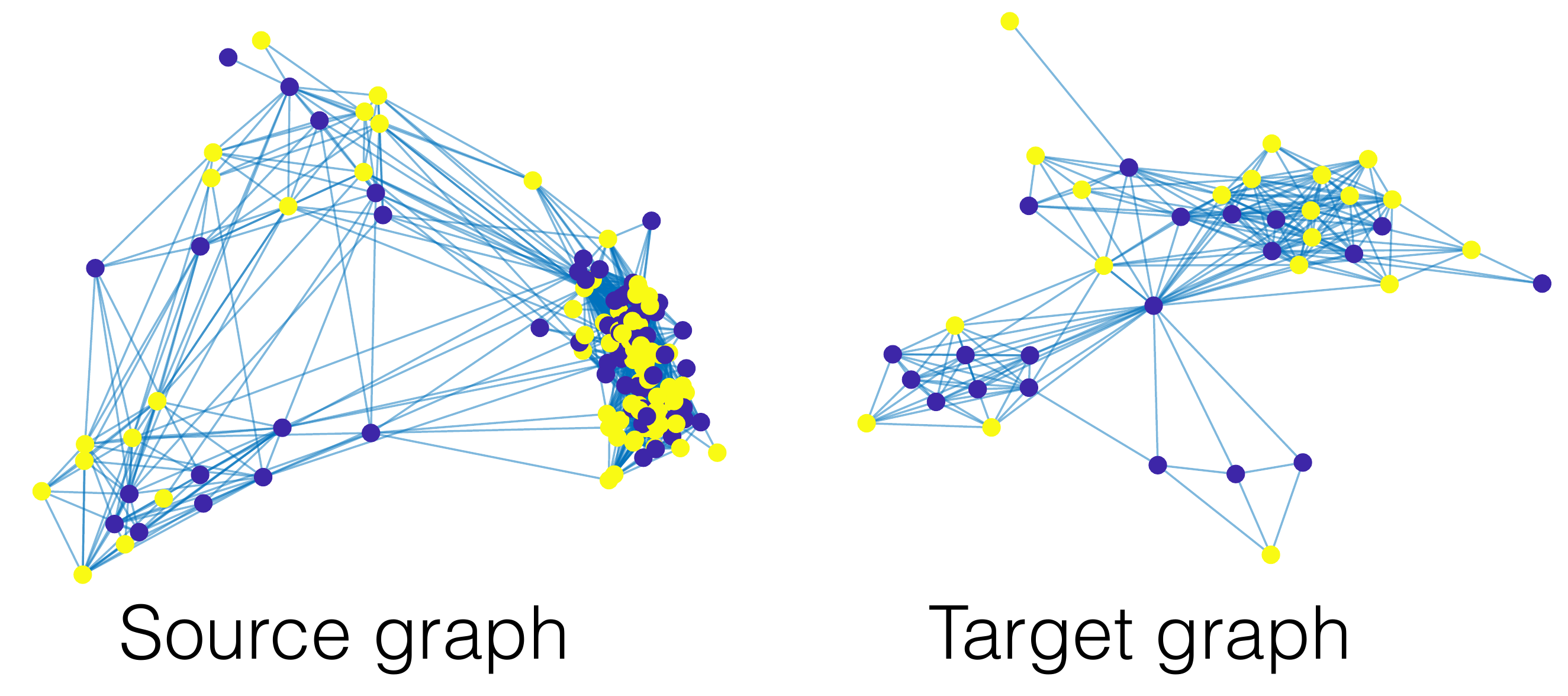}}
 \caption{Source and target community graphs for the Facebook data}
  \vspace{-0cm}
  \label{fig:Facebook_graphs}
\end{figure}

The following data sets are used in the experiments.

\textit{Synthetic data sets.} The two synthetic data sets shown in Figures \ref{fig:toydata_mix1} and \ref{fig:toydata_mix3} are generated by drawing 100 samples for each class from a normal distribution in $\R^3$, with different means for the two classes. The means of the source classes and the corresponding target classes are symmetric along the $y$-direction. The variance of the distribution is higher in Synthetic dataset-2 compared to Synthetic dataset-1; hence the difficulty level of the classification task is higher. The source and the target graphs are constructed by connecting each data sample to their $25$ nearest neighbors.  The edge weights are computed with a Gaussian kernel as $w^s_{ij}=\exp{(-\| x^s_i - x^s_j \|^2/\sigma^2)}$ in the source graph and similarly in the target graph, where $x^s_i$ and $x^s_j$ are the data sample coordinates and the scale parameter $\sigma$ is chosen proportionally to the typical distance between neighboring samples.

\textit{MIT-CBCL face image data set.}  The MIT-CBCL face recognition database \cite{MITCBCL} consists of a total of 3240 face images rendered from the 3D head models of 10 participants under varying illumination and poses. The images of each participant are rendered under 9 different poses varying from the frontal view (Pose 1) to a nearly profile view (Pose 9), and 36 illumination conditions at each pose. Some sample images are shown in Figure \ref{fig:face_dataset}. We downsample the images to a resolution of $100 \times 100$ pixels and consider the images taken under each pose as samples from a different domain. Raw features consisting of pixel intensities are used in the experiments. Two settings are considered, where the source domain is taken as Pose 1 in both settings. The target domain is taken as Pose 5 in the first setting, and as Pose 9 in the second setting.  Source and target data graphs are constructed independently in the source and the target domains, by connecting each image to its nearest $38$ neighbors with respect to the Euclidean distance. The weight matrices $W^s$, $W^t$ are constructed with a Gaussian kernel as in the synthetic data sets.

\textit{COIL object image data set.} The COIL-20 object database \cite{NeneNM96} consists of a total of 1440 images of 20 objects. Each object has 72 images taken from different viewpoints rotating around it. We downsample the images to a resolution of $32 \times 32$ pixels. We consider a transfer learning setting by dividing the 20 objects in the data set into two groups and matching each object in the first group to another object in the second group with respect to their similarity computed via pairwise distances. The experiments are then done by considering each group of 10 objects as a different domain, and regarding the images of the matched objects across the two domains as having the same class label. The two groups and the matched object pairs are shown in Figure \ref{fig:Coil_data_matches}. The source and the target graphs are constructed by connecting each sample to its $3$ nearest neighbors and the weights are set with a Gaussian kernel. A small number of neighbors is chosen deliberately to be coherent with the small intrinsic dimension of the data set as the images are formed by rotating the camera around each object in only one direction.

\textit{Amazon product ratings data set.} The Amazon data set \cite{AmazonMcAuley} is used in the task of predicting user ratings on books. The data set contains scores from users who purchased a book from Amazon, where the scores are integers in the range $[1, 5]$. The experiment is conducted on the first $150000$ ratings in the data set. The users who rated less than three books are excluded from the experiment. In each repetition of the experiment, two bestsellers are chosen from the book catalogue of Amazon. The source graph consists of the users who read the first bestseller, and the target graph consists of the users who read the second bestseller. Each graph node represents a user, and the scores that the users gave to the first and the second bestsellers are regarded as signals (label functions), respectively on the source and the target graphs. The source and the target graphs are constructed with respect to the similarities between the users, where two users are considered similar if their past reading records agree. Thus, if two users have read books in common, they are connected with an edge in the graphs. The edge weights are determined as inversely proportional to the average difference of the scores the users assigned to the same books, in order to capture the similarity of their literary preferences. Given the scores on the source bestseller, and the available scores on the target bestseller, we consider the task of predicting the unavailable scores on the target bestseller.

\begin{figure}[t]
\begin{center}
       \subfigure[Synthetic dataset-1]
       {\label{fig:Fourier_coef_Toydata1}\includegraphics[height=2.2cm]{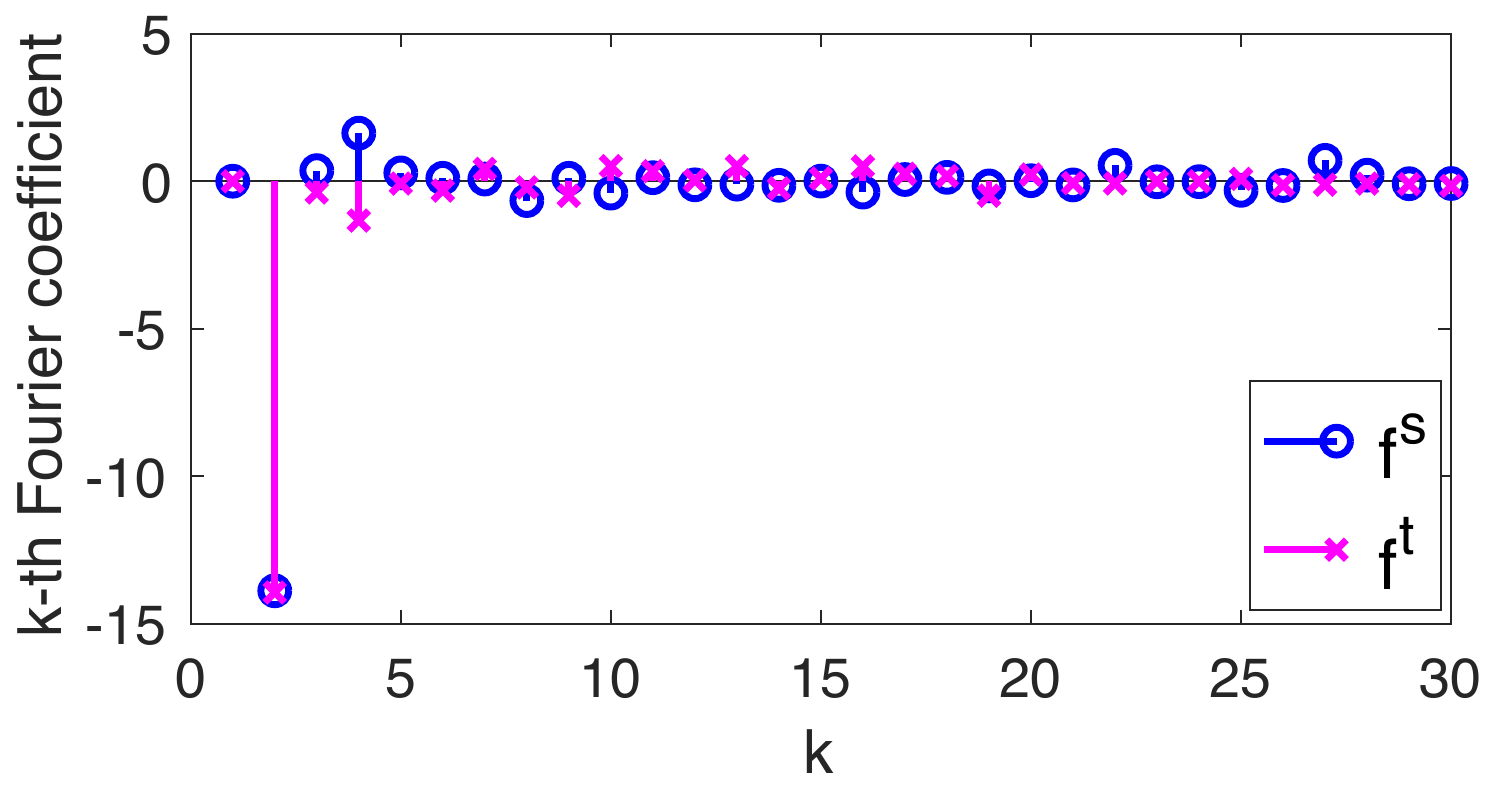}}  
      \subfigure[Synthetic dataset-2]
       {\label{fig:Fourier_coef_Toydata3}\includegraphics[height=2.2cm]{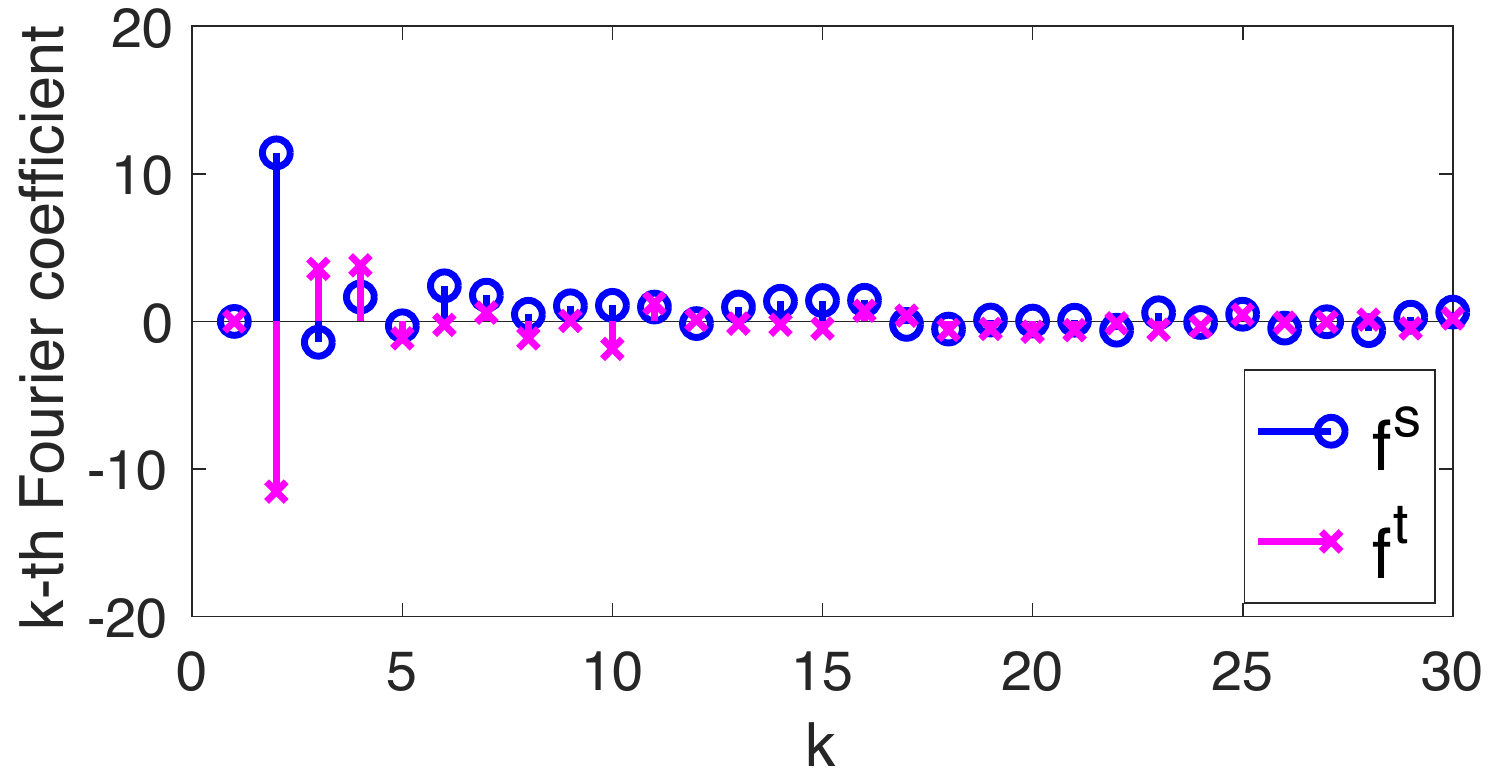}}   
 \end{center}
 \caption{Source and target label spectra on synthetic data sets}
  \label{fig:toydata_spectra}
 \end{figure}

\begin{figure}[t]
\begin{center}
       \subfigure[Source: Pose 1, Target: Pose 5]
       {\label{fig:Fourier_coef_MITCBCL_1_5}\includegraphics[height=2.8cm]{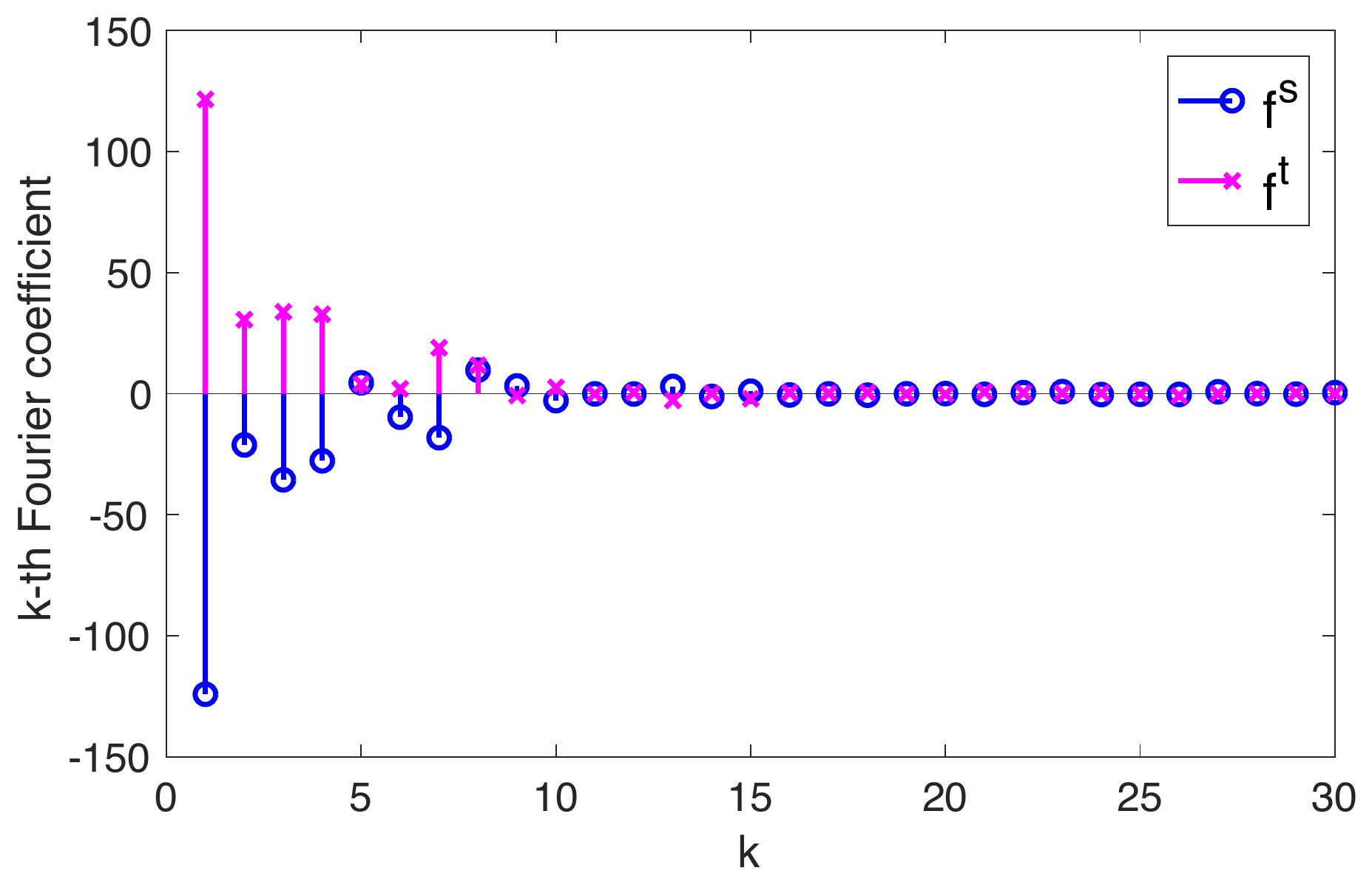}}  
      \subfigure[Source: Pose 1, Target: Pose 9]
       {\label{fig:Fourier_coef_MITCBCL_1_9}\includegraphics[height=2.8cm]{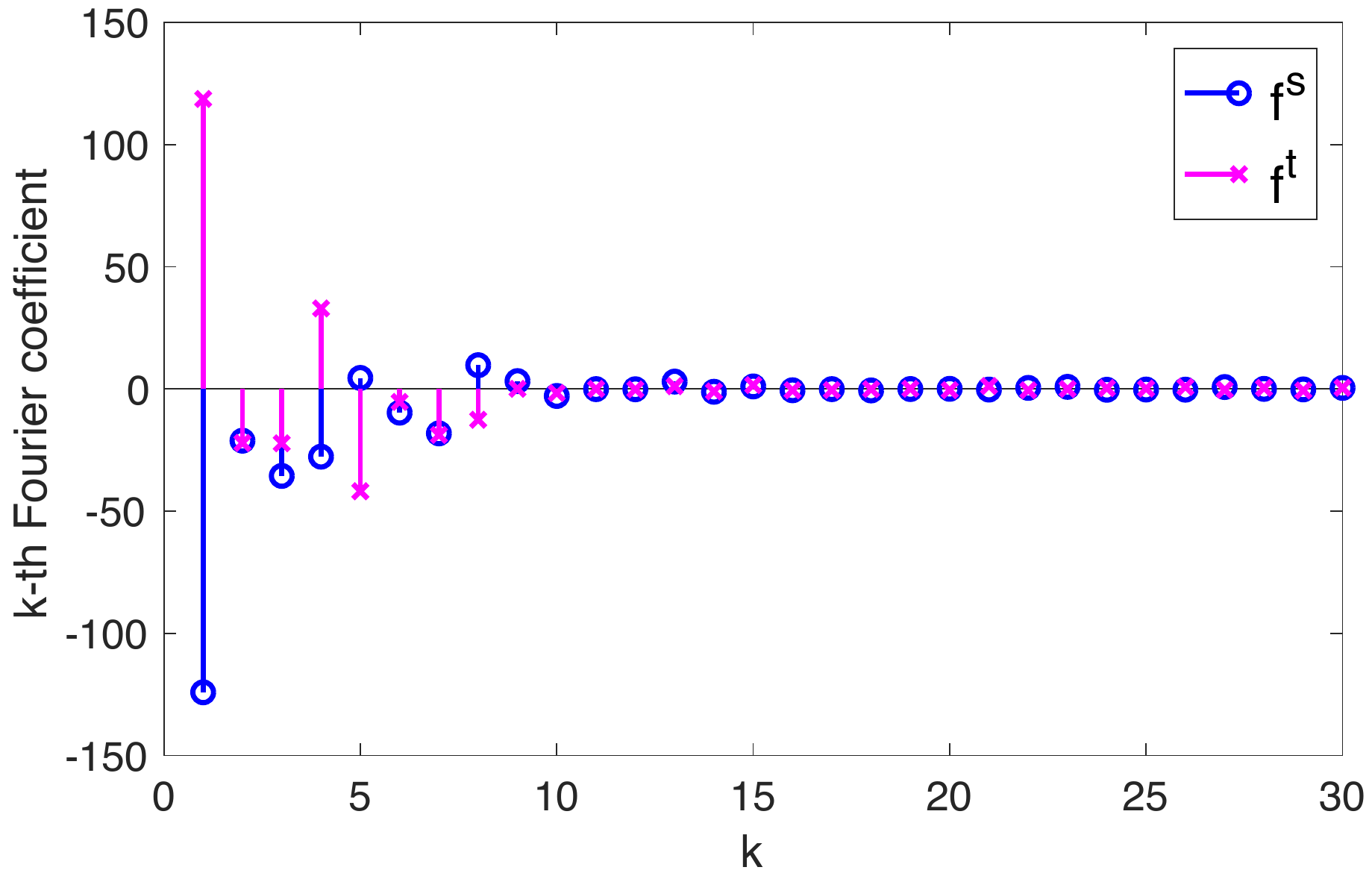}}   
 \end{center}
 \caption{Source and target label spectra on the MIT-CBCL data set}
  \label{fig:mitcbcl_spectra}
 \end{figure}

\begin{figure}[t]
  \centering
  \centerline{\includegraphics[width=5.0cm]{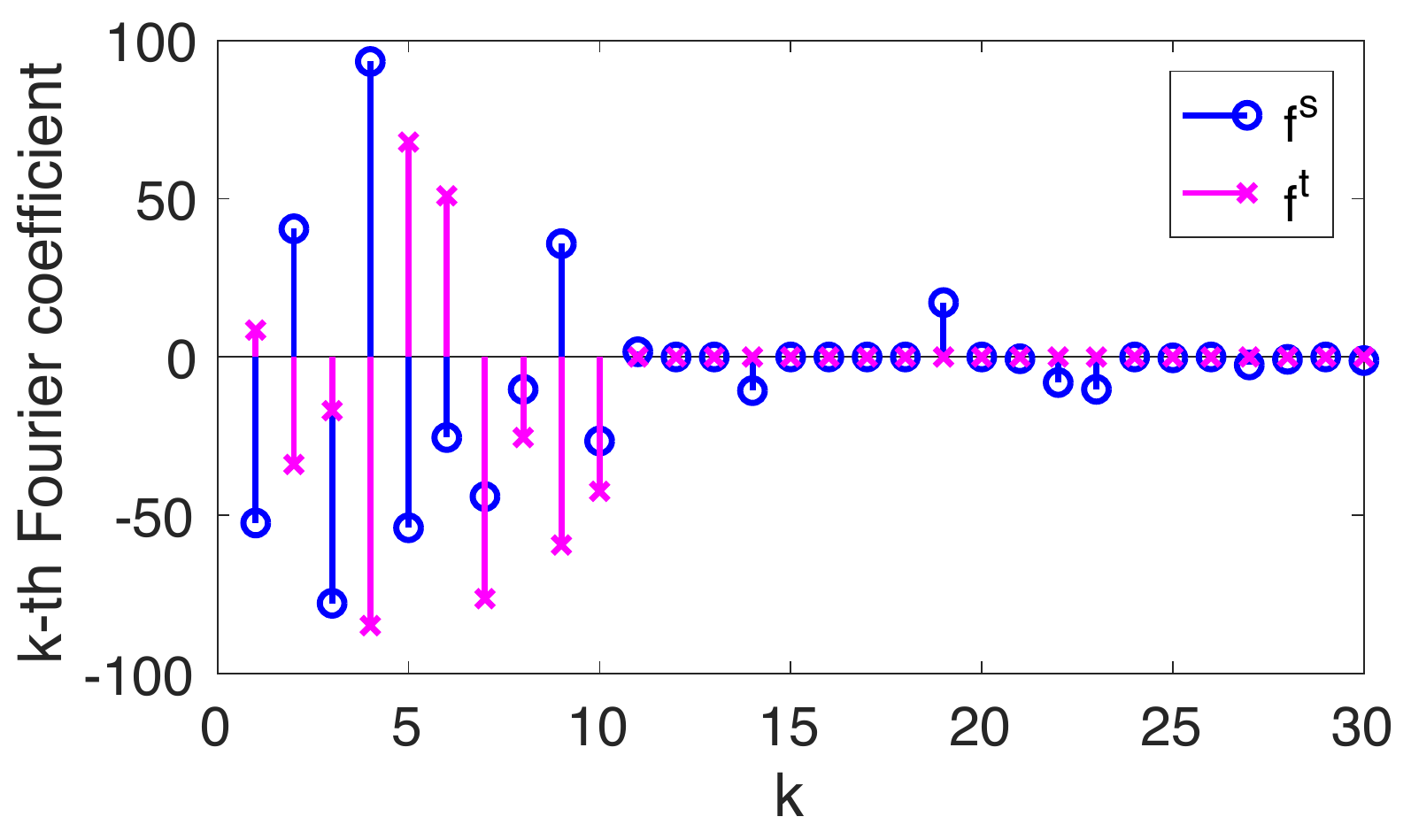}}
 \caption{Source and target label spectra on the COIL-20 data set}
  \vspace{-0cm}
  \label{fig:coil_spectra}
\end{figure}

\begin{figure}[t]
  \centering
  \centerline{\includegraphics[width=5.0cm]{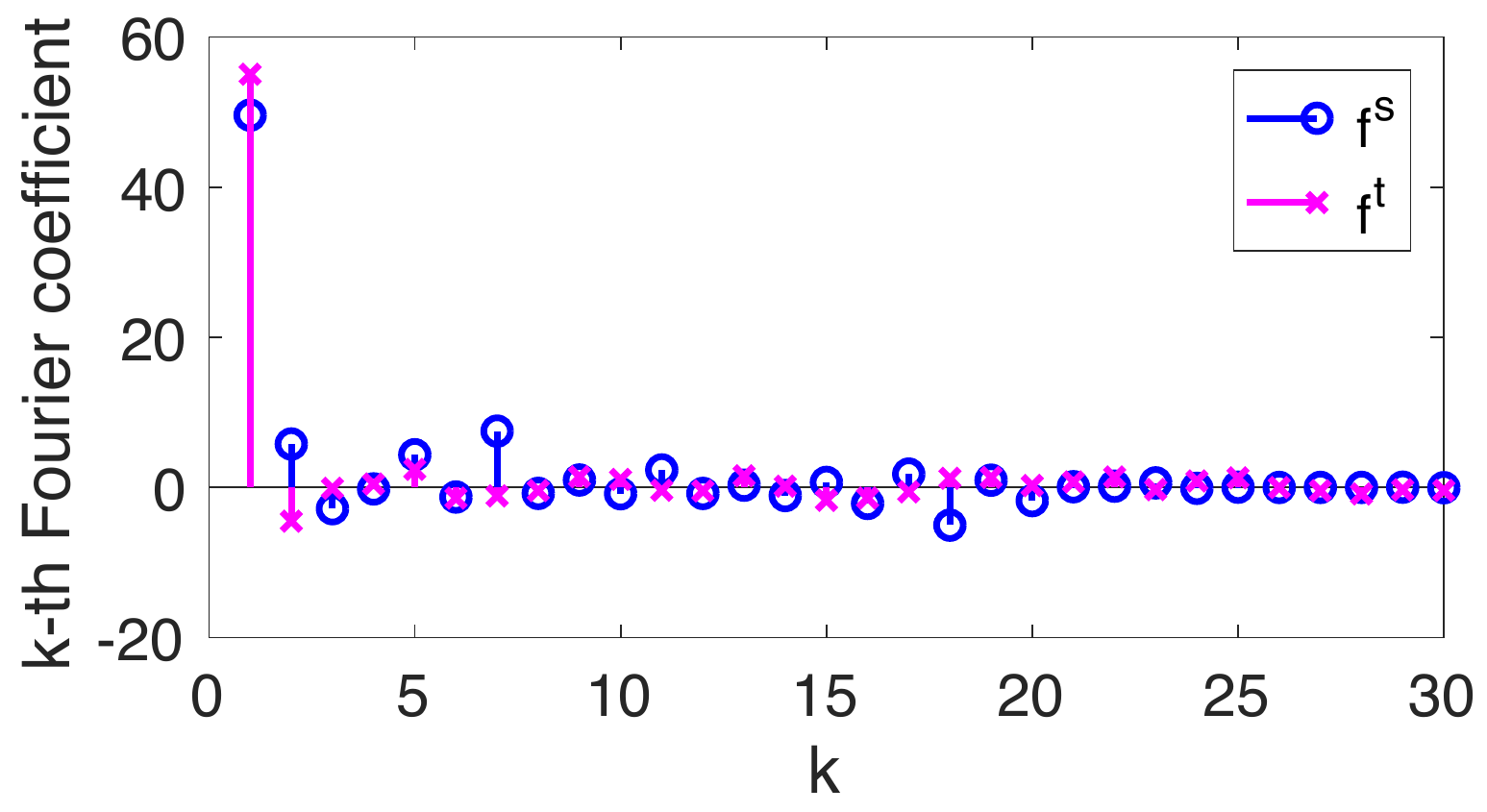}}
 \caption{Source and target label spectra on the Amazon data set}
  \vspace{-0cm}
  \label{fig:amazon_spectra}
\end{figure}

\begin{figure}[h!]
  \centering
  \centerline{\includegraphics[width=5.0cm]{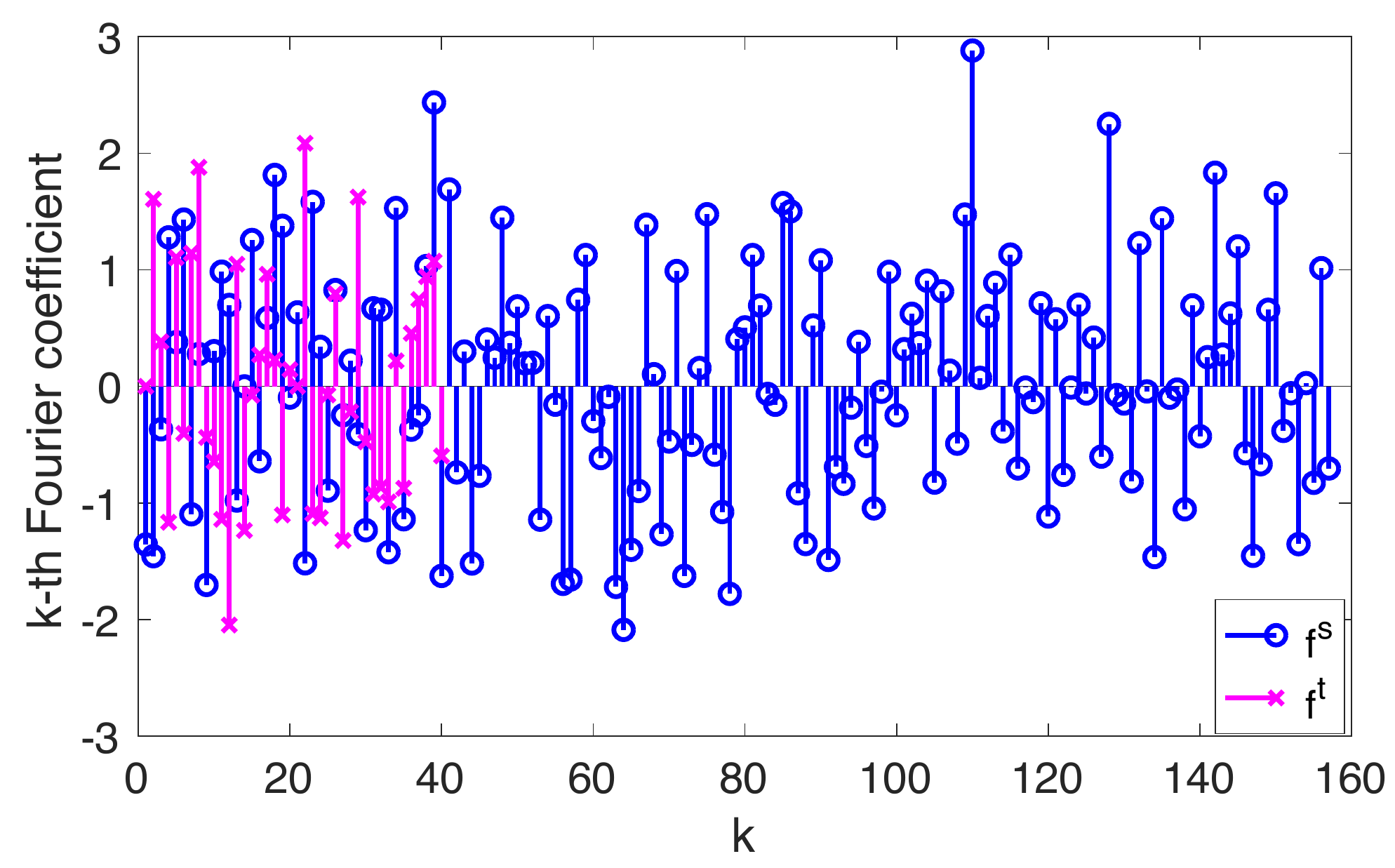}}
 \caption{Source and target label spectra on the Facebook data set}
  \vspace{-0cm}
  \label{fig:facebook_spectra}
\end{figure}

\textit{Facebook data set.} The Facebook data set \cite{McAuleyL12} consists of various communities (friend circles) extracted from the Facebook network. Graph nodes and edges respectively represent Facebook users and their friendship relations. In our experiments two different communities are chosen as the source graph and the target graph. Isolated users or user cliques are removed and the weights of all edges are set to the constant value 1. The gender of the Facebook users is taken as the binary label function to be predicted. The source and the target graphs, consisting respectively of 157 and 40 users, are shown in Figure \ref{fig:Facebook_graphs}, where the values of the label function are represented with two different colors.\\

In the following, we first verify the validity of the main assumption of the proposed method that the frequency content of the label function is similar on the source and the target graphs. The Fourier coefficients of the source and the target label functions are plotted in Figures \ref{fig:toydata_spectra}-\ref{fig:facebook_spectra} for all data sets. Heavily concentrated at low frequencies, the source and the target label functions seem to have similar frequency contents for the two synthetic data sets in Figure \ref{fig:toydata_spectra}. In Figure \ref{fig:Fourier_coef_MITCBCL_1_5}, the source and the target Fourier coefficients at the same frequency have quite similar magnitudes due to the high similarity between the source and the target images captured under nearby camera angles. On the other hand, in Figure \ref{fig:Fourier_coef_MITCBCL_1_9} where the two domains bear smaller resemblance, the source and the target Fourier coefficients at the same frequency do not always have similar magnitudes. Nevertheless, the shape of the spectrum is similar between the source and the target graphs, with similar amplitudes at nearby frequencies. The results on the COIL-20 and Amazon data sets in Figures \ref{fig:coil_spectra} and \ref{fig:amazon_spectra} can be interpreted similarly, as the spectra of the source and the target label functions decay with the frequency and have similar characteristics in both data sets. The label function is seen to have a rather flat spectrum in Figure \ref{fig:coil_spectra} for the Facebook data set, on both the source and the target graphs (note that the two graphs have different sizes and all Fourier coefficients are plotted in the figure). This is due to the highly irregular nature of the label function, which is observable in Figure \ref{fig:Facebook_graphs}. The experiments on all of the data sets lead to the common conclusion that the assumption that the label function has similar frequency content on the source and the target graphs is realistic in practice.


\subsection{Evaluation of the Algorithm Performance }
\label{ssec:exp_alg_comp}

The performance of the proposed DASGA method is compared to the domain adaptation methods Heterogeneous Domain Adaptation using Manifold Alignment (DAMA) \cite{WangM11}, Easy Adapt++ (EA++) \cite{Daume2010}, Subspace Alignment (SA) \cite{Fernando2013}, Geodesic Flow Kernel for Unsupervised Domain Adaptation (GFK) \cite{GongSSG12},  Scatter Component Analysis (SCA) \cite{GhifaryBKZ17}, LDA-Inspired Domain Adaptation (LDADA) \cite{LuSC0H18}, Joint Geometrical and Statistical Alignment (JGSA) \cite{ZhangLO17};  as well as the baseline classifiers Support Vector Machine (SVM), Nearest-Neighbor classification (NN), and the graph-based Semi-Supervised Learning with Gaussian  fields (SSL) algorithm \cite{ZhuGL03}. The baseline classifiers SVM and NN are evaluated under the ``source+target'' setting using the labeled samples from both the source and the target domains for training, and the SSL algorithm is used in the ``target only'' setting, which give the best results. When testing the SA and GFK algorithms, once the source and the target domains are aligned in an unsupervised way as proposed in \cite{Fernando2013} and \cite{GongSSG12}, the known source and target labels are both used in the final classification. In all experiments the source labels are assumed to be known and the ratio of known target labels is varied gradually. The class labels of the unlabeled target samples are then estimated with the tested methods and the classification performances are compared.

\begin{figure}[t]
\begin{center}
     \subfigure[Synthetic dataset-1]
       {\label{fig:toydata_mix1_errors}\includegraphics[height=2.7cm]{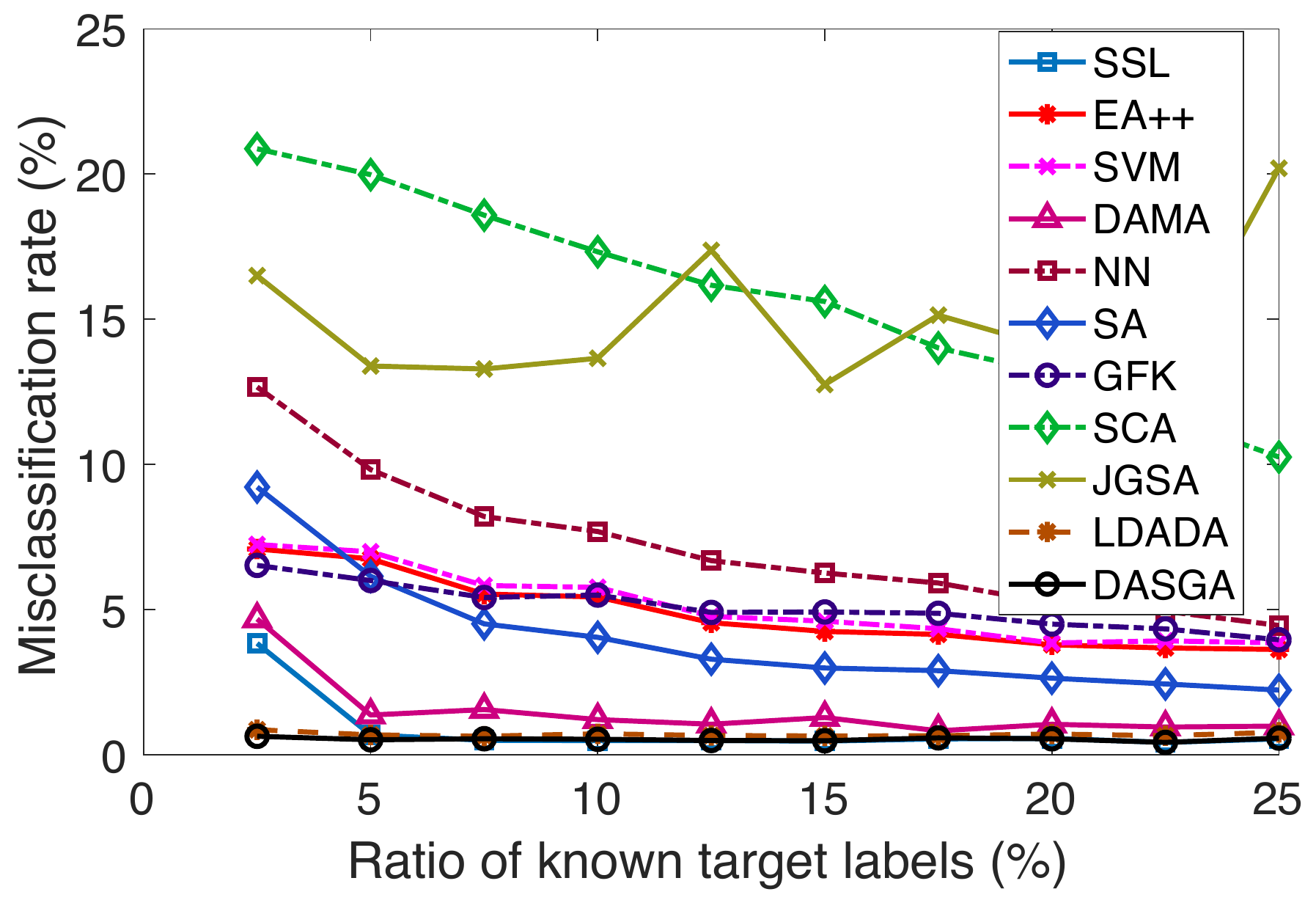}}
        \subfigure[Synthetic dataset-2]
       {\label{fig:toydata_mix3_errors}\includegraphics[height=2.7cm]{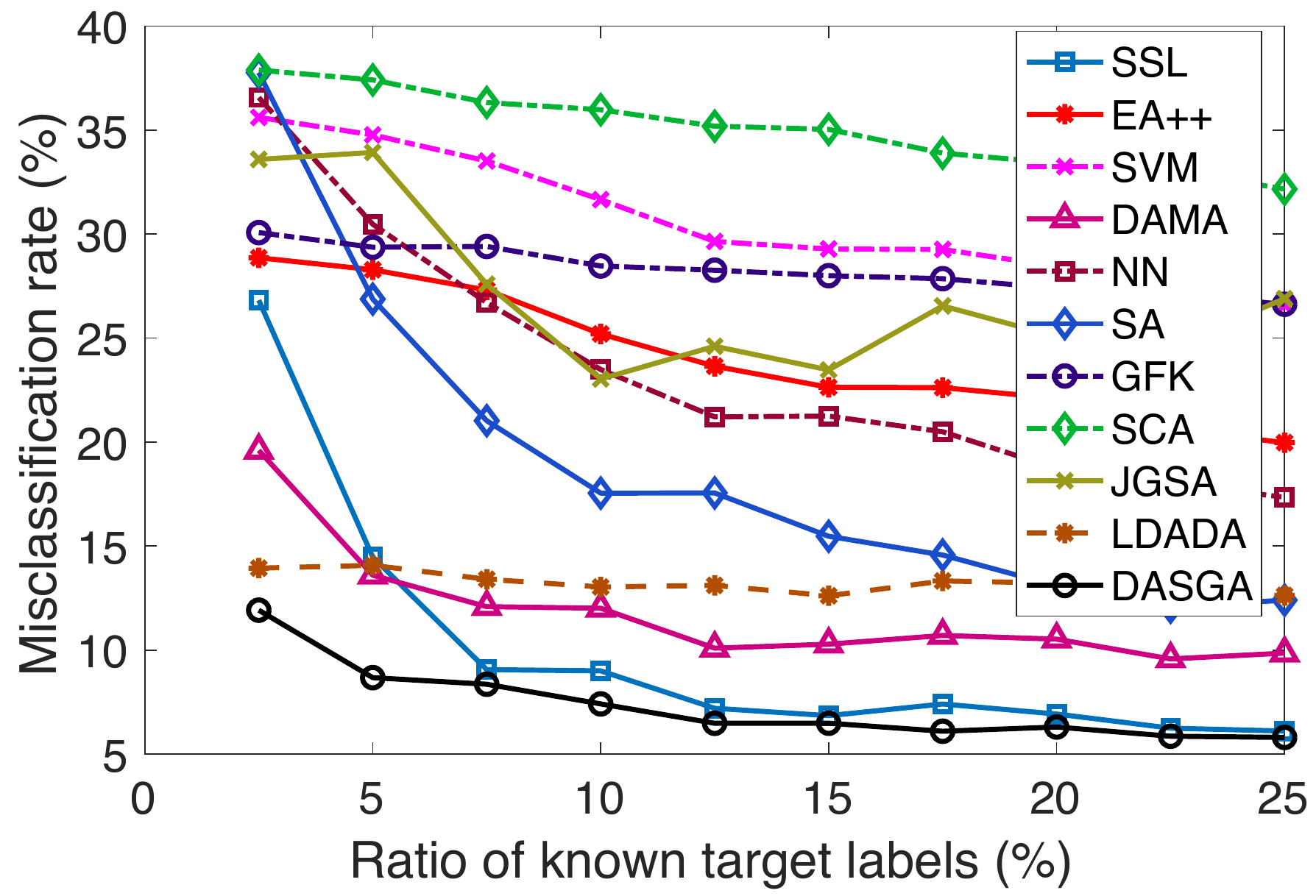}}  
 \end{center}
 \caption{Misclassification rates of target samples for synthetic data sets}
 \label{fig:toydata_errors}
\end{figure}

\subsubsection{Experiments on synthetic data sets}
\label{ssec:exp_syn_data}

The proposed DASGA algorithm is used with the parameters $\mu_1=0.1$, $\mu_2 = 1$, $\sizeU = 9$ in the experiments with synthetic data sets. In Figure \ref{fig:toydata_errors}, the misclassification rates of unlabeled target samples in percentage are plotted with respect to the ratio of labeled target samples in percentage. The results are averaged over 50 repetitions of the experiment with random selections of the labeled samples. As expected, the misclassification rates of the algorithms tend to decrease as the ratio of known target labels increases. The proposed DASGA algorithm is observed to outperform the compared methods in both data sets.  The performance gap between DASGA and the other methods is larger in Synthetic dataset-2, which is a more challenging data set due to the relatively high distribution variance. 
Among the domain adaptation methods, DAMA \cite{WangM11} and LDADA \cite{LuSC0H18} give the closest performance to the proposed DASGA method. The approach in both of these methods is to learn supervised projections, which is relatively successful in this synthetic data set consisting of normally distributed data. On the other hand, the proposed DASGA method relies on a pure graph representation of data, therefore its performance is less affected by the  ambient space properties of the data. This feature is seen to provide some robustness against the challenges such as large variance and poor separation between the classes.

\subsubsection{Experiments on image data sets}
\label{ssec:exp_image_data}

\begin{figure}[t]
\begin{center}
     \subfigure[Source: Pose 1, Target: Pose 5]
       {\label{fig:errors_1_5}\includegraphics[height=4.0cm]{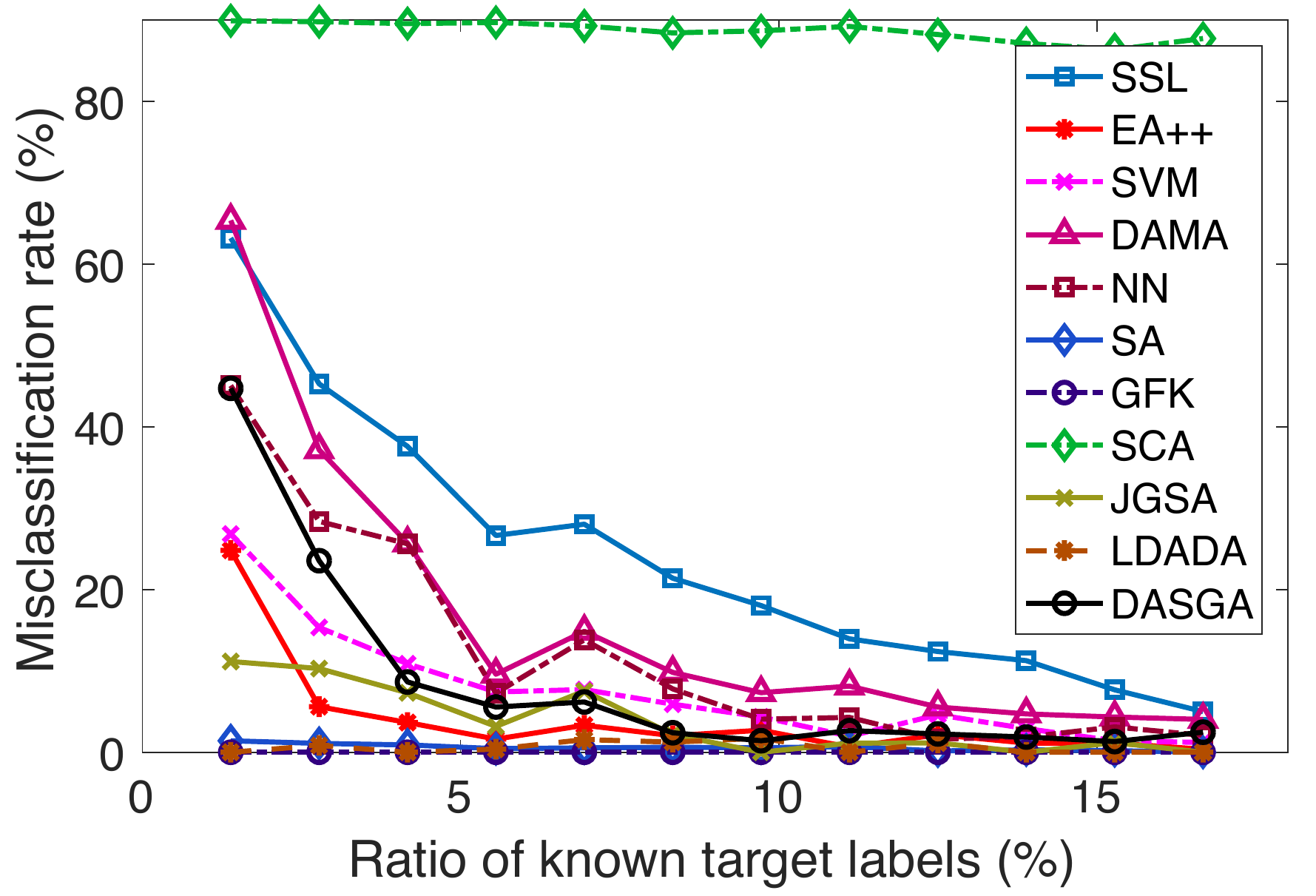}}
      \subfigure[Source: Pose 1, Target: Pose 9]
       {\label{fig:errors_1_9}\includegraphics[height=4.0cm]{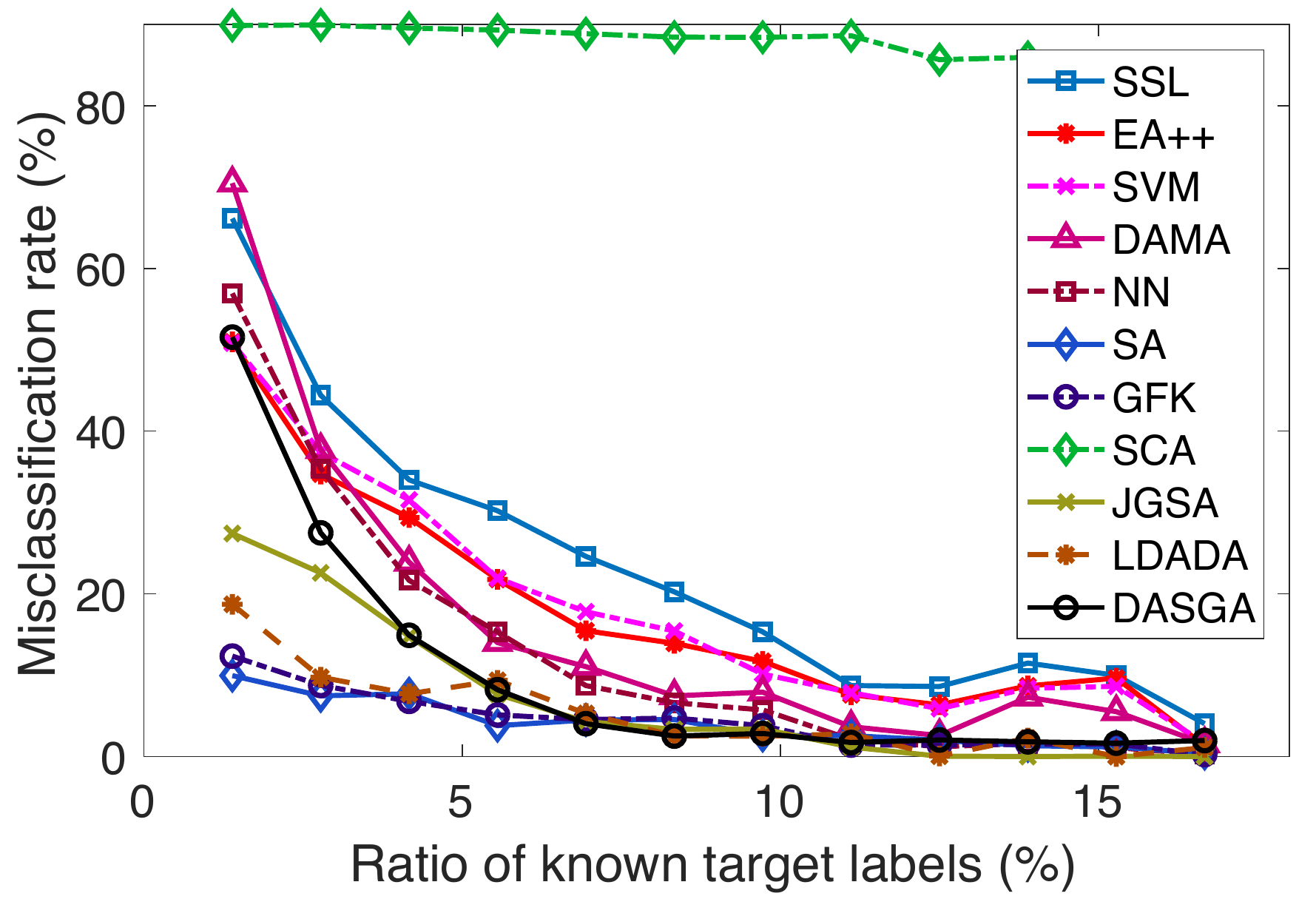}}
 \end{center}
 \caption{Misclassification rates of target samples for the MIT-CBCL data}
 \label{fig:mitcbcl_errors}
\end{figure}

We next evaluate the performance of the proposed algorithm on the image data sets.  In the experiments with the MIT-CBCL face image data set, the parameters of the proposed DASGA method are set as $\mu_1=0.1$, $\mu_2=0.85$, and $\sizeU=9$.  The experiment is repeated over $10$ realizations with random selections of the labeled samples and the results are averaged. The misclassification rates of the unlabeled target images are plotted with respect to the ratio of labeled target images in Figures  \ref{fig:errors_1_5} and \ref{fig:errors_1_9}, where the target domain is respectively taken as Pose 5 and Pose 9. The misclassification errors of the algorithms are seen to be larger in Figure \ref{fig:errors_1_9} compared to Figure \ref{fig:errors_1_5}, as the source and the target poses differ more significantly and the similarity between the two domains is weaker. In Figure \ref{fig:errors_1_5}, where the source and the target domains are relatively similar, the proposed DASGA method is seen to be outperformed by the domain adaptation methods EA++ and JGSA,  as well as the  SA, GFK and LDADA methods which yield almost zero error. Capturing the face images of the same participants from nearby poses in a clean and controlled environment, the two domains in this experiment are quite convenient to align with methods using projections and geometric transformations, which explains the success of these algorithms. On the other hand, the proposed graph-based DASGA algorithm does not use the pixel intensity values of image data samples once the source and target graphs are constructed, hence, it does not employ the same type of information as these methods. Nevertheless, in Figure \ref{fig:errors_1_9}, where the source and target images differ more significantly, the performance of DASGA catches up with the other methods when the ratio of known target labels reaches $7\%$.

\begin{figure}[t]
  \centering
  \centerline{\includegraphics[width=7.0cm]{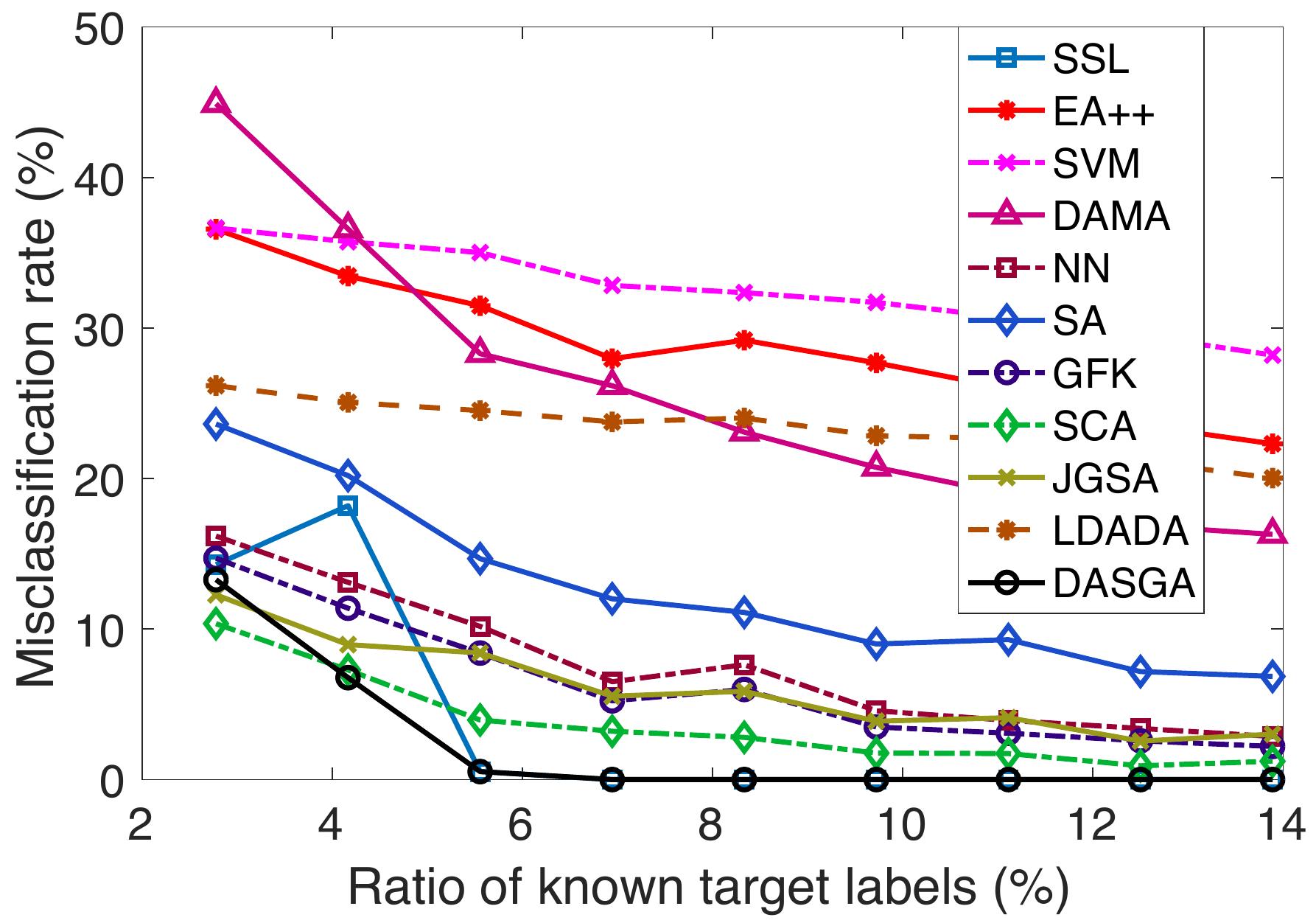}}
 \caption{Misclassification rates of target samples for the COIL-20 data }
  \vspace{-0cm}
  \label{fig:Coil_errors}
\end{figure}

In the experiments with the COIL-20 image data set, the parameters of the proposed method are set as $\mu_1=1$,  $\mu_2 = 1$ and $\sizeU=10$. The misclassification rates of the algorithms are plotted with respect to the ratio of known target labels in Figure \ref{fig:Coil_errors}. The proposed DASGA method is observed to often yield the best classification performance. The misclassification rate of the proposed algorithm falls to zero when about $ 7\% $ of the samples are labeled in the target domain. The graph-based semi-supervised learning algorithm SSL  also performs well in this experiment. The regular sampling of the images on the image manifold in this data set allows the construction of well-organized graphs, which can be successfully exploited by graph-based learning methods. The performances of the domain adaptation methods SA, DAMA, and LDADA fall behind that of the simple NN classifier in this experiment. Relying on the alignment of the source and the target domains via transformations, these methods fail in the transfer learning problem considered in this experiment. The source and the target images belong to different objects; hence, they are difficult to align via linear projections or transformations. It is also interesting to note that the relatively more sophisticated SCA method based on nonlinear kernel transformations, is more successful in this challenging data set compared to the previous data sets of simpler structure.

\subsubsection{Experiments on the Amazon book ratings data set}

In the experiments with the Amazon book ratings data, the parameters of DASGA are set as $\mu_1=0.001$, $\mu_2=0.8$,  and $\sizeU=10$, which are selected by trials on a test setup with two arbitrarily chosen bestsellers that are not used in the experiments. Being a purely graph-based method, the proposed DASGA algorithm requires only the source and the target user graphs and the available ratings. Meanwhile, the other algorithms in comparison require the coordinates of the data samples; thus, need an embedding of the data in an ambient space. Unlike the image data and the synthetic data used in the previous experiments, the data samples do not have a physical embedding in this experiment. One could possibly regard the user ratings given to previously read books as feature vectors. However, due to the very large number of books in the Amazon catalogue and the  small number of books users typically read, such feature vectors are very sparse in a very high-dimensional ambient space. This increases the complexity and impairs the performance and feasibility of most of the compared methods. Another solution could be to represent graph nodes using graph-theoretic features as in \cite{WattsS98},  \cite{GuptaLF16}; however, such features should be selected and used carefully. In order to test the compared methods, we follow an alternative approach and embed the source and the target graphs into an Euclidean domain of optimal dimension using the Multidimensional Scaling (MDS) algorithm \cite{Torgerson1952}. The coordinates learnt for each user with MDS are then used as training features by the compared algorithms.

\begin{figure}[t]
\begin{center}
      \subfigure[]
       {\label{fig:Amazon_Percent_Error}\includegraphics[height=4.0cm]{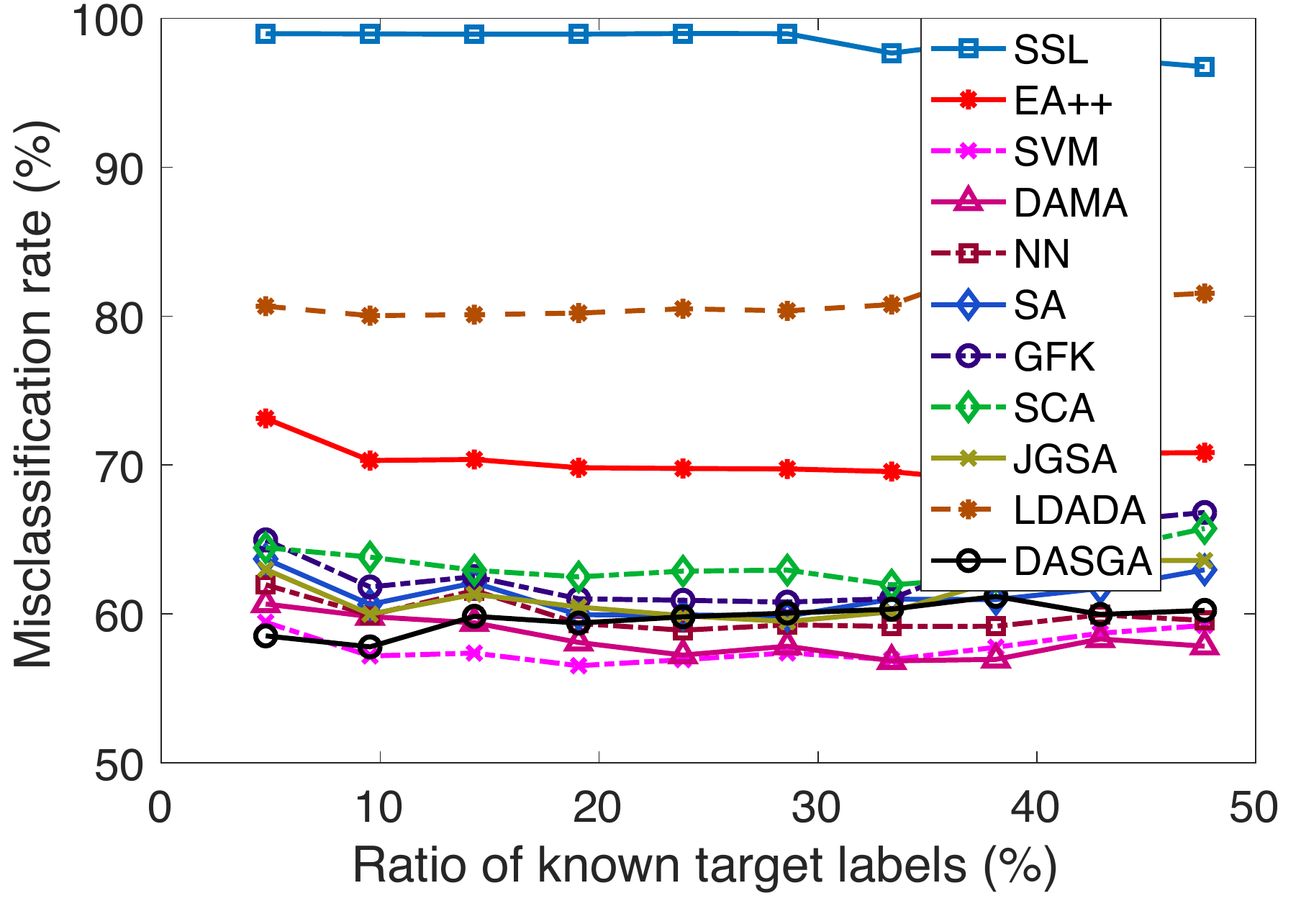}}
            \subfigure[]
       {\label{fig:Amazon_RMS_Error}\includegraphics[height=4.0cm]{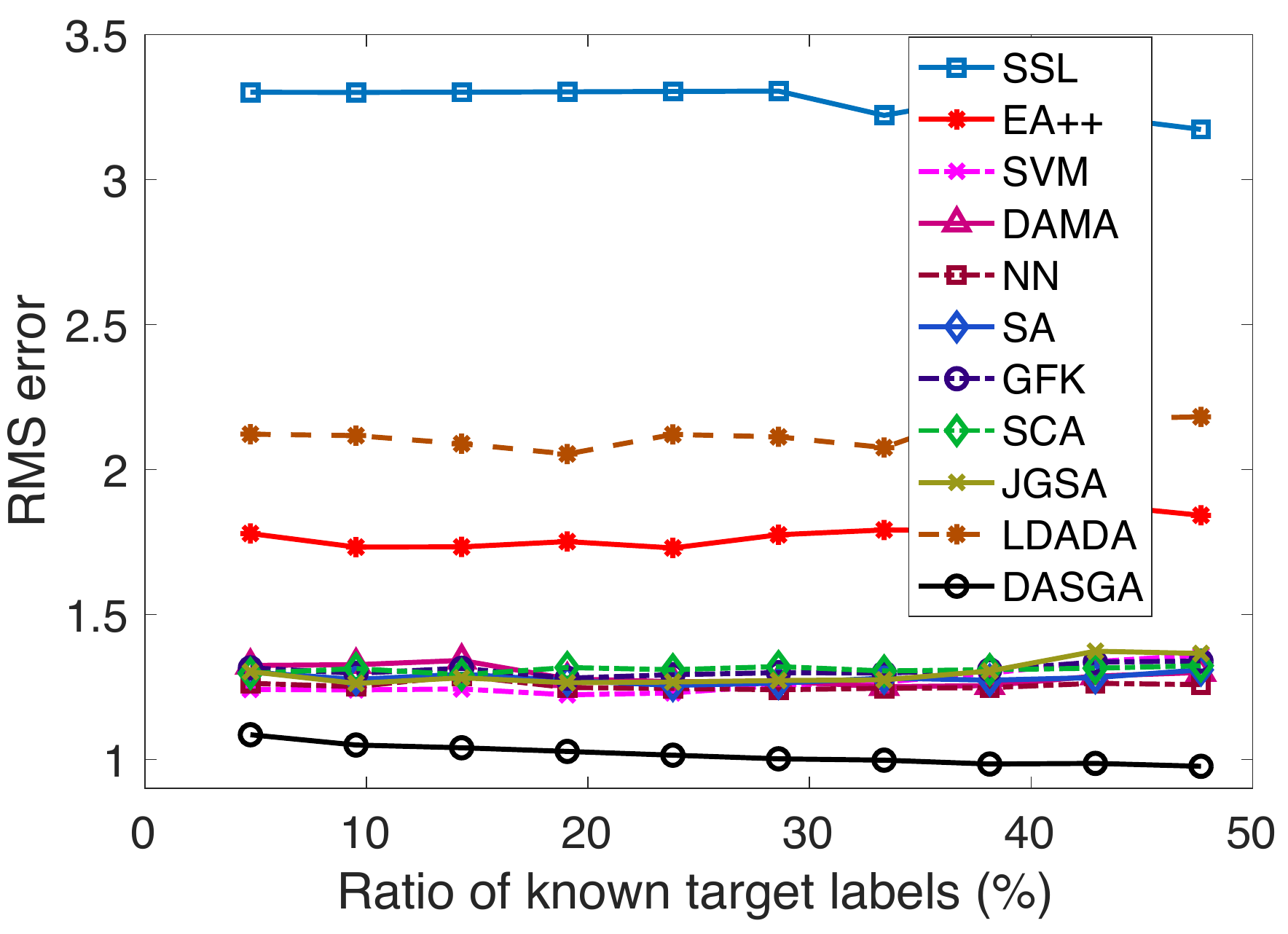}}
 \end{center}
  \vspace{-0.3cm}
 \caption{RMS errors and misclassification rates of target user score predictions for Amazon book ratings}
 \label{fig:Amazon_rms_misclas_errors}
\end{figure}

The experiment is conducted over 10 different pairs of source and target bestsellers, with 10 repetitions of the experiment for each bestseller pair by randomly selecting the labeled nodes. The average misclassification rates of the score predictions (considering each score from 1 to 5 as a different class label) are plotted in Figure \ref{fig:Amazon_Percent_Error}, and Figure \ref{fig:Amazon_RMS_Error} shows the root mean square (RMS) error of the predictions. The results in Figure \ref{fig:Amazon_Percent_Error} show that most of the methods including DASGA yield similar misclassification errors. Although DASGA does not provide smaller misclassification error than the other methods, Figure \ref{fig:Amazon_RMS_Error} shows that it clearly outperforms the other methods in terms of the RMS prediction error. The ensemble of the results in Figure \ref{fig:Amazon_rms_misclas_errors} suggests that the proposed DASGA algorithm is well-fit to the regression problem inherent to this setting as it relies on the analysis of the rate of variation of the user rating functions over the graphs.

\begin{figure}[t]
  \centering
  \centerline{\includegraphics[width=5.0cm]{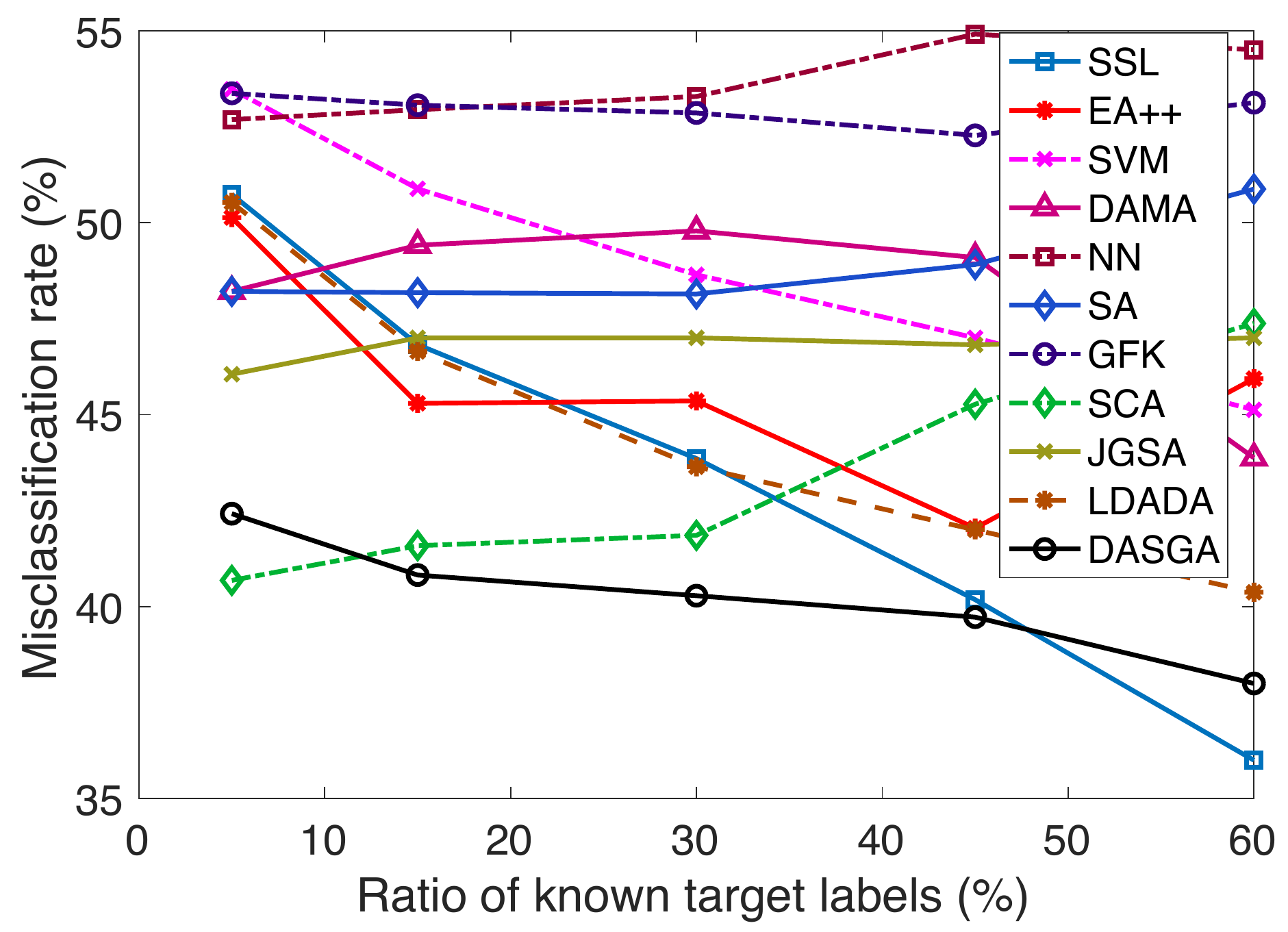}}
 \caption{Misclassification rates of target samples for the Facebook data }
  \vspace{-0cm}
  \label{fig:Facebook_errors}
\end{figure}

\subsubsection{Experiments on the Facebook data set}

As the Facebook data set involves a pure graph environment, the two graphs are embedded into an Euclidean domain via the MDS algorithm as in the Amazon data set in order to provide feature representations for the other algorithms than DASGA and SSL. The parameters of DASGA are set as $\mu_1=1$, $\mu_2=1$, and $\sizeU=8$. The misclassification rates of the compared methods are presented in Figure \ref{fig:Facebook_errors}. The classification errors of all methods are relatively high in this experiment, which can be explained by observing the challenging structure of the data set in Figure \ref{fig:Facebook_graphs}. The proposed DASGA method is seen to generally outperform the other methods. It is interesting to compare DASGA to the reference graph-based SSL method. When the ratio of available target labels is relatively small, DASGA performs better than SSL thanks to the information of the label spectrum transmitted from the source graph. Meanwhile, when the ratio of available target labels exceeds $50\%$, the SSL method has sufficient information to diffuse in the target graph and it can guess the label function more accurately than DASGA. This is coherent with the principle of domain adaptation: learning the label spectrum from an exemplar source graph improves the performance in the target graph when the label information is restricted in the target graph, which is typically the case in a domain adaptation problem.

\subsection{Stabilization and Sensitivity Analysis of the Proposed Algorithm}

\begin{figure}[t]
\begin{center}
     \subfigure[COIL-20 data]
       {\label{fig:coil_err_obj_iter}\includegraphics[height=3.35cm]{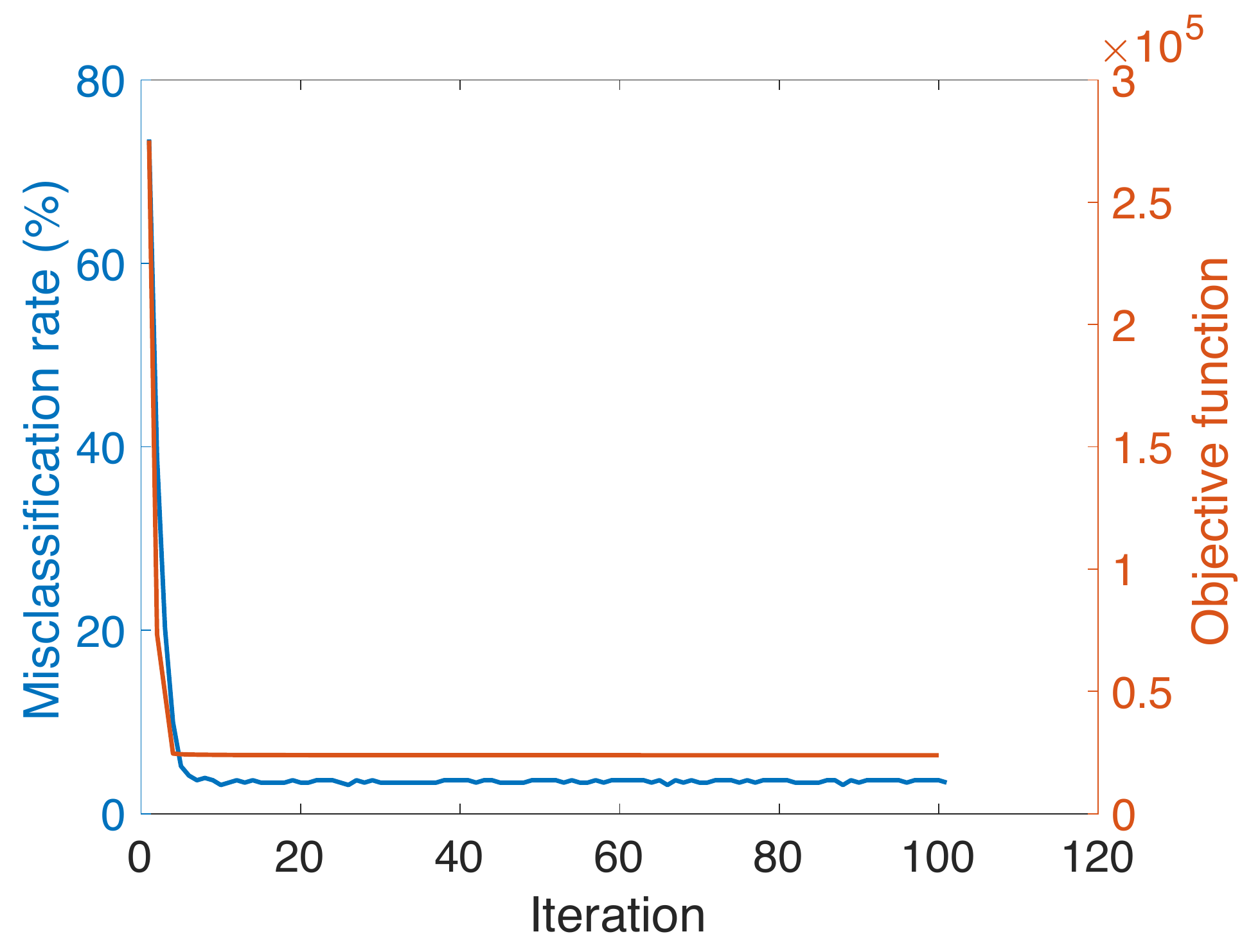}}
      \subfigure[MIT-CBCL data]
       {\label{fig:mit_err_obj_iter}\includegraphics[height=3.2cm]{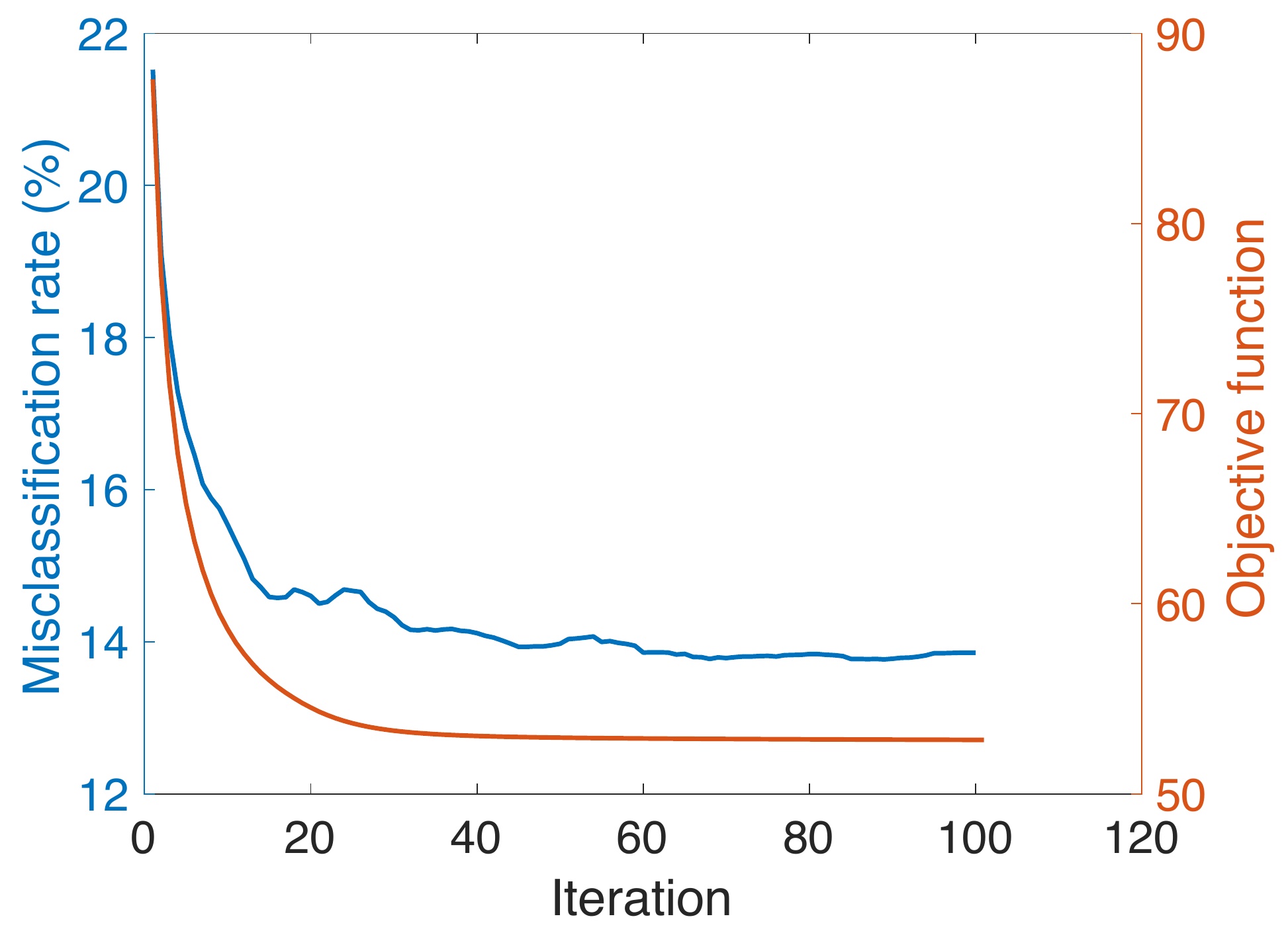}}
 \end{center}
 \caption{Evolution of the objective function and the misclassification rate throughout the iterations}
 \label{fig:coil_mit_obj_err_vs_iter}
\end{figure}

We first study the behavior of the proposed DASGA algorithm throughout the iterative optimization procedure. We examine the variations of the objective function and the misclassification rate of target samples during the iterations. The value of the objective function \eqref{eq:prob_form_general} is evaluated in each iteration of the alternating optimization procedure, as well as the misclassification rate given by the solution computed in each iteration. The evolutions of the objective function and the misclassification rate are shown for the COIL-20 and the MIT-CBCL data sets in Figure \ref{fig:coil_mit_obj_err_vs_iter}. The results confirm that the objective function decreases monotonically throughout the iterations and converges as discussed in Section \ref{ssec:overall_opt}. The misclassification rate also has the general tendency to decrease during the iterations. The rate of decrease of the misclassification error follows closely that of the objective function in both data sets. This suggests that the objective function \eqref{eq:prob_form_general} underlying the proposed method captures well the actual performance of classification.

%
%

Next, we study the sensitivity of the DASGA method to the choice of the algorithm parameters.

\begin{table}[!h]
\centering
\begin{tabular}{|c|c|c|c|c|}
	\hline
	 & $\mu_2=0.1$ & $\mu_2=0.5$ & $\mu_2=1$  & $\mu_2=1.5$\\
	\hline
	$\mu_1=10^{-4}$ & 2.09 & 2.13 &2.13 & 2.13\\
	\hline
	$\mu_1=10^{-3}$ & 1.35 & 1.35 & 1.35 & 1.35\\
	\hline
	$\mu_1=10^{-2}$ & 0.63 & 0.58 & 0.62 & 0.61\\
	\hline
	$\mu_1=0.1$ & 0.56 & 0.54 & 0.54 & 0.55\\
	\hline
	$\mu_1=0.5$ & 0.51 & 0.51 & 0.53 & 0.51\\
	\hline
	$\mu_1=1$ & 0.49 & 0.52 & 0.52 & 0.53\\
	\hline
		\hline
\end{tabular}
\begin{tabular}{|c|c|c|c|c|}
	\hline
	 & $\mu_2=0.1$ & $\mu_2=0.5$ & $\mu_2=1$  & $\mu_2=1.5$\\
	\hline
	$\mu_1=10^{-4}$ & 16.09 & 15.45 & 15.26 & 15.23\\
	\hline
	$\mu_1=10^{-3}$ & 13.08 & 12.11 & 11.87 & 11.83\\
	\hline
	$\mu_1=10^{-2}$ & 10.74 & 10.63 & 10.36 & 10.48\\
	\hline
	$\mu_1=0.1$ & 10.20 & 8.42 & 8.41 & 9.83\\
	\hline
	$\mu_1=0.5$ & 10.21 & 8.82 & 8.62 & 8.41\\
	\hline
	$\mu_1=1$ & 10.21 & 8.83 & 8.66 & 8.53\\
	\hline
\end{tabular}
\caption{Variation of the target misclassification rate (in percentage) with algorithm weight parameters $\mu_1$, $\mu_2$ on the synthetic data sets. Ratio of known target labels is $5\%$. Upper table: Synthetic dataset-1. Lower table: Synthetic dataset-2.}
\label{table:syn_err_vs_mu1mu2}
\end{table}

\begin{table}[!h]
\centering
\begin{tabular}{|c|c|c|c|c|}
	\hline
	 & $\mu_2=0.1$ & $\mu_2=0.5$ & $\mu_2=1$  & $\mu_2=1.5$\\
	\hline
	$\mu_1=10^{-4}$ & 15.97 & 14.61 & 13.48 & 12.75\\
	\hline
	$\mu_1=10^{-3}$ & 15.65 & 15.16 & 12.75 & 12.99\\
	\hline
	$\mu_1=10^{-2}$ & 12.26 & 10.06 & 11.25 & 10.46\\
	\hline
	$\mu_1=0.1$ & 11.97 & 7.86 & 6.81 & 8.12\\
	\hline
	$\mu_1=0.5$ & 11.97 & 8.90 & 7.88 & 7.80\\
	\hline
	$\mu_1=1$ & 12.03 & 8.87 & 7.30 & 6.93\\
	\hline
		\hline
\end{tabular}
\begin{tabular}{|c|c|c|c|c|}
	\hline
	 & $\mu_2=0.1$ & $\mu_2=0.5$ & $\mu_2=1$  & $\mu_2=1.5$\\
	\hline
	$\mu_1=10^{-4}$ & 15.74 & 15.28 & 17.36 & 16.84\\
	\hline
	$\mu_1=10^{-3}$ & 15.42 & 16.43 & 16.99 & 17.33\\
	\hline
	$\mu_1=10^{-2}$ & 17.30 & 18.03 & 15.91 & 15.83\\
	\hline
	$\mu_1=0.1$ & 17.48 & 16.64 & 16.78 & 15.71\\
	\hline
	$\mu_1=0.5$ & 17.45 & 14.14 & 14.09 & 14.55\\
	\hline
	$\mu_1=1$ & 17.80 & 15.91 & 12.84 & 14.90\\
	\hline
\end{tabular}
\caption{Variation of the target misclassification rate (in percentage) with algorithm weight parameters $\mu_1$, $\mu_2$ on the MIT-CBCL face image data set. Ratio of known target labels is $4.16\%$. Upper table: Source: Pose 1, Target: Pose 5. Lower table: Source: Pose 1, Target: Pose 9.}
\label{table:mit_err_vs_mu1mu2}
\end{table}

\begin{table}[!h]
\centering
\begin{tabular}{|c|c|c|c|c|}
	\hline
	 & $\mu_2=0.1$ & $\mu_2=0.5$ & $\mu_2=1$  & $\mu_2=1.5$\\
	\hline
	$\mu_1=10^{-4}$ & 0.53 & 0.53 & 0.53 & 0.53\\
	\hline
	$\mu_1=10^{-3}$ & 0.53 & 5.71 & 4.66 & 5.71\\
	\hline
	$\mu_1=10^{-2}$ & 3.50 & 2.04 & 4.04 & 11.60\\
	\hline
	$\mu_1=0.1$ & 0.53 & 0.53 & 0.53 & 1.05\\
	\hline
	$\mu_1=0.5$ & 0.53 & 0.53 & 0.53 & 1.05\\
	\hline
	$\mu_1=1$ & 0.53 & 0.53 & 0.53 & 2.06\\
	\hline
\end{tabular}
\caption{Variation of the target misclassification rate (in percentage) with algorithm weight parameters $\mu_1$, $\mu_2$ on the COIL-20 object image data set. Ratio of known target labels is $5.6\%$.}
\label{table:coil_err_vs_mu1mu2}
\end{table}

\begin{table}[!h]
\centering
\begin{tabular}{|c|c|c|c|c|}
	\hline
	 & $\mu_2=0.1$ & $\mu_2=0.5$ & $\mu_2=1$  & $\mu_2=1.5$\\
	\hline
	$\mu_1=10^{-4}$ & 57.69 & 57.76 & 57.76 & 57.78\\
	\hline
	$\mu_1=10^{-3}$ & 57.66 & 58.08 & 57.86 & 57.73\\
	\hline
	$\mu_1=10^{-2}$ & 59.58 & 58.95 & 59.94 & 60.60\\
	\hline
	$\mu_1=0.1$ & 61.27 & 61.51 & 62.76 & 62.87\\
	\hline
	$\mu_1=0.5$ & 61.21 & 62.45 & 62.30 & 62.36\\
	\hline
	$\mu_1=1$ & 61.82 & 62.14 & 64.25 & 63.80\\
	\hline
		\hline
\end{tabular}
\begin{tabular}{|c|c|c|c|c|}
	\hline
	 & $\mu_2=0.1$ & $\mu_2=0.5$ & $\mu_2=1$  & $\mu_2=1.5$\\
	\hline
	$\mu_1=10^{-4}$ & 1.10 & 1.10 & 1.11 & 1.10\\
	\hline
	$\mu_1=10^{-3}$ & 1.06 & 1.06 & 1.05 & 1.06\\
	\hline
	$\mu_1=10^{-2}$ & 1.04 & 1.03 & 1.02 & 1.03\\
	\hline
	$\mu_1=0.1$ & 1.05 & 1.05 & 1.05 & 1.05\\
	\hline
	$\mu_1=0.5$ & 1.06 & 1.06 & 1.05 & 1.05\\
	\hline
	$\mu_1=1$ & 1.07 & 1.05 & 1.06 & 1.06\\
	\hline
\end{tabular}
\caption{Variation of the target error with algorithm weight parameters $\mu_1$, $\mu_2$ on the Amazon data set. Ratio of known target labels is $11.43\%$. Upper table: Misclassification rate (in percentage). Lower table: RMS error}
\label{table:amazon_err_vs_mu1mu2}
\end{table}

\begin{table}[!h]
\centering
\begin{tabular}{|c|c|c|c|c|}
	\hline
	 & $\mu_2=0.1$ & $\mu_2=0.5$ & $\mu_2=1$  & $\mu_2=1.5$\\
	\hline
	$\mu_1=10^{-4}$ & 45.15 & 43.82 & 43.68 & 43.68\\
	\hline
	$\mu_1=10^{-3}$ & 43.82 & 43.38 & 43.24 & 43.24\\
	\hline
	$\mu_1=10^{-2}$ & 41.62 & 41.91 & 41.47 & 41.47\\
	\hline
	$\mu_1=0.1$ & 41.18 & 41.03 & 41.91 & 41.91\\
	\hline
	$\mu_1=0.5$ & 41.76 & 39.71 & 39.71 & 37.94\\
	\hline
	$\mu_1=1$ & 42.21 & 40.44 & 40.15 & 39.71\\
	\hline
\end{tabular}
\caption{Variation of the target misclassification rate (in percentage) with algorithm weight parameters $\mu_1$, $\mu_2$ on the Facebook data set. Ratio of known target labels is $15\%$.}
\label{table:facebook_err_vs_mu1mu2}
\end{table}

\textit{Sensitivity to the weight parameters $\mu_1$ and $\mu_2$.} 
We first examine how the choice of the weight parameters affects the algorithm performance. The target misclassification rate of the algorithm is reported for various $(\mu_1, \mu_2)$ pairs in a region of interest in Tables \ref{table:syn_err_vs_mu1mu2}-\ref{table:facebook_err_vs_mu1mu2} for all five data sets. The results indicate that although the optimal values of the $\mu_1$ and $\mu_2$ parameters may vary among different data sets, the $(\mu_1, \mu_2)$ pair yielding the smallest misclassification error is found within the region $\mu_1 \in [0.1, 1]$, $\mu_2 \in [0.5, 1.5]$ for all four data sets except Amazon. While the smallest misclassication error is obtained outside this region for the Amazon data set (with $\mu_1$ taking a smaller value), the RMS errors obtained by choosing $\mu_1$ and $\mu_2$ within this region are quite close to the optimal value.  We also observe that the algorithm error does not vary dramatically within this region. In particular, the relative difference in the misclassification error varies between  $1\%-4\%$ in all data sets inside the region. These findings suggest that it is safe to choose the weight parameters within the intervals  $\mu_1 \in [0.1, 1]$ and $\mu_2 \in [0.5, 1.5]$, where the algorithm performs sufficiently well. 

\begin{figure}[t]
\begin{center}
     \subfigure[Synthetic data set]
       {\label{fig:syn_err_vs_K}\includegraphics[height=3.2cm]{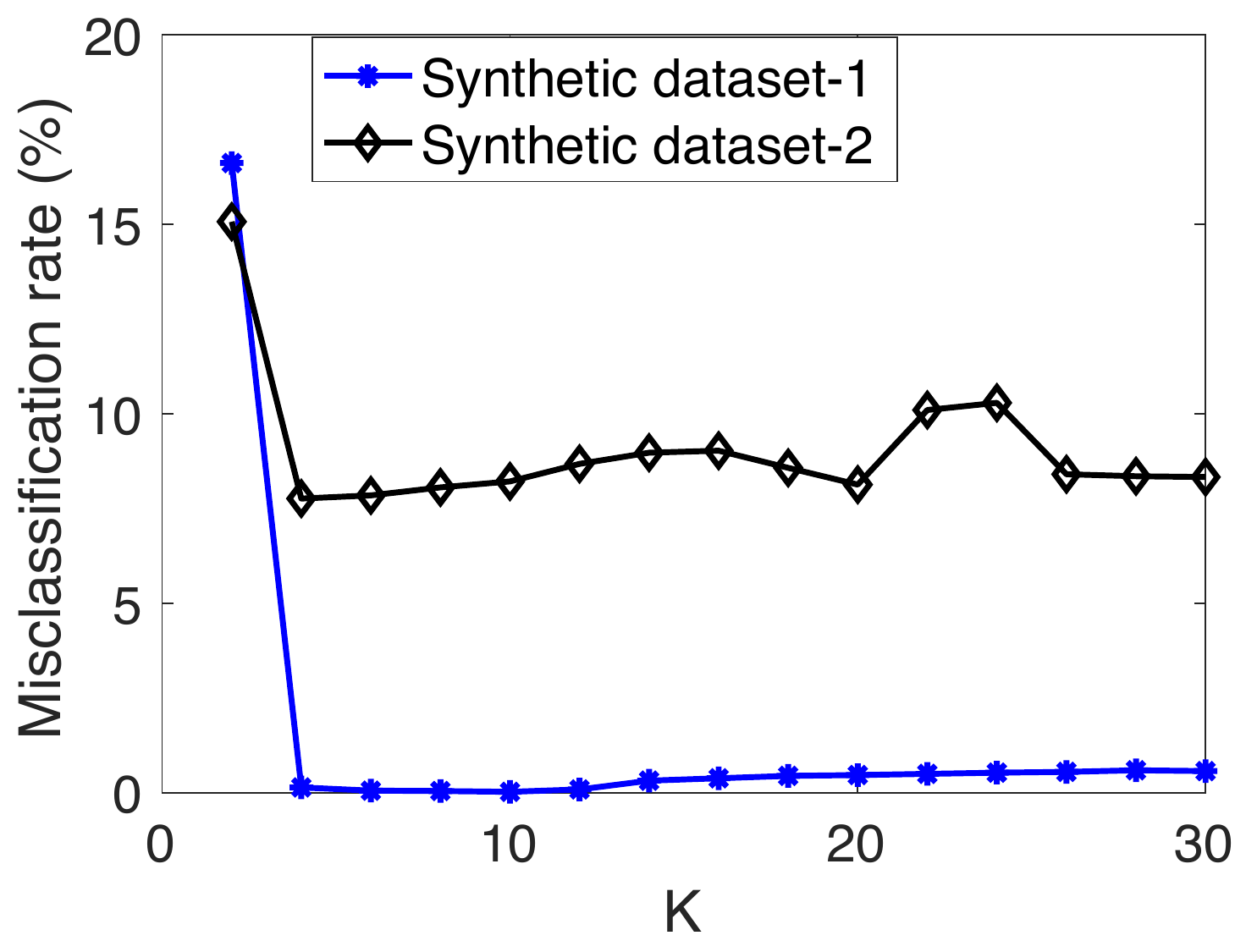}}
      \subfigure[MIT-CBCL face data set]
       {\label{fig:mit_err_vs_K}\includegraphics[height=3.2cm]{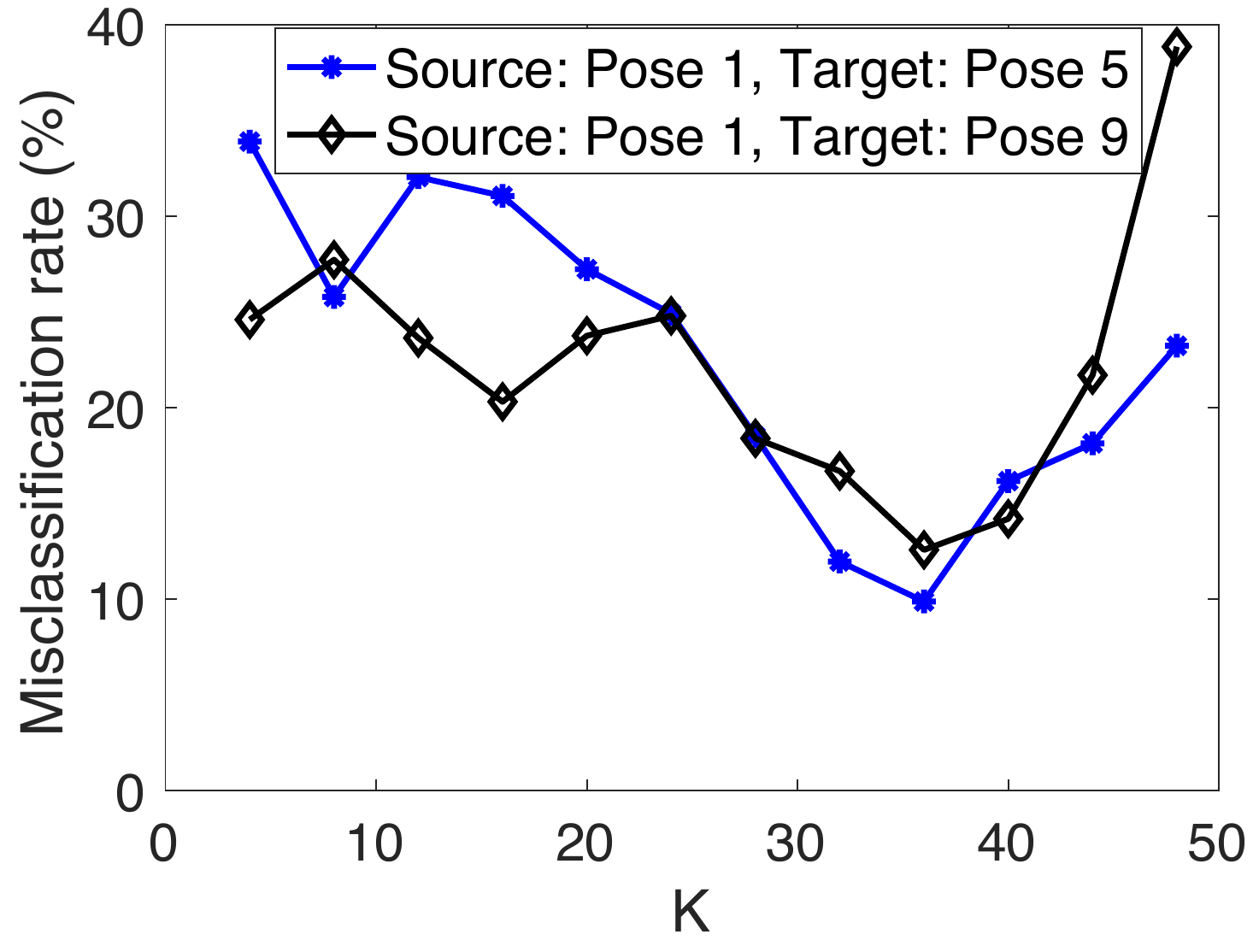}}
       \subfigure[COIL-20 object data set]
       {\label{fig:coil_err_vs_K}\includegraphics[height=3.2cm]{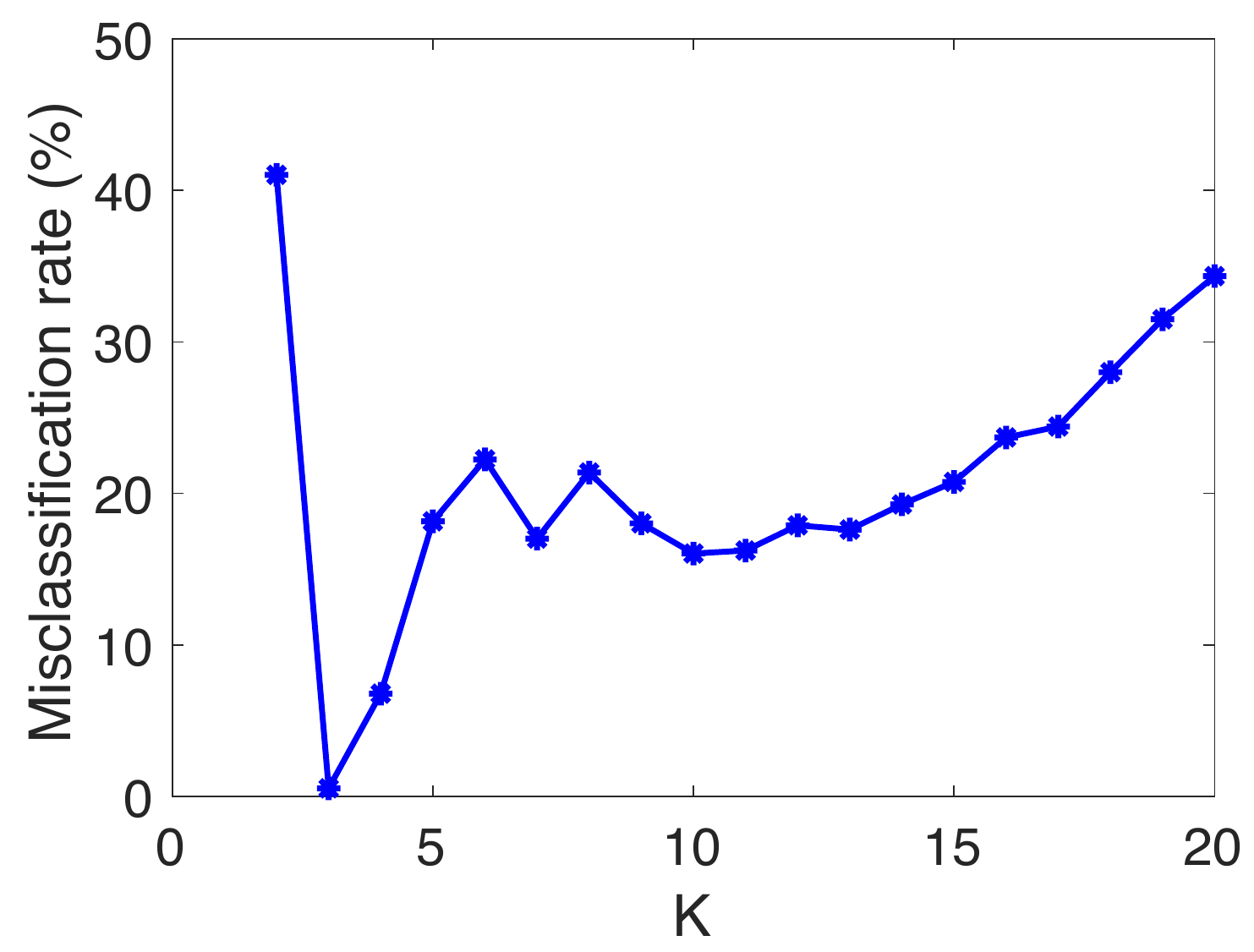}}
 \end{center}
 \vspace{-0.3cm}
 \caption{Variation of the misclassification rates of target samples with the number of neighbors $K$}
 \label{fig:err_vs_K}
\end{figure}

\textit{Sensitivity to the number of neighbors $K$.} 
Next, we study the effect of the choice of the number of nearest neighbors $K$ when constructing the source and the target graphs. The variation of the target misclassification rate with the number of nearest neighbors $K$ is examined on the synthetic, MIT-CBCL, and the COIL-20 data sets, where the source and the target graphs need to be constructed from data. The target misclassification rates are given in Figure \ref{fig:err_vs_K}. In Figure \ref{fig:syn_err_vs_K}, the algorithm performance is seen to be stable over a relatively wide range of $K$ values for the two synthetic data sets. On the other hand, we see in Figure \ref{fig:coil_err_vs_K} that the proposed method is more sensitive to the choice of the $K$ parameter in the COIL-20 data set. In particular, the optimal value of $K$ is quite small (around $3-4$). This result is in line with the intrinsic geometric properties of this data set: As the images of the objects are taken by rotating the camera around the object by varying a single camera angle parameter, the intrinsic dimension of COIL-20 is quite low. The best performance is then achieved when the graphs are constructed with a small number of neighbors, which conforms to the geometric structure of data. By comparison, the face images in the MIT-CBCL data set are rendered under a larger set of parameters related to the illumination conditions, hence the intrinsic dimension of the data set is higher than that of COIL-20. As a result, the optimal value of the $K$ parameter is seen to be larger for MIT-CBCL in Figure \ref{fig:mit_err_vs_K}.

\begin{figure}[t]
\begin{center}
     \subfigure[Synthetic data set]
       {\label{fig:syn_err_vs_R}\includegraphics[height=3.2cm]{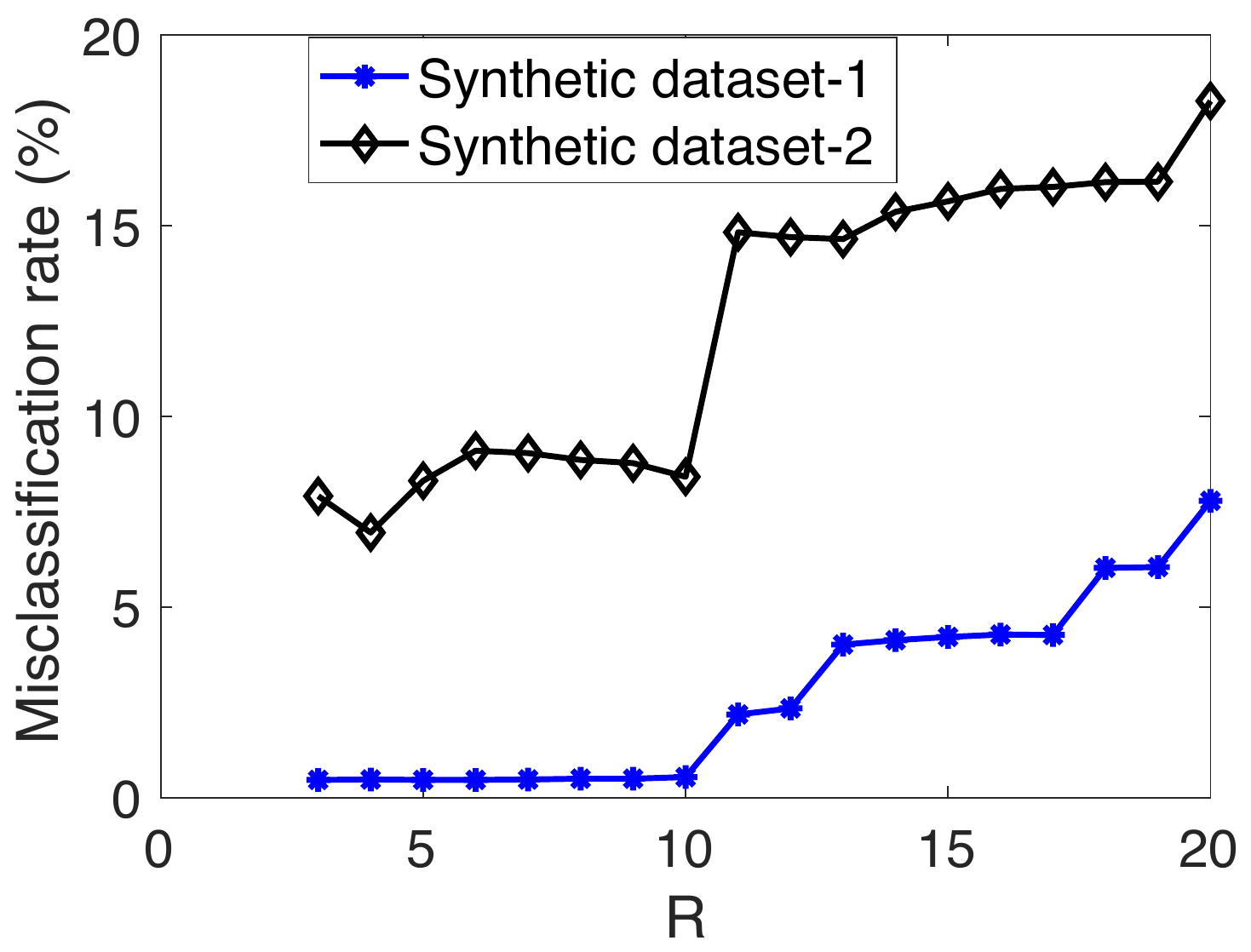}}
      \subfigure[MIT-CBCL face data set]
       {\label{fig:mit_err_vs_R}\includegraphics[height=3.2cm]{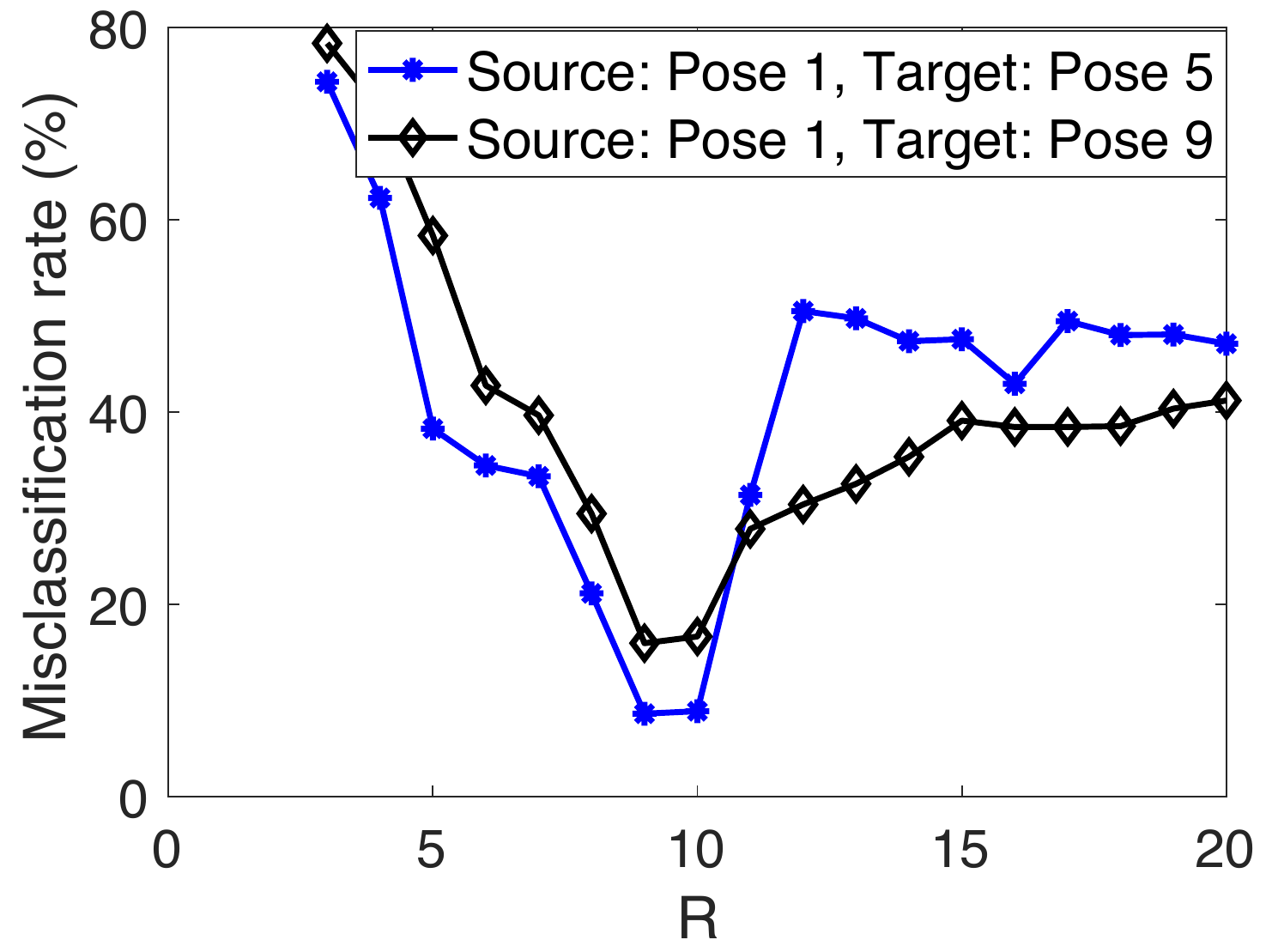}}
       \subfigure[COIL-20 object data set]
       {\label{fig:coil_err_vs_R}\includegraphics[height=3.2cm]{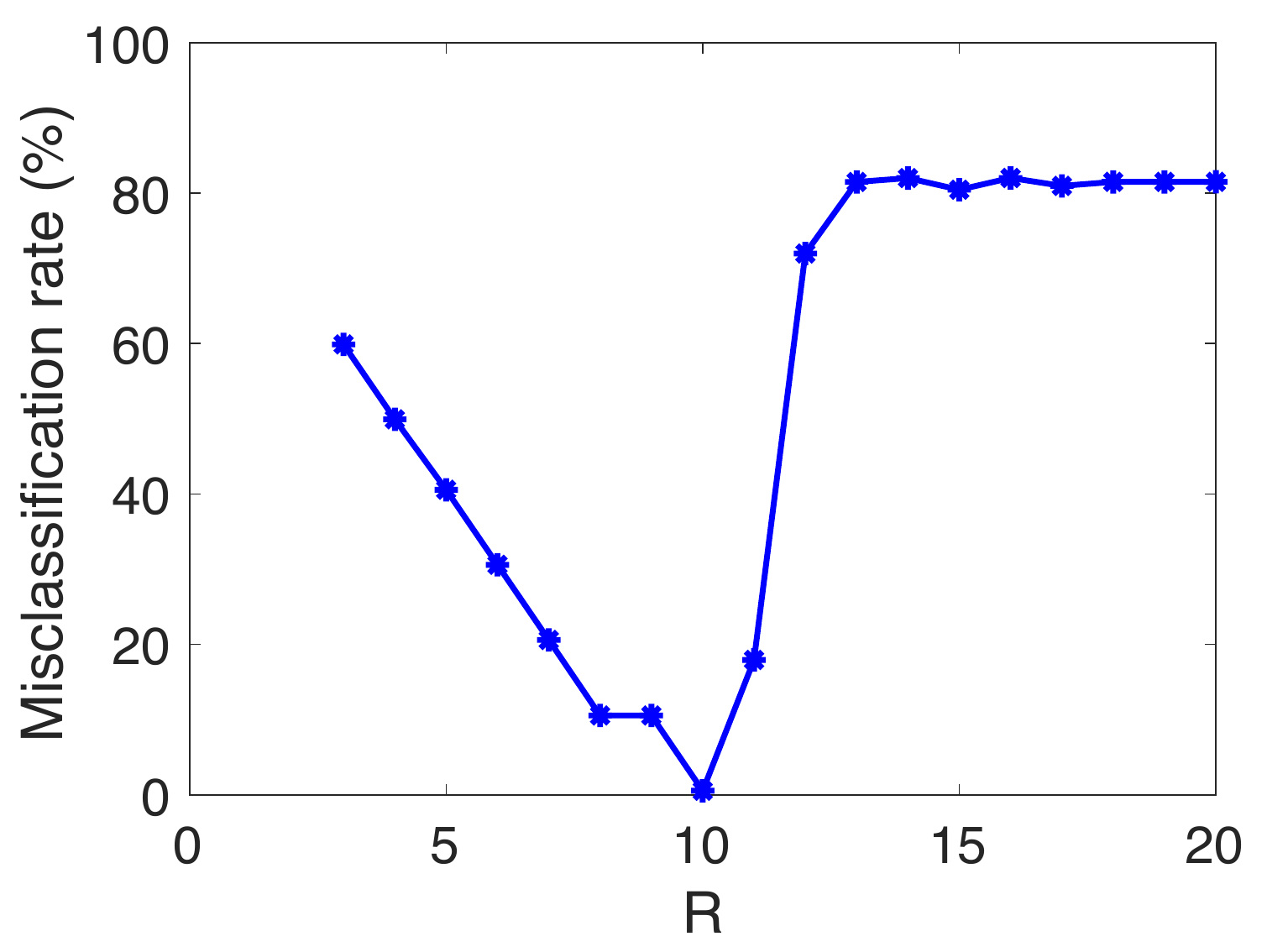}}
       \subfigure[Amazon data set]
       {\label{fig:amazon_err_vs_R}\includegraphics[height=3.2cm]{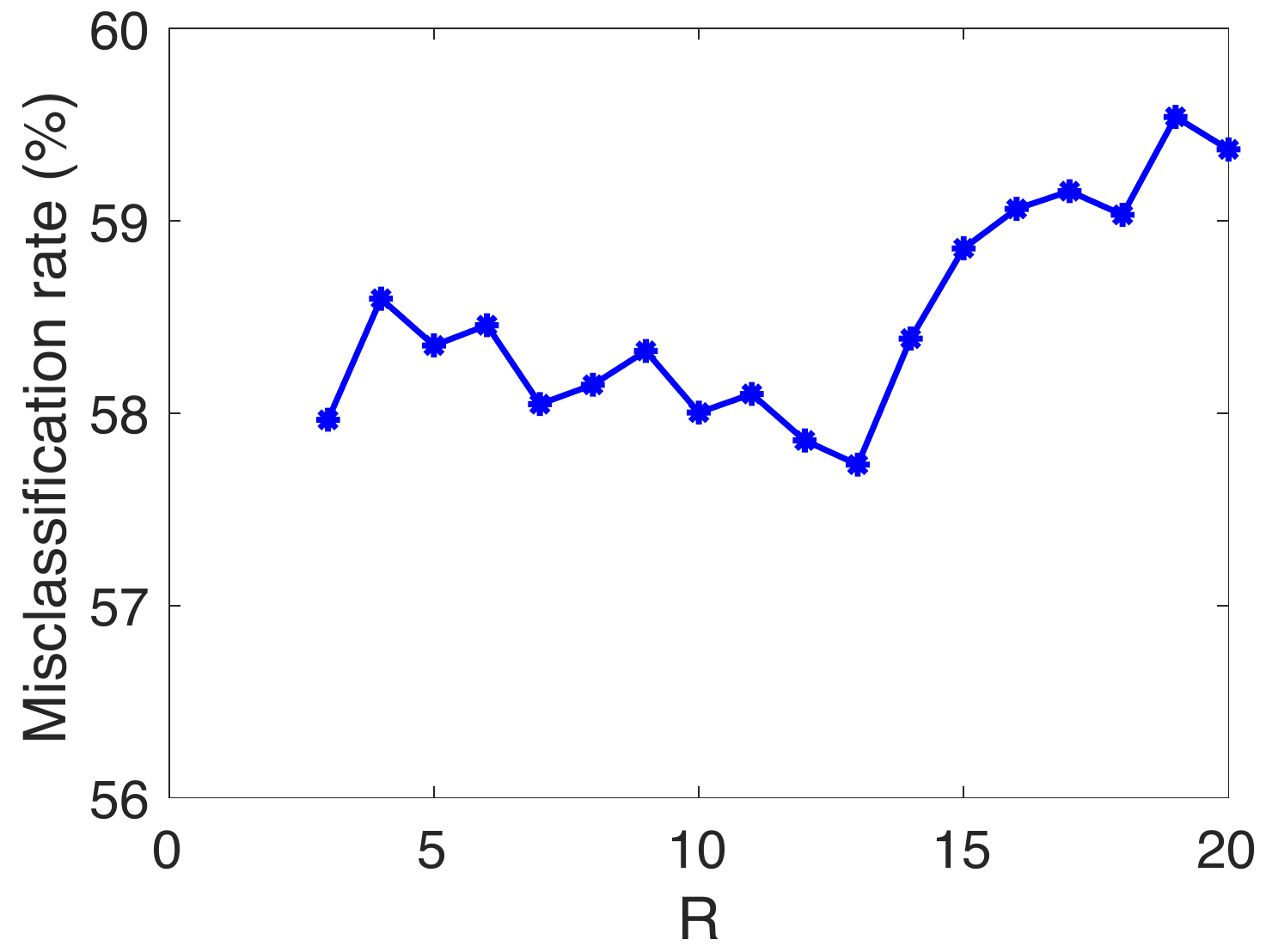}}
              \subfigure[Amazon data set]
       {\label{fig:amazon_rms_err_vs_R}\includegraphics[height=3.2cm]{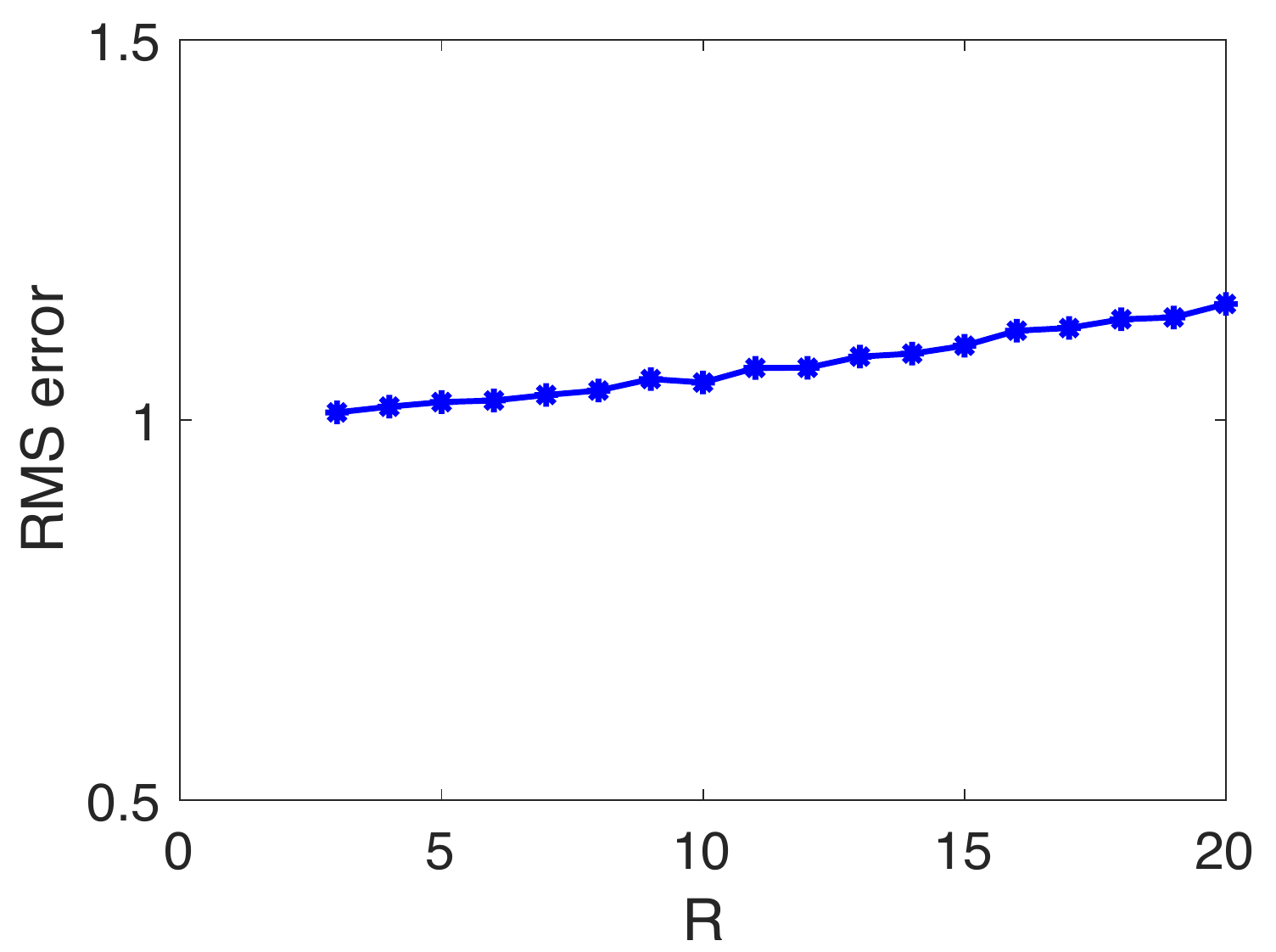}}
      \subfigure[Facebook data set]
       {\label{fig:facebook_err_vs_R}\includegraphics[height=3.2cm]{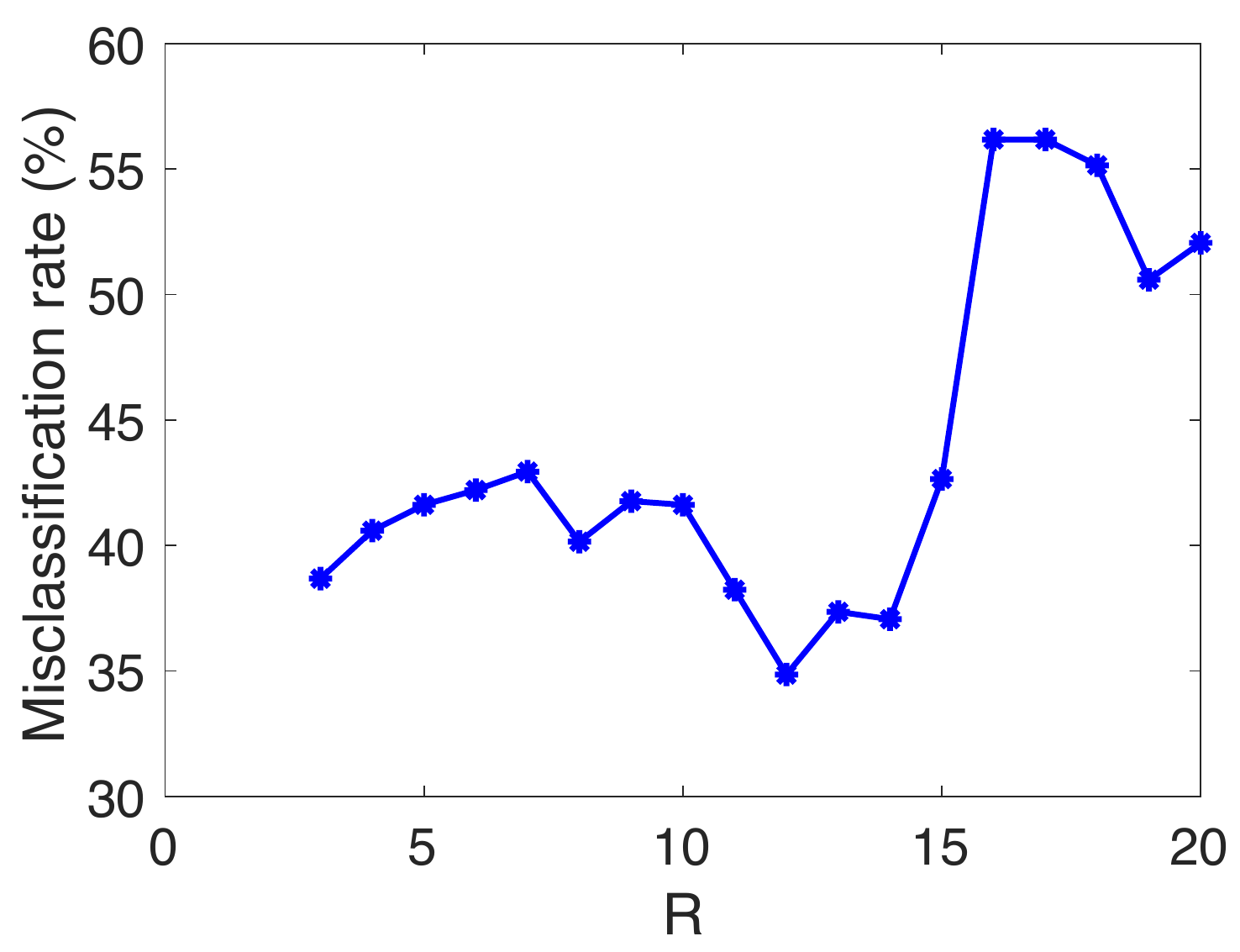}}
 \end{center}
 \vspace{-0.3cm}
 \caption{Variation of the misclassification rates of target samples with the number of basis vectors $R$}
 \label{fig:err_vs_R}
\end{figure}

\textit{Sensitivity to the number of eigenvectors $\sizeU$.} Finally, we investigate how the choice of the number of graph basis vectors $\sizeU$ used in the objective \eqref{eq:prob_form_general} affects the algorithm performance. 
The variation of the target misclassification rate is plotted with respect to the number of basis vectors $\sizeU$ in Figure \ref{fig:err_vs_R} for all data sets. The results  suggest that the variation of the misclassification rate with $\sizeU$ has similar characteristics among different data sets.  At small  $\sizeU$  values, the classification performance improves as  $\sizeU$ increases, since the label function can be approximated more accurately when more basis vectors are used. The optimal value of  $\sizeU$ is often around 9-12, and the performance tends to degrade when  $\sizeU$ is increased beyond these values. This is because increasing $\sizeU$ too much results in poor regularization and increases the misclassification error, which is also consistent with the theoretical bound in Proposition \ref{prop:diff_fs_ft_rates}.


%
%

\section{Conclusion}
\label{sec:concl}

We have considered the problem of domain adaptation on graphs. Given a source graph with sufficiently many labeled nodes and a target graph, we have proposed a graph-based domain adaptation algorithm that estimates a label function on the target graph, relying on the assumption that the frequency content of the source and target label functions have similar characteristics. Our method is based on the idea of learning a pair of coherent bases on the source and the target graphs. The learnt bases not only resemble in terms of their spectral content, but also ``align'' the two graphs such that the label functions on the two graphs can be reconstructed with similar coefficients. The proposed domain adaptation algorithm is completely graph-based and is particularly applicable in learning problems defined purely on graph domains where no physical embedding of data samples is available. The performance of the proposed method is demonstrated mainly in data classification applications; however, it can potentially be applied to a wide range of machine learning problems concerning the inference of the unknown values of a graph function from available values. The exploration of information transfer based on more elaborate graph kernels than the graph Fourier basis, or the extension of the method to explicitly employ data embeddings in addition to graph models in order to improve its performance on data sets with available ambient space representations remain as future directions.


%

\appendices
\section{Proof of Proposition \ref{prop:diff_fs_ft_rates}}
\label{app:pf:prop:diff_fs_ft_rates}

\begin{proof}

The solution $\ralpha^s, \ralpha^t, \rT$ of Problem 3 gives the estimated source and target label functions as $ f^s = \rU^s  \ralpha^s$ and $f^t = \rU^t \rT  \ralpha^t$. The rates of variation of $f^s$ and $f^t$ on the source and target graphs are given by
\begin{equation}
\begin{split}
(f^s)^T L^s f^s &=   (\ralpha^s)^T  (\rU^s)^T  L^s \rU^s  \ralpha^s =  (\ralpha^s)^T  \rLambda^s  \ralpha^s \\
(f^t)^T L^t f^t &=   (\rT \ralpha^t)^T    (\rU^t)^T  L^t \rU^t \rT  \ralpha^t =  (\rT \ralpha^t)^T  \rLambda^t \rT  \ralpha^t
\end{split}
\end{equation}
where $\rLambda^s$ and $\rLambda^t$ are the diagonal matrices consisting of the $\sizeU$ smallest eigenvalues of respectively $L^s$ and $L^t$, such that $\rLambda^s_{ii} = \lambda_i^s$ and $\rLambda^t_{ii} = \lambda_i^t$, for $i=1, \dots, \sizeU$.

The difference between the rates of variations of $f^s$ and $f^t$ can then be bounded as
\begin{equation}
\label{eq:fs_ft_sep3terms}
\begin{split}
& | (f^s)^T L^s f^s  -  (f^t)^T L^t f^t   | = |  (\ralpha^s)^T  \rLambda^s  \ralpha^s -   (\rT \ralpha^t)^T   \rLambda^t \rT  \ralpha^t  | \\
&= | 
  (\ralpha^s)^T  \rLambda^s  \ralpha^s 
 -  (\ralpha^s)^T  \rLambda^t  \ralpha^s  +   (\ralpha^s)^T  \rLambda^t  \ralpha^s    \\
& -   (\ralpha^t)^T  \rLambda^t  \ralpha^t   + (\ralpha^t)^T  \rLambda^t  \ralpha^t  
 -  (\rT \ralpha^t)^T   \rLambda^t \rT  \ralpha^t  
 | \\
& \leq  
| (\ralpha^s)^T ( \rLambda^s -  \rLambda^t) \ralpha^s  |
+ |  (\ralpha^s)^T \rLambda^t \ralpha^s -  (\ralpha^t)^T \rLambda^t \ralpha^t   |  \\
&+  | (\ralpha^t)^T \rLambda^t \ralpha^t -  (\rT \ralpha^t)^T \rLambda^t \rT \ralpha^t   |.
\end{split}
\end{equation}

In the following, we derive an upper bound for each one of the three terms at the right hand side of the inequality in \eqref{eq:fs_ft_sep3terms}. The first term is bounded as
\begin{equation}
|  (\ralpha^s)^T ( \rLambda^s -  \rLambda^t) \ralpha^s   | 
\leq \| \ralpha^s \|^2  \| \rLambda^s -  \rLambda^t  \|
\leq  \calpha^2 \dlambda.
\end{equation}
Here the first inequality is due to the Cauchy-Schwarz inequality, and the second inequality follows from the fact that the operator norm of the matrix $\rLambda^s -  \rLambda^t $ is given by the magnitude of its largest eigenvalue, which is upper bounded by $\dlambda$ due to the assumption $| \lambda_i^s - \lambda_i^t | \leq \dlambda$ for all $i$.

Next, we bound the second term in \eqref{eq:fs_ft_sep3terms} as 
\begin{equation}
\begin{split}
& |  (\ralpha^s)^T \rLambda^t \ralpha^s -  (\ralpha^t)^T \rLambda^t \ralpha^t    | \\
&= |  
(\ralpha^s)^T \rLambda^t \ralpha^s  - (\ralpha^s)^T \rLambda^t \ralpha^t
+
(\ralpha^s)^T \rLambda^t \ralpha^t  - (\ralpha^t)^T \rLambda^t \ralpha^t
| \\
& \leq 
|  (\ralpha^s)^T \rLambda^t  (\ralpha^s -  \ralpha^t) |
+ | (\ralpha^s -  \ralpha^t)^T  \rLambda^t  \ralpha^t  | \\
& \leq \|  \ralpha^s \|   \| \rLambda^t \|     \|  \ralpha^s -  \ralpha^t  \|  
+  \|  \ralpha^s -  \ralpha^t  \|   \| \rLambda^t \|   \|  \ralpha^t \| 
\leq 2 \calpha \lambdaR \dalpha
\end{split}
\end{equation}
where the last equality follows from the fact that the matrix norm $\|  \rLambda^t \|$ is bounded by the largest eigenvalue of $\rLambda^t $, which is smaller than $\lambdaR$ by our assumption.

Lastly, the third term in \eqref{eq:fs_ft_sep3terms} can be bounded as
\begin{equation}
\label{eq:third_term_fs_ft}
\begin{split}
& | (\ralpha^t)^T \rLambda^t \ralpha^t -  (\rT \ralpha^t)^T \rLambda^t \rT \ralpha^t   | \\
& \leq 
 | (\ralpha^t)^T \rLambda^t \ralpha^t -   (\ralpha^t)^T \rLambda^t  \rT \ralpha^t  \\
 & \quad + (\ralpha^t)^T \rLambda^t  \rT \ralpha^t 
  -  (\rT \ralpha^t)^T \rLambda^t \rT \ralpha^t   | \\
& \leq    | (\ralpha^t)^T \rLambda^t ( \ralpha^t -   \rT \ralpha^t ) |
+ |  (\ralpha^t - \rT \ralpha^t)^T \rLambda^t  \rT \ralpha^t   |  \\
& \leq   \|  \ralpha^t \|^2   \|  \rLambda^t  \|  \| I - \rT \|  
+ \|  \ralpha^t \|^2   \| I - \rT \|    \|  \rLambda^t  \|  \|  \rT \|.
\end{split}
\end{equation}

Bounding the norm of $ \rT $ as
\begin{equation}
\|  \rT  \| = \| I + \rT - I  \| \leq \| I \| +  \| \rT - I \|
\leq 1 + \dT
\end{equation}
and using also the assumption $\| \rT - I \| \leq \dT$ in \eqref{eq:third_term_fs_ft}, we get
\begin{equation}
\begin{split}
| (\ralpha^t)^T \rLambda^t \ralpha^t &-  (\rT \ralpha^t)^T \rLambda^t \rT \ralpha^t   | \\
&\leq \calpha^2 \lambdaR \dT + \calpha^2 \lambdaR \dT (1 + \dT).
\end{split}
\end{equation}
Finally, putting together the upper bounds for all the three terms in \eqref{eq:fs_ft_sep3terms}, we get the stated result
\begin{equation}
\begin{split}
 | (f^s)^T L^s f^s  -  (f^t)^T L^t f^t   | & \leq 
 \calpha^2 \dlambda + 2 \calpha \lambdaR \dalpha  \\
 &+ \calpha^2 \lambdaR (2 \dT + \dT^2).
\end{split}
\end{equation}
\end{proof}

\section{Convergence Analysis of the SQP Method}
\label{app:conv_anly_sqp}
In order to analyze the convergence of the SQP algorithm, we first inspect the Hessian $\nabla_{\rt \rt}^2 \L(\rt, \eta)$ of the Lagrangian function. Recall that the matrices $2 \mu_2 F^T F$ and $\nabla^2_{\rt \rt} \, g(\rt, \eta)$ in the expression of $\nabla_{\rt \rt}^2 \L(\rt, \eta)$ are both diagonal. From the assumption that  $\mu_2> \eta_j^*$, we get that at the local solution  $(\rt^*, \eta^*)$ the entries of the diagonal matrix $2 \mu_2 F^T F - \nabla^2_{\rt \rt} \, g(\rt, \eta) $ are lower bounded as
\begin{equation}
\begin{split}
\label{eq:sqp_pf_diagmat_pd}
[2 \mu_2 F^T F - \nabla^2_{\rt \rt} \, g(\rt, \eta) |_{( \rt^*, \eta^*)}  ]_{kk} &= 2 \mu_2 \rW_{ij}^2 - 2 \eta^*_j \\
&> 2 \mu_2 - 2 \eta^*_j 
>0
\end{split}
\end{equation}
for $k= \sizeU(j - 1) + i$ with $i,j=1, \dots, \sizeU$, where the first inequality simply follows from the fact that $\rW_{ij} > 1$ as the definition \eqref{eq:Wij_defn} implies. This shows that the diagonal matrix $2 \mu_2 F^T F - \nabla^2_{\rt \rt} \, g(\rt, \eta) $ is positive definite at the local solution $(\rt^*, \eta^*)$. Then, the Hessian $\nabla_{\rt \rt}^2 \L(\rt, \eta)$ of the Lagrangian is also positive definite at $(\rt^*, \eta^*)$, since for any $x \in \R^{\sizeU^2 }$ with $x \neq 0$ we have
\begin{equation}
\begin{split}
& x^T [ \nabla_{\rt \rt}^2 \L(\rt, \eta) |_{(\rt^*, \eta^*)} ] x  \\
&= 2 x^T A^T A x 
+ x^T [2 \mu_2 F^T F - \nabla^2_{\rt \rt} \, g(\rt, \eta) |_{( \rt^*, \eta^*)}  ] x >0
\end{split}
\end{equation}
which follows from \eqref{eq:sqp_pf_diagmat_pd} and the positive semi-definiteness of $A^T A$.

Next, from the form of the constraint gradients  $\nabla_{\rt}  \, g_j$ in \eqref{eq:constraint_gradient}, and the constraint that the columns of $\rT$ must have unit norm at a local solution, it is easy to observe that the set of the equality constraint gradients $\{ \nabla_{\rt}  \, g_j, \ j=1, \dots, \sizeU \}$ is linearly independent, which are the only active constraints of the optimization problem. Since the objective and the constraint functions are twice differentiable with Lipschitz continuous second derivatives, the active constraint gradients are linearly independent, and the Hessian of the Lagrangian function is positive definite at the local solution $( \rt^*, \eta^*)$, by \cite[Theorem 18.4]{NocedalW06}, the SQP algorithm converges to the local solution $( \rt^*, \eta^*)$ provided that the initial solution $(\rt, \eta)$ is sufficiently close to  $( \rt^*, \eta^*)$.

\ifCLASSOPTIONcompsoc
  \section*{Acknowledgments}
\else
  \section*{Acknowledgment}
\fi
This work has been supported by the T\"UB\.ITAK 2232 research scholarship.

\ifCLASSOPTIONcaptionsoff
  \newpage
\fi



%


\bibliographystyle{IEEEtran}
\bibliography{refs}

\end{document}